\pdfoutput=1

\documentclass[10pt,twocolumn,letterpaper]{article}

\usepackage[pagenumbers]{cvpr} %

\usepackage{cuted} %

\usepackage{currfile} %

\usepackage{caption} %
\usepackage{microtype} %

\usepackage{amsmath,amssymb,amsthm}
\usepackage{booktabs}
\usepackage{multirow}
\newtheoremstyle{named}{\topsep}{\topsep}{\itshape}{0pt}{\bfseries}{.}{.5em}{\thmname{#1}\thmnumber{ #2}\thmnote{. #3}}
\theoremstyle{named}
\newtheorem{lemma}{Lemma}
\newtheorem{proposition}{Proposition}

\usepackage{colortbl}
\usepackage{etoolbox}
\newcommand{\tabunderline}[1]{\setbox0\hbox{#1}\rlap{\rule[-0.9pt]{\wd0}{0.35pt}}\box0}
\AtBeginEnvironment{table}{\let\underline\tabunderline}
\AtBeginEnvironment{table*}{\let\underline\tabunderline}
\usepackage{morefloats}
\usepackage{algorithm}      %
\usepackage{algpseudocode}

\usepackage{placeins}
\makeatletter

\newcommand{\pairfigtab}[4]{%
\begin{figure}[htbp]\centering
\begin{minipage}[t]{0.49\textwidth}\vspace{0pt}\centering\includegraphics{#1}\caption{#2}\label{#3}\end{minipage}\hfill
\begingroup
\renewenvironment{table}[1][]{\begin{minipage}[t]{0.48\textwidth}\vspace{0pt}\centering\def\@captype{table}}{\end{minipage}}%
\input{#4}%
\endgroup
\end{figure}}
\newcommand{\sidecapfig}[3]{%
\begin{figure}[htbp]\centering
\begin{minipage}[c]{0.49\textwidth}\centering\includegraphics{#1}\end{minipage}\hfill
\begin{minipage}[c]{0.46\textwidth}\caption{#2}\label{#3}\end{minipage}
\end{figure}}
\makeatother

\definecolor{cvprblue}{rgb}{0.21,0.49,0.74}
\usepackage[pagebackref,breaklinks,colorlinks,allcolors=cvprblue]{hyperref}

\title{Sightline: Long-Window Speculative Decoding for Fast SVG Generation}

\author{%
Yanchao Li$^{1,2}$\quad
Ben Gao$^{3}$\quad
Jiaqing Xie$^{4,2}$\quad
Yusu Zhang$^{5}$\quad
Beilun Wang$^{6}$\\
Yanbo Wang$^{7}$\quad
Xiaohua Xu$^{8}$\quad
Nanqing Dong$^{2,9}$\quad
Tianfan Fu$^{1,2*}$\quad
Zhongyu Huang$^{2*}$\\[0.6em]
\normalsize
$^{1}$Nanjing University\quad
$^{2}$Shanghai Artificial Intelligence Laboratory\quad
$^{3}$Wuhan University\\
\normalsize
$^{4}$Fudan University\quad
$^{5}$Shanghai Jiao Tong University\quad
$^{6}$Southeast University\\
\normalsize
$^{7}$North University of China\quad
$^{8}$University of Science and Technology of China\\
\normalsize
$^{9}$Shanghai Innovation Institute\\[0.4em]
\small
$^{*}$Corresponding authors
}

\hypersetup{pdftitle={Sightline: Long-Window Speculative Decoding for Fast SVG Generation},
  pdfauthor={Yanchao Li, Ben Gao, Jiaqing Xie, Yusu Zhang, Beilun Wang, Yanbo Wang, Xiaohua Xu, Nanqing Dong, Tianfan Fu, Zhongyu Huang}}

\begin{document}
\maketitle
\begin{abstract}
Scalable Vector Graphics (SVG) generation models write a graphic as code with one forward pass per token, so drawing is slow.
Speculative decoding speeds them up losslessly, and block drafters guess many tokens per pass.
On SVG, many of their rounds accept the whole draft and end only at the verification window.
We find that their misses are rare and clumped, and that they stay accurate far deeper into a block than on natural language.
A run-length identity and three propositions show that a longer window pays once the candidates are wide, up to the drafter's horizon.
We propose Sightline, which pairs a window as long as the drafter's horizon with a draft tree wide enough to fill it.
Experiments cover text-to-SVG and image-to-SVG with three models on six benchmarks from MMSVG, MMSVG-2M and SArena.
In every setting, Sightline is the fastest of the compared lossless decoders.
On SArena, it stays the fastest when every decoder shares one loop control.
It also leads in almost every setting of six models from four other families, which write SVG code, compressed SVG tokens or Lottie JSON.
Analyses trace the lead to rounds that shorter windows cut off, both in loop-free content and in loops.
Speculative decoding also saves time for choosing samples by rendering, which beats choosing by likelihood.
These results suggest that speculative decoding on vector graphics should verify as far as the drafter sees.
Code and data are available at \href{https://github.com/LiZaiyuan0619/Sightline}{\texttt{github.com/LiZaiyuan0619/Sightline}}.
\end{abstract}

\section{Introduction}
\label{sec:intro}
\begin{figure}[t]
\centering
\includegraphics{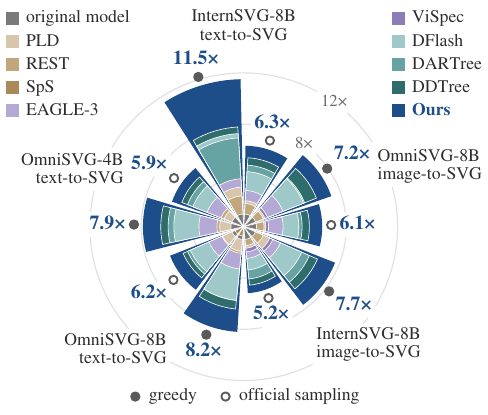}
\caption{\textbf{Speedups of nine lossless decoders over the original model.} Each petal averages one model--task pair over its six benchmarks under one decoding. Each band runs from the center to a decoder's speedup, so the outermost color is the fastest. Numbers give our mean speedups, and Figure~\ref{fig:main} gives all 60 settings.}
\label{fig:teaser}
\end{figure}

Scalable Vector Graphics (SVG) generation models draw a graphic by writing its code one token at a time~\cite{Rodriguez_2025_CVPR,yang2025omnisvg,wang2026internsvg}, so a drawing of a thousand tokens or more is slow.
The official pipeline of OmniSVG even samples eight candidates per prompt and keeps the first valid one.
Speculative decoding speeds up generation losslessly, as a cheap drafter guesses the next tokens and the target model checks them all in one forward pass~\cite{3618408.3619203}.
With a draft tree, block drafters that guess a block of tokens at once are among the fastest speculative decoders~\cite{ringel2026accelerating}.

On SVG, these drafters stop short.
The target often accepts every token they guess, so a round ends only because the window does.
Such rounds appear on some text benchmarks too~\cite{li2026dartree,wu2026ceiling}, but on SVG they are common.
The windows were set on natural language, where a longer block adds little under standard training~\cite{chen2026dflash,zhang2026d,lin2026libraspec}.
SVG generation models also often repeat code up to the length limit, and choosing among their samples needs a judge, since likelihood favors the samples that loop~\cite{Holtzman2020The}.

We find that, on SVG, block drafters miss rarely and in clumps, and see far within a block.
Most tokens of an SVG are settled by the code before them, so, with enough candidates per position, the runs between misses grow far longer than existing windows.
The commands after a coordinate stay the same whatever value it takes, which sets vector graphics apart from natural language.
A run-length identity and three propositions make this precise.
With enough candidates per position, the gain from a longer window lasts until the drafter's horizon, how deep into a block the drafter stays accurate.

We turn this finding into Sightline, a speculative decoder whose window reaches as far as its drafter can see.
It drafts with a head trained for twice the window of existing block drafters, and verifies a draft tree with enough candidates to fill the whole window.
A loop control for every decoder escapes a repeated cycle once and stops the output if it repeats.

\begin{figure*}[t]
\centering
\includegraphics{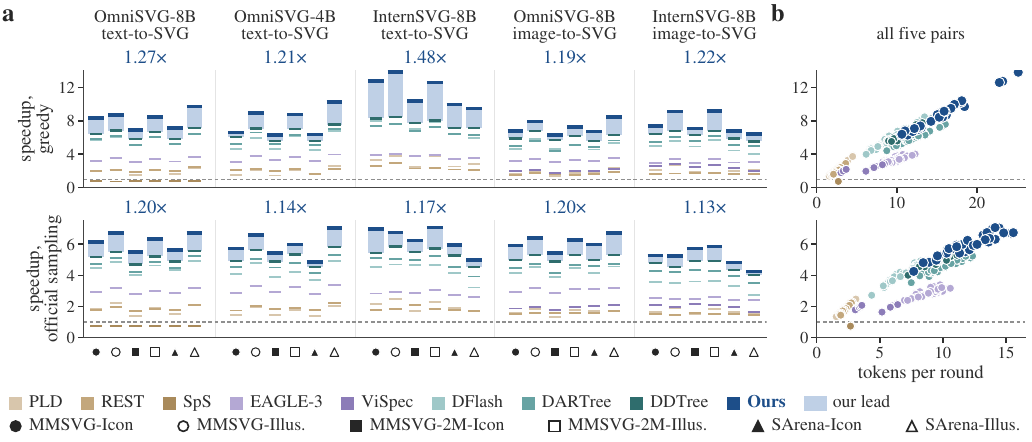}
\caption{\textbf{Lossless speedup and tokens per round in all 60 settings.} (a)~One tick per decoder and benchmark. The number above each model--task pair is our mean speedup over its six benchmarks divided by that of the fastest other decoder. (b)~Speedup against tokens per round in the settings of a. Speedups are over the original model timed in the same job, and count all tokens (Appendix~\ref{app:speed-all}).}
\label{fig:main}
\end{figure*}

On OmniSVG-8B, OmniSVG-4B and InternSVG-8B, Sightline is faster than every compared lossless decoder in every setting, and 4.25$\times$ to 13.83$\times$ faster than the original model (Figures~\ref{fig:teaser} and~\ref{fig:main}).
Per model--task pair, its mean speedup is 1.13$\times$ to 1.48$\times$ that of the fastest other decoder.
When every decoder and the original model share one loop control, it stays the fastest and runs 4.65$\times$ to 7.58$\times$ faster than the original model (Table~\ref{tab:quality}).
The time that speculative decoding saves also buys better drawings.
With loop control, a best of eight chosen by rendering takes less time on average than one official sample of the original model.

Our main contributions are summarized as follows.
\begin{itemize}
\item We identify why speculative decoding on SVG stops at the verification window, and trace it to rare and clumped misses.
\item We explain with a run-length identity and three propositions when a longer window pays, where its gain ends and how loops change it, each with predictions registered before the experiments.
\item We propose Sightline, whose head is trained up to the drafter's horizon and whose draft tree fills the longer window.
\item We evaluate Sightline against eight existing decoders on three SVG generation models, two tasks and six benchmarks, and on six more models from four families.
\item We locate its lead in the rounds that existing windows truncate, and show that rendering judges SVG samples better than likelihood does.
\end{itemize}

\section{Related Work}
\label{sec:related}

Speculative decoders differ in where their drafts come from~\cite{xia-etal-2024-unlocking}.
SpS drafts with a smaller model of the same family~\cite{3618408.3619203,chen2023accelerating}, PLD and REST copy from the context or a datastore~\cite{saxena2023prompt,he-etal-2024-rest}, and EAGLE-3 drafts one token at a time from the target's hidden states~\cite{3692070.3693232,li2025eagle}.
Block drafters write a whole block per pass~\cite{3692070.3692273,arriola2025block}, as DFlash does~\cite{chen2026dflash}, and draft trees raise the tokens per round~\cite{10.1145/3620666.3651335,li-etal-2024-eagle}.
DDTree builds such trees for block drafters, and DARTree corrects their branches with an autoregressive head~\cite{ringel2026accelerating,li2026dartree}, while theory covers several drafts and speedup limits~\cite{sun2023spectr,pankratov-alistarh-2026-speculative}.
Recent block drafters enlarge the drafter, condition on earlier tokens, draft twice, adapt the verification length or post-train for a longer block~\cite{zhang2026dflare,huang-etal-2026-domino,hu2026jetspec,zhang2026d,cheng2026dspark,oh2026bastion,lin2026libraspec,wu2026ceiling}, all on natural language, math and code.
Adaptive verification lengths trim the window where acceptance is unlikely, while Sightline stretches the window of a wide tree to the drafter's horizon on vector graphics.

For images, speculative decoding drafts by Jacobi iterations or relaxes the verification~\cite{teng2025accelerating,jang2025lantern,11444739,dong2026vvs}, drafters for vision language models such as ViSpec compress, skip or block-draft the image~\cite{kang2025vispec,Xie_2026_CVPR,lee2026vision}, and meshes and tables have structure-aware drafters~\cite{shen2026flashmesh,Zaytsev_2026_CVPR}.
SVG generation has moved from optimization through a differentiable rasterizer~\cite{Li:2020:DVG,10204913,10658565} and sequence models of drawing commands~\cite{3495724.3497096,10.1145/3618364} to models that write the code~\cite{Rodriguez_2025_CVPR,Xing_2025_CVPR,11092564,yang2025omnisvg,wang2026internsvg}, and compact tokenizations need retrained models~\cite{yang2026omnilottie,xing2026hierarchical}.
Repetition grows under likelihood-seeking decoding~\cite{Holtzman2020The} and is treated in training or in the attention cache~\cite{Welleck2020Neural,3600270.3600493,xu-etal-2026-loopguard}, and choosing among samples raises quality~\cite{snell2025scaling,Ma_2025_CVPR,wu2026canvas} and can stop weak candidates early~\cite{sun2024fast}.
Sightline keeps the target's tokens, verifies every drafted token exactly and leaves time for the selection.

\begin{figure}[t]
\centering
\includegraphics{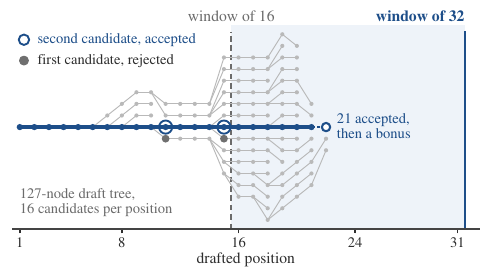}
\caption{\textbf{One round of Sightline.} OmniSVG-8B text-to-SVG, first round on an MMSVG prompt. The accepted path takes the second candidate at positions 11 and 15, where a chain of first candidates stops. It then runs past the end of a window of 16 (Appendix~\ref{app:detail}).}
\label{fig:method}
\end{figure}
\begin{figure*}[t]
\centering
\includegraphics{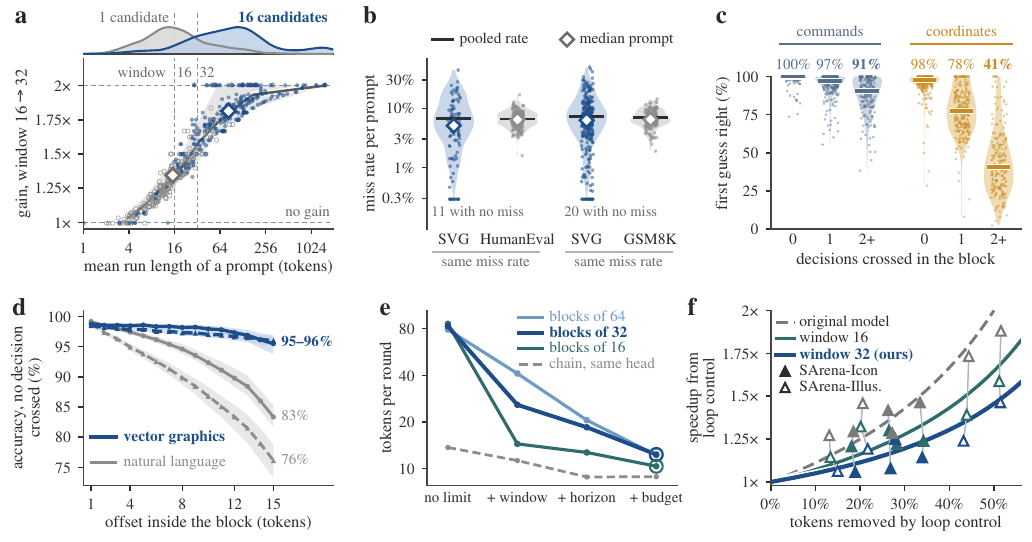}
\caption{\textbf{Propositions 1 to 3 measured.} (a)~One point per prompt, the median prompt a diamond, and a band over the middle 80\%. (b)~One candidate per position. (c)~One point per prompt at positions that are not decisions, with bars at the median. (d)~95\% bootstrap intervals over prompts, solid and dashed for the two pairs of b. (e)~Each step adds one limit, and rings are measured. (f)~One point per SArena setting and decoder, and curves from~\eqref{eq:loopgain}. Vector graphics in a to e are loop-free greedy outputs of OmniSVG-8B text-to-SVG (Appendix~\ref{app:theory-more}).}
\label{fig:theory}
\end{figure*}

\begin{table*}[t]
\caption{\textbf{Speed and quality on the full SArena test sets.} Seconds per prompt and two official metrics on 6,013 icon and 2,001 illustration prompts, under OmniSVG-8B's released sampling and InternSVG-8B's greedy decoding. Each group holds the original model (AR) and the four fastest baselines of Figure~\ref{fig:main}, lossless and with the same loop control. \textbf{Fastest} in bold, \underline{second}{} underlined, by seconds within a group.}
\label{tab:quality}
\centering
\small
\setlength{\tabcolsep}{2.2pt}
\renewcommand{\arraystretch}{0.9}
\begin{tabular*}{\textwidth}{@{\extracolsep{\fill}}ll*{12}{c}@{}}
\toprule
 &  & \multicolumn{6}{c}{text-to-SVG} & \multicolumn{6}{c}{image-to-SVG} \\
\cmidrule(lr){3-8}\cmidrule(lr){9-14}
 &  & \multicolumn{3}{c}{SArena-Icon} & \multicolumn{3}{c}{SArena-Illus.} & \multicolumn{3}{c}{SArena-Icon} & \multicolumn{3}{c}{SArena-Illus.} \\
\cmidrule(lr){3-5}\cmidrule(lr){6-8}\cmidrule(lr){9-11}\cmidrule(lr){12-14}
 & Method & s$\downarrow$ & FID$\downarrow$ & CLIP$\uparrow$ & s$\downarrow$ & FID$\downarrow$ & CLIP$\uparrow$ & s$\downarrow$ & DINO$\uparrow$ & LPIPS$\downarrow$ & s$\downarrow$ & DINO$\uparrow$ & LPIPS$\downarrow$ \\
\midrule
\multicolumn{14}{l}{\textbf{OmniSVG-8B}} \\
\multirow{6}{1.05cm}{\centering\itshape lossless} & AR & 5.72 & 10.81 & 22.76 & 15.92 & 38.93 & 17.40 & 8.85 & 0.903 & 0.178 & 19.82 & 0.803 & 0.272 \\
 & EAGLE-3 & 1.91 & 10.93 & 22.79 & 4.76 & 39.09 & 17.31 & 2.96 & 0.902 & 0.178 & 6.14 & 0.805 & 0.270 \\
 & DFlash & 1.29 & 10.76 & 22.80 & 3.08 & 39.30 & 17.39 & 2.11 & 0.901 & 0.179 & 4.24 & 0.803 & 0.270 \\
 & DDTree & \underline{1.14} & 10.95 & 22.79 & \underline{2.79} & 39.19 & 17.33 & \underline{1.76} & 0.902 & 0.178 & \underline{3.68} & 0.804 & 0.271 \\
 & DARTree & 1.27 & 10.89 & 22.79 & 3.06 & 39.16 & 17.36 & 1.96 & 0.902 & 0.177 & 4.08 & 0.803 & 0.272 \\
 & \textbf{Ours} & \textbf{1.01} & 10.86 & 22.77 & \textbf{2.28} & 39.24 & 17.32 & \textbf{1.48} & 0.902 & 0.178 & \textbf{2.94} & 0.805 & 0.272 \\
\cmidrule(lr){1-14}
\multirow{6}{1.05cm}{\centering\itshape\shortstack{loop\\control}} & AR & 4.46 & 10.89 & 22.76 & 8.46 & 39.07 & 17.37 & 6.05 & 0.903 & 0.177 & 11.43 & 0.805 & 0.272 \\
 & EAGLE-3 & 1.54 & 10.99 & 22.81 & 2.81 & 39.22 & 17.33 & 2.33 & 0.903 & 0.177 & 4.26 & 0.807 & 0.270 \\
 & DFlash & 1.08 & 10.87 & 22.80 & 2.05 & 39.34 & 17.38 & 1.77 & 0.902 & 0.178 & 3.25 & 0.805 & 0.270 \\
 & DDTree & \underline{0.94} & 10.99 & 22.79 & \underline{1.76} & 39.27 & 17.33 & \underline{1.42} & 0.903 & 0.177 & \underline{2.64} & 0.806 & 0.273 \\
 & DARTree & 1.08 & 11.02 & 22.79 & 1.96 & 38.99 & 17.35 & 1.57 & 0.903 & 0.177 & 2.91 & 0.805 & 0.272 \\
 & \textbf{Ours} & \textbf{0.80} & 10.94 & 22.76 & \textbf{1.56} & 39.18 & 17.33 & \textbf{1.30} & 0.903 & 0.177 & \textbf{2.37} & 0.806 & 0.272 \\
\midrule
\multicolumn{14}{l}{\textbf{InternSVG-8B}} \\
\multirow{6}{1.05cm}{\centering\itshape lossless} & AR & 11.21 & 11.83 & 23.71 & 43.58 & 52.91 & 17.95 & 10.98 & 0.965 & 0.077 & 43.93 & 0.864 & 0.204 \\
 & EAGLE-3 & 3.28 & 11.75 & 23.76 & 12.62 & 52.84 & 17.92 & 3.66 & 0.963 & 0.078 & 14.65 & 0.869 & 0.202 \\
 & DFlash & 1.81 & 11.88 & 23.74 & 7.31 & 54.32 & 17.90 & 2.37 & 0.965 & 0.077 & 10.95 & 0.870 & 0.201 \\
 & DDTree & \underline{1.60} & 11.94 & 23.71 & \underline{6.18} & 53.83 & 17.89 & \underline{1.88} & 0.963 & 0.078 & \underline{7.86} & 0.864 & 0.204 \\
 & DARTree & 1.77 & 11.89 & 23.74 & 6.81 & 54.62 & 17.88 & 2.03 & 0.965 & 0.077 & 8.81 & 0.868 & 0.202 \\
 & \textbf{Ours} & \textbf{1.14} & 11.84 & 23.73 & \textbf{4.70} & 54.37 & 17.88 & \textbf{1.54} & 0.963 & 0.078 & \textbf{6.96} & 0.866 & 0.203 \\
\cmidrule(lr){1-14}
\multirow{6}{1.05cm}{\centering\itshape\shortstack{loop\\control}} & AR & 7.91 & 12.22 & 23.69 & 29.88 & 53.42 & 17.96 & 8.49 & 0.965 & 0.077 & 34.55 & 0.866 & 0.203 \\
 & EAGLE-3 & 2.48 & 12.09 & 23.72 & 9.33 & 54.18 & 17.95 & 3.01 & 0.964 & 0.077 & 12.54 & 0.868 & 0.204 \\
 & DFlash & 1.47 & 12.07 & 23.74 & 5.81 & 54.93 & 17.86 & 2.13 & 0.965 & 0.077 & 10.12 & 0.871 & 0.202 \\
 & DDTree & \underline{1.23} & 12.38 & 23.69 & \underline{4.67} & 54.15 & 17.87 & \underline{1.56} & 0.964 & 0.078 & \underline{6.86} & 0.866 & 0.204 \\
 & DARTree & 1.38 & 12.22 & 23.74 & 5.14 & 54.87 & 17.88 & 1.78 & 0.965 & 0.077 & 7.98 & 0.870 & 0.202 \\
 & \textbf{Ours} & \textbf{1.05} & 12.10 & 23.72 & \textbf{3.94} & 54.93 & 17.87 & \textbf{1.46} & 0.963 & 0.079 & \textbf{6.54} & 0.866 & 0.204 \\
\bottomrule
\end{tabular*}
\end{table*}

\begin{figure*}[t]
\centering
\includegraphics{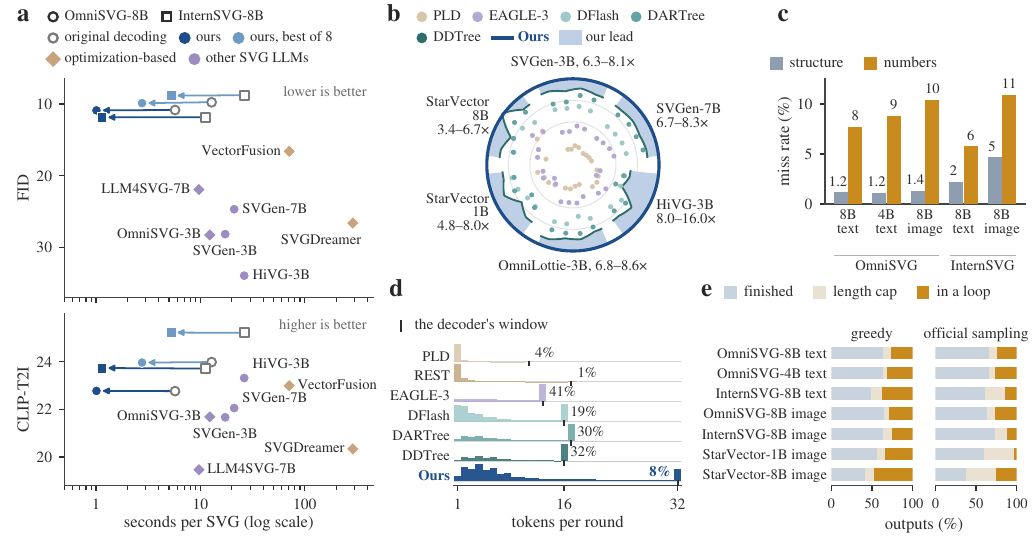}
\caption{\textbf{Comparison with SVG generators (a), other model families (b), and analysis (c--e).} (a)~SArena-Icon text-to-SVG, with arrows from each model's own decoding to ours. Diamonds are reported by InternSVG. We measure SVGen~\cite{10.1145/3746027.3755011}, HiVG and our two targets on the same GPU (Appendix~\ref{app:generators}). (b)~Radius is a decoder's speedup as a share of ours, four spokes per model for two subsets and two decodings. (c)~Miss rate of our draft head. (d)~Tokens per round, each ridge scaled to its peak, with the share of rounds that stop at the decoder's window. (e)~Endings of the outputs. Panels c and d are greedy and count loop-free content (Appendices~\ref{app:full} and~\ref{app:lead}).}
\label{fig:ex1}
\end{figure*}
\begin{figure*}[t]
\centering
\includegraphics{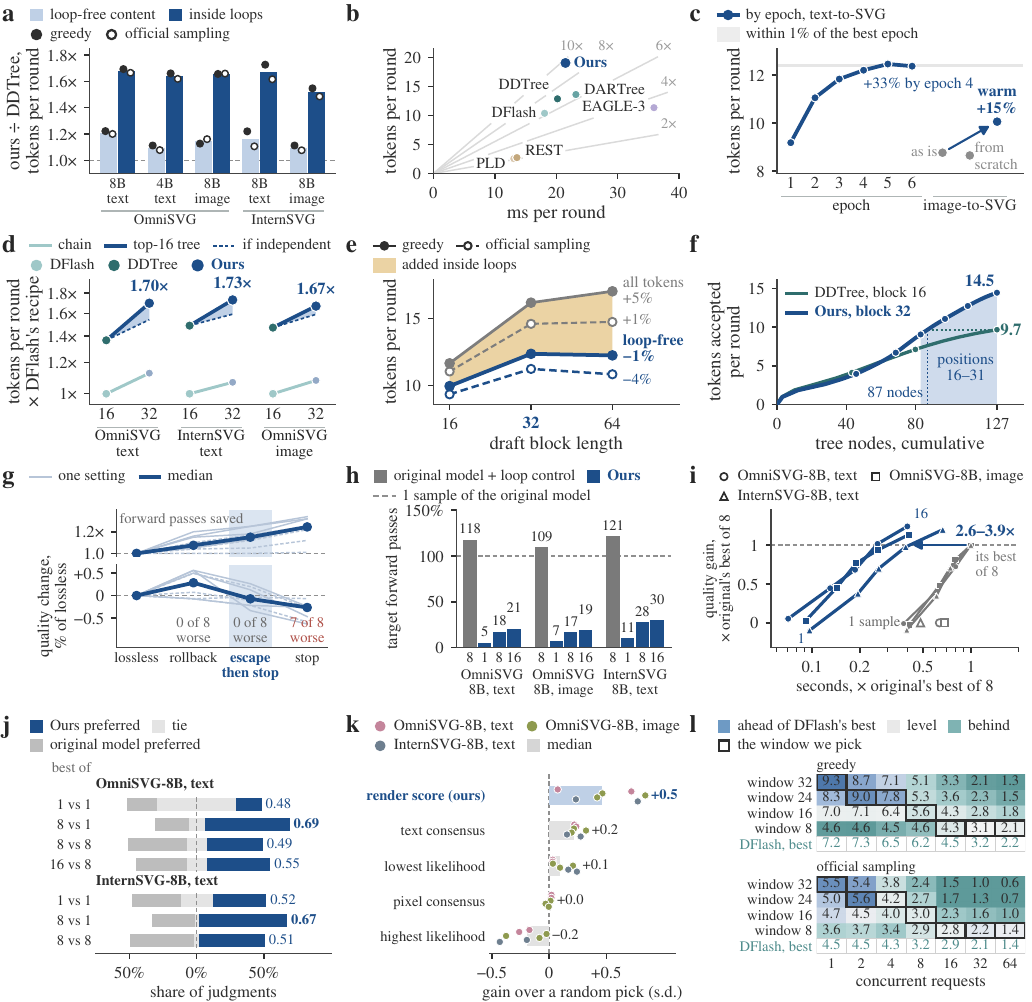}
\caption{\textbf{Sources of the lead (a, b), training and design (c--f), loop control (g), the best of eight (h--k) and serving (l).} (a)~Our tokens per round over DDTree's, averaging the two decodings. (b)~Greedy, rays of equal speedup $S$. (c)~Training epochs and three starts of the image-to-SVG head. (d)~Gains of the two changes over DFlash's recipe, their product dashed. (f)~Tokens accepted as the tree grows. (g)~Change of the main metric from the lossless run. (h--j)~300 prompts per pair, original model in gray. (i)~Relative to the original model's best of 8, hollow for one official sample. (j)~Three VLM judges, 1,800 judgments per row, ties counted half. (k)~Gain over a random pick in standard deviations of the per-prompt difference. (l)~InternSVG-8B text-to-SVG in SGLang with chain verification, colored against DFlash's head at its best window. Panels b, c, e and f use OmniSVG-8B text-to-SVG, and c, d and f count loop-free content (Appendices~\ref{app:lead} to~\ref{app:serving}).}
\label{fig:ex2}
\end{figure*}

\section{Method}
\label{sec:method}

\subsection{Preliminaries}
\label{sec:prelim}

We write $x_{\le t}$ for the prompt and the tokens so far, and $p(\cdot\mid x_{\le t})$ for the target's next-token distribution under its decoding settings.
In one round, a drafter offers $k$ candidates for each of the next $W-1$ positions, and the draft tree $\mathcal{T}$ holds at most $B$ continuations of them, closed under prefixes and a chain when $k=1$.
The target scores $\mathcal{T}$ in one forward pass and walks it by its decoding rule, drawing $y_d$ from $p(\cdot\mid x_{\le t},y_{<d})$ at depth $d$ and moving on while $(y_1,\dots,y_d)\in\mathcal{T}$.
The round accepts $A=\max\{d:(y_1,\dots,y_d)\in\mathcal{T}\}$ drafted tokens and adds the target's next draw as a bonus, so it writes $A+1\le W$ tokens.
It stops after all $W-1$ drafted positions or at a miss, a position where the target's draw is not a candidate.
Every written token is a draw of the target, and in exact arithmetic the output therefore follows $p$.
Greedy decoding then reproduces the original model.

We call $W$ the window and the mean of $A+1$ over the rounds the tokens per round $\tau$.
Summing $\Pr(A\ge d)$ over the depths the walk reaches gives $\mathbb{E}[A\mid x_{\le t}]=\sum_{u\in\mathcal{T}}p(u\mid x_{\le t})$, where $p(u\mid x_{\le t})$ is the target's probability of $u$.
If a round takes $T_{\mathrm{draft}}$ to draft and $T_{\mathrm{verify}}$ to verify, and the original model takes $T_{\mathrm{AR}}$ per token, the speedup is $S=\tau\,T_{\mathrm{AR}}/(T_{\mathrm{draft}}+T_{\mathrm{verify}})$~\cite{sadhukhan2025magicdec}.
A block drafter writes all $W-1$ positions in one pass, so a longer window adds little to $T_{\mathrm{draft}}$ (Appendix~\ref{app:round-cost}).

\subsection{Sightline}
\label{sec:sightline}

Since a block drafter misses rarely on SVG and sees far within a block, the window sets how much a round collects.
Sightline lengthens the window to the drafter's horizon.
Its draft tree has DDTree's node budget and the fewest candidates per position at which $\tau$ levels off (Appendix~\ref{app:tree-shape}).

Its draft head reads the target's hidden states and the last written token, and returns in one pass a distribution $q_j$ for each position $t+j$ (Figure~\ref{fig:method}).
Each $q_j$ is a distribution for position $t+j$ alone, because the head sees none of the tokens it drafts.
A head drafts only as far as its training block.
Ours follows the design and training of DFlash at a block length of 32 (Appendix~\ref{app:training}).
The same head also serves every shorter window, which verifies a prefix of its block (Appendix~\ref{app:runs}).

Position $t+j$ offers the $k$ most likely tokens of $q_j$.
As in DDTree, the tree keeps the $B$ continuations of highest drafter probability $\prod_j q_j(u_j)$.
The kept set maximizes the drafter's estimate of $\mathbb{E}[A\mid x_{\le t}]$ and is closed under prefixes.
Appendix~\ref{app:detail} gives the round as an algorithm.

Lossless decoding also keeps the loops of the target, where the output repeats one cycle up to the length limit.
Loop control fires when the output ends with three exact copies of a cycle of 8 to 128 tokens.
At the first $m$ firings, it bans the token $c_t$ that would start the next copy, and at the next firing, it stops the output, which replaces the target by
\begin{equation}
  \tilde p_m(\cdot\mid x_{\le t})=p(\cdot\mid x_{\le t};\mathcal{V}_t),
  \label{eq:loop}
\end{equation}
its decoding rule on the vocabulary $\mathcal{V}_t$, which is $\mathcal{V}\setminus\{c_t\}$ at the first $m$ firings, $\{\mathrm{eos}\}$ at later ones and $\mathcal{V}$ otherwise.
Since $\tilde p_m$ depends only on the written tokens, every decoder stays exact with respect to it.

Speculative decoding also leaves time to choose among samples.
For a best of $N$, we draw $N$ samples of $\tilde p_0$ in one batch, each verifying a chain.
Each sample stops at its first loop, since a batch runs until its longest sample ends.
We keep the sample whose rendering scores highest, by CLIP score~\cite{shen2022klite} against the prompt or DINOv2 similarity~\cite{oquab2024dinov} to the input image.
Rendered selection is established for SVG~\cite{Rodriguez_2025_CVPR}, and fast decoding makes it affordable.

\subsection{Theoretical Analysis}
\label{sec:theory}

The rounds spent on a given output can be counted exactly~\cite{zhao2026training,3618408.3619203}.
We registered a prediction for each statement below before the experiments, and Appendix~\ref{app:predictions} lists each with its outcome.
Appendix~\ref{app:proofs} gives the proofs.

\begin{lemma}[Run-length identity]
\label{lem:runlength}
Fix a drafter and a drafting offset.
Call a position of the output a miss when the token written there is not in the drafter's candidate set for it, and cut the output after every miss into runs of lengths $n_1,\dots,n_P$, where the last run may be unclosed.
If the labels do not depend on where a round starts, a decoder with window $W$ spends $\lceil n_i/W\rceil$ rounds on run $i$ and writes
\begin{equation}
  \tau_W=\frac{\sum_{i=1}^{P}n_i}{\sum_{i=1}^{P}\lceil n_i/W\rceil}
  \label{eq:runlength}
\end{equation}
tokens per round.
\end{lemma}

A window acts only inside runs longer than itself.
As $W$ grows, \eqref{eq:runlength} tends to the ceiling $\tau_\infty=\sum_i n_i/P$, set by the miss rate alone.
How much of the ceiling a window collects is left to the run lengths, and a model that keeps the accuracy of every position but draws misses independently reaches about four fifths of the measured $\tau_{32}$.
The identity also prices a head's shorter windows from its full window's labels, to a median error of 2.7\% (Appendix~\ref{app:runs}).
For independent misses at rate $\lambda$, \eqref{eq:runlength} gives $\tau_W=(1-(1-\lambda)^W)/\lambda$ on long outputs, and its reciprocal $r_W(\lambda)$ gives the rounds per token.

\begin{proposition}[Window share and clustering tax]
\label{prop:share}
Widen the candidate set at every position, and write $\tau'$ for the tokens per round under the labels of Lemma~\ref{lem:runlength} with the wider sets.
Then $\tau'_\infty\ge\tau_\infty$ and $\tau'_W/\tau'_\infty\le\tau_W/\tau_\infty$ at every window $W$.
Suppose further that the output splits into parts that are long compared with $W$ and take shares $w_i$ of it, and that misses in part $i$ are independent at rate $\lambda_i$, with mean rate $\bar\lambda=\sum_i w_i\lambda_i$.
As the parts grow, the rounds per token satisfy
\begin{equation}
  \frac{1}{\tau_W}\longrightarrow\sum_i w_i\,r_W(\lambda_i)\ \ge\ r_W(\bar\lambda),
  \label{eq:tax}
\end{equation}
with equality for $W\ge 2$ only when all $\lambda_i$ are equal.
\end{proposition}

Widening turns some misses into correct positions and joins the runs on either side, so the ceiling rises while the joined runs take more rounds to cross.
Clumping gathers the same misses into fewer places, and windows shorter than the long runs it leaves pay this tax.
Once the candidates are wide, lengthening the window pays until it reaches the drafter's horizon, the distance within a block over which the drafter stays accurate.
Chains and trees share one curve of the gain from doubling the window, and the tree moves the median prompt up it (Figure~\ref{fig:theory}a).
At matched miss rates, vector graphics prompts range from no miss to frequent misses, while natural language prompts cluster (Figure~\ref{fig:theory}b).

A drafter that writes a block in one pass sees the prefix of the block but not the tokens inside it~\cite{ankner2024hydra,qiang2026beyond}.

\begin{proposition}[Horizon in decisions]
\label{prop:horizon}
Let a round start after $x_{\le t}$, let $b$ range over the ways the positions $t+1,\dots,t+j-1$ can be resolved, and write $p_b=p(\cdot\mid x_{\le t},b)$ for the target's distribution at $t+j$ given $b$.
A set $C_j$ of $k$ candidates chosen from $x_{\le t}$ alone contains the target's token at $t+j$ with probability $\bar p(C_j)$, where $\bar p=\sum_b p(b\mid x_{\le t})\,p_b$ is the target's marginal there, and
\begin{equation}
  \bar p(C_j)\le\max_{|S|=k}\bar p(S)\le\sum_b p(b\mid x_{\le t})\max_{|S|=k}p_b(S).
  \label{eq:horizon}
\end{equation}
The right side is the probability for a drafter that also sees $b$.
The last two terms of \eqref{eq:horizon} are equal exactly when one set of $k$ tokens is among the $k$ most likely of every $p_b$ with $p(b\mid x_{\le t})>0$.
\end{proposition}

We call a position a decision when the target's own next-token entropy there exceeds $\eta=0.2$ nats, and thresholds of 0.05 and 0.5 nats give the same picture (Appendix~\ref{app:decisions}).
A decision crossed within the block costs only the positions that depend on it.
In SVG, the commands do not depend on the coordinate decisions they cross, so a crossed decision costs the coordinates and spares the commands (Figure~\ref{fig:theory}c and Appendix~\ref{app:robust}).
Classes of natural language with a similar share of the stream keep less than half of their accuracy past two decisions (Appendix~\ref{app:decisions}).
Between decisions, the drafter also stays accurate farther into the block on vector graphics than on natural language (Figure~\ref{fig:theory}d).
Its accuracy still falls with distance, and this fall and the node budget end the gain from a longer window (Figure~\ref{fig:theory}e).

\begin{proposition}[Loop limit]
\label{prop:loop}
Suppose loop control removes a share $\varphi$ of the output and leaves the rest unchanged, and the drafter does not miss inside the removed loops.
A decoder with window $W$ writes $W$ tokens per round inside such a loop, and loop control divides its number of rounds by
\begin{equation}
  \gamma_W=1+\frac{\varphi\,\tilde\tau_W}{(1-\varphi)\,W},
  \label{eq:loopgain}
\end{equation}
where $\tilde\tau_W$ is its tokens per round on the rest.
Hence $\gamma_W\le\gamma_1=1/(1-\varphi)$, and loop control multiplies the lead of a window $W'$ over a window $W$ by $\gamma_{W'}/\gamma_W$.
This factor is below one when $\tilde\tau_{W'}/W'<\tilde\tau_W/W$.
\end{proposition}

A loop is a run with no end, so a window collects it at $W$ tokens per round.
Loop control therefore belongs to every decoder in a comparison, and it narrows the lead of a longer window without closing it (Figure~\ref{fig:theory}f and Table~\ref{tab:quality}).
With the tokens per round measured inside loops in place of $W$, \eqref{eq:loopgain} predicts the gains to a median error of 6\% (Appendix~\ref{app:loops-theory}).

\section{Experiments}
\label{sec:exp}

\subsection{Experimental Setup}
\label{sec:setup}

\noindent\textbf{Models, benchmarks and baselines.}
We accelerate five model--task pairs, OmniSVG-8B and InternSVG-8B on text-to-SVG and image-to-SVG, and OmniSVG-4B on text-to-SVG, on the icon and illustration sets of MMSVG and MMSVG-2M~\cite{yang2025omnisvg} and of SArena~\cite{wang2026internsvg} (Appendix~\ref{app:setup}).
Every draft head trains on its target's outputs for MMSVG-2M prompts outside the two MMSVG-2M benchmarks, and ours follows the training settings of DFlash (Appendix~\ref{app:training}).
We compare with eight existing decoders, PLD~\cite{saxena2023prompt}, REST~\cite{he-etal-2024-rest}, SpS~\cite{chen2023accelerating}, EAGLE-3~\cite{li2025eagle}, ViSpec~\cite{kang2025vispec}, DFlash~\cite{chen2026dflash}, DDTree~\cite{ringel2026accelerating} and DARTree~\cite{li2026dartree}, all verified by the rule of Section~\ref{sec:prelim} (Appendix~\ref{app:baselines}).

\noindent\textbf{Protocol and metrics.}
Each pair runs its six benchmarks under greedy decoding and official sampling, 60 settings in all.
InternSVG-8B releases greedy decoding only, so we sample it at temperature 0.6 and top-$p$ 0.9.
Decoders are compared losslessly, with one shared loop control ($m=1$), or as the best of $N=8$.
Lossless decoders write the original model's tokens in exact arithmetic, and fp32 audits of 1.05 million tokens found no mismatch.
In bf16, passes of other shapes can break near-ties differently, so every decoder parts from about the same share of the original model's outputs (Appendix~\ref{app:lossless}).
The share is as large for PLD and REST, which train no drafter.
Speed is the tokens per round $\tau$ and the wall-clock speedup over the original model timed in the same job on one GPU.
Quality uses the official metrics of SArena, with FID~\cite{3295222.3295408} and CLIP score for text-to-SVG, and DINOv2 similarity and LPIPS~\cite{severo2024the} for image-to-SVG.
The main results count every token, and the analyses count loop-free content, the rounds that start before the output enters a loop.
Reruns at three seeds and three temperatures keep the order of the methods, and the lead over DDTree narrows as the temperature rises (Appendix~\ref{app:variation}).

\subsection{Main Results}
\label{sec:main}\label{sec:general}
\noindent\textbf{Sightline is the fastest decoder in every setting (Figure~\ref{fig:main} and Table~\ref{tab:quality}).}
Its lead over the fastest other decoder holds on every model--task pair under both decodings (Figure~\ref{fig:main}a).
The speedup grows with the tokens per round, and Sightline writes the most per round in every setting (Figure~\ref{fig:main}b).
On the full SArena test sets, it takes the fewest seconds in every column of both groups, and quality within each group stays close (Table~\ref{tab:quality}).
Among SVG generators, Sightline brings its target to the fast end without losing quality (Figure~\ref{fig:ex1}a).

\noindent\textbf{The lead holds beyond our three models (Figure~\ref{fig:ex1}b).}
We train every method's draft head with one shared budget on six more models, which come from four families and emit SVG code, compressed SVG tokens and Lottie JSON (Appendix~\ref{app:training}).
Sightline is the fastest in almost every setting.

\subsection{Analysis}
\label{sec:analysis}
\noindent\textbf{Misses fall on numbers, and the structure between them stays predictable (Figure~\ref{fig:ex1}c).}
Numbers carry the bulk of our head's misses in every pair, and Proposition~\ref{prop:horizon} explains why the commands after them stay correct.

\noindent\textbf{Existing windows cut SVG rounds short (Figure~\ref{fig:ex1}d and e).}
The rounds of existing block drafters pile up at their windows, while ours run past them (Figure~\ref{fig:ex1}d).
Trees widen the candidates and so fill their windows more often than chains, as Proposition~\ref{prop:share} predicts.
Loops make such rounds more common still, and many outputs end in one, more often under greedy decoding (Figure~\ref{fig:ex1}e).

\noindent\textbf{The lead comes from rounds that a longer window lets continue (Figure~\ref{fig:ex2}a and b).}
Sightline writes 1.1 to 1.2 times DDTree's tokens per round on loop-free content and 1.5 to 1.7 times in loops, where a shorter window caps every round.
Its rounds cost about as much, since the target's pass dominates and the head drafts every position at once.

\noindent\textbf{The lead does not come from more training (Figure~\ref{fig:ex2}c and e).}
With the training settings of DFlash, ours levels off after four epochs.
A head trained on blocks of 16 for twice as many epochs as ours still writes a fifth fewer tokens per round on loop-free content.
Like existing image heads, ours starts from the text head, which beats a cold start.

\noindent\textbf{Window and tree are complements, as Proposition~\ref{prop:share} predicts (Figure~\ref{fig:ex2}d and f).}
Added to DFlash's recipe, the tree alone gives DDTree, and the longer window alone adds little.
Together they gain more in every pair than their separate gains multiplied.
With the same 127 nodes, our tree accepts as much as DDTree with about two thirds of them, and spends the rest beyond the reach of shorter windows.

\noindent\textbf{The useful window ends at the drafter's horizon, and a longer training block does not extend it (Figures~\ref{fig:ex2}e and~\ref{fig:theory}e).}
From block length 32 to 64, all added tokens come from loops, and loop-free content loses a little.

\noindent\textbf{Loop control narrows the lead without closing it, as Proposition~\ref{prop:loop} predicts (Table~\ref{tab:quality} and Figure~\ref{fig:ex2}g).}
The same rule~\eqref{eq:loop} acts on every decoder and on the original model, which it speeds up most.
With it, Sightline still runs 1.05$\times$ to 1.19$\times$ faster than DDTree in Table~\ref{tab:quality}.
An offline check finds most of the outputs that loop control stops already in a loop (Appendix~\ref{app:loop-stopped}).
Escaping once before stopping keeps quality within noise in every setting, while stopping at once lowers it almost everywhere (Figure~\ref{fig:ex2}g).

\begin{figure}[t]
\centering
\includegraphics{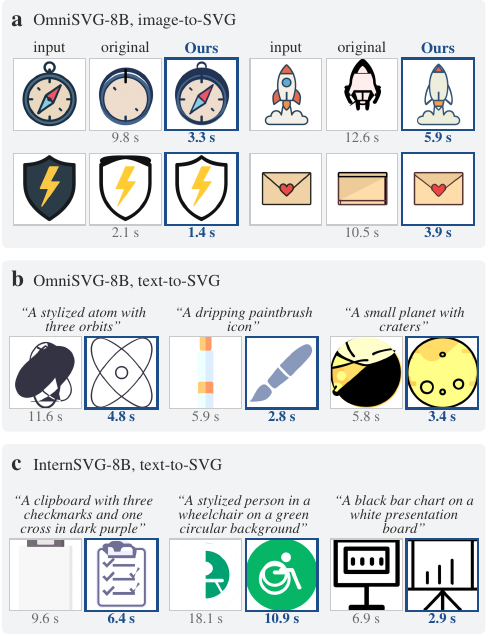}
\caption{\textbf{Qualitative comparison of one original sample and our best of eight.} One sample of the original model against our best of eight (blue), with the seconds per prompt on one GPU. Both use the same loop control. The prompts follow a fixed rule (Appendix~\ref{app:examples}). On them, our best of eight and the original model's both beat one sample, and ours finishes first.}
\label{fig:cases}
\end{figure}
\noindent\textbf{Speculative decoding leaves time for better drawings (Figures~\ref{fig:ex2}h to j and~\ref{fig:cases}).}
With the same loop control and rendered selection, our best of eight matches the original model's best of eight in quality in a fraction of its time (Figure~\ref{fig:ex2}i).
It needs under a third of the target passes of one original sample (Figure~\ref{fig:ex2}h).
Counting rendering and scoring, it also finishes on average before one official sample (Figure~\ref{fig:ex2}i).
Three VLM judges prefer it two to one over one original sample, and rate it even with the original model's best of eight (Figure~\ref{fig:ex2}j).
Rendered selection also improves FID and LPIPS, which it ignores.
In a batch, each sample verifies a chain, so DFlash's chain head is about as fast (Appendix~\ref{app:quality-full}).
Figure~\ref{fig:cases} shows ten examples.

\noindent\textbf{Likelihood is the wrong judge of a drawing (Figure~\ref{fig:ex2}k).}
On metrics the render score does not use, rendering beats four other ways to choose among eight samples.
The likeliest sample does worse than a random pick, as it often loops.

\noindent\textbf{One head adapts to the load by shortening its window (Figure~\ref{fig:ex2}l).}
SGLang~\cite{zheng2024sglang} verifies chains.
With 1 to 64 concurrent requests, each head runs at the window that a sweep finds fastest for each load.
At low load, ours verifies a long window and runs up to 1.29$\times$ faster than DFlash's head.
As load grows, both heads verify fewer positions, as in DSpark~\cite{cheng2026dspark}.
Ours then runs at 0.91$\times$ to 1.03$\times$ the speed of DFlash's head (Appendix~\ref{app:serving-load}).

\section{Conclusion}
\label{sec:conclusion}
Looking at where block drafters miss on vector graphics, we showed that the misses are few and clumped, and that the commands following a coordinate keep their accuracy.
A run-length identity and three propositions explain why a longer window pays only with wide candidates, and why its gain ends at the drafter's horizon.
Sightline pairs the two and outpaces every compared decoder in every setting.
The long window pays more when requests are few and the target is confident.
On vector graphics, how far each verification looks may matter as much as how well the drafter guesses.

\section*{LLM Usage Statement}

We used Claude, GPT, Gemini, and DeepSeek to assist with literature searches, running experimental code, preparing figures and tables, and editing the manuscript. In the experiments, InternSVG-8B reads each input image with a one-sentence description, and a DeepSeek model wrote these descriptions for MMSVG, which ships none (Appendix~\ref{app:setup}). The authors directed the work and take full responsibility for the paper.

{
    \small
    \bibliographystyle{ieeenat_fullname}
    \bibliography{ref}
}

\clearpage
\onecolumn
{\centering\Large\textbf{Appendix}\par\vspace{1.0em}}
\renewcommand{\arraystretch}{0.9}%
\setcounter{topnumber}{3}\setcounter{bottomnumber}{2}\setcounter{totalnumber}{6}
\renewcommand{\topfraction}{0.9}\renewcommand{\bottomfraction}{0.6}\renewcommand{\textfraction}{0.07}
\renewcommand{\floatpagefraction}{0.7}
\makeatletter\setlength{\@fptop}{0pt}\setlength{\@fpbot}{0pt plus 1fil}
\let\appx@xfloat\@xfloat\def\@xfloat#1[#2]{\appx@xfloat#1[htbp]}%
\makeatother
\appendix
\setcounter{lemma}{0}\renewcommand{\thelemma}{\Alph{lemma}}

Appendix~\ref{app:examples} shows examples, and the rest of this appendix follows the order of the paper.
Appendices~\ref{app:detail} to~\ref{app:theory-more} give Sightline in detail, the proofs and further measurements for Section~\ref{sec:theory}.
Appendices~\ref{app:setup} to~\ref{app:serving} give the experimental setup and further results in the order of Section~\ref{sec:exp}.
Appendix~\ref{app:limits} discusses limitations.

\FloatBarrier
\section{Examples}
\label{app:examples}
Every example compares one sample of the original model with our best of eight for the same prompt, with the same loop control for both, and gives the seconds per prompt on one H200.
The prompts of Figure~\ref{fig:cases} are those where both best-of-eight rows beat one sample, by at least 0.05 in LPIPS for image-to-SVG and in all six verdicts of the three judges for text-to-SVG, and where our best of eight finishes first.
We rank them by the lower aesthetic score of the two best-of-eight renders (Appendix~\ref{app:metrics}), skip repeated objects, and replace prompts whose drawings read poorly at print size with the next in rank.

\subsection{Ten Prompts in Full}
\label{app:ex-full}
\begin{figure*}[t]
\centering
\includegraphics{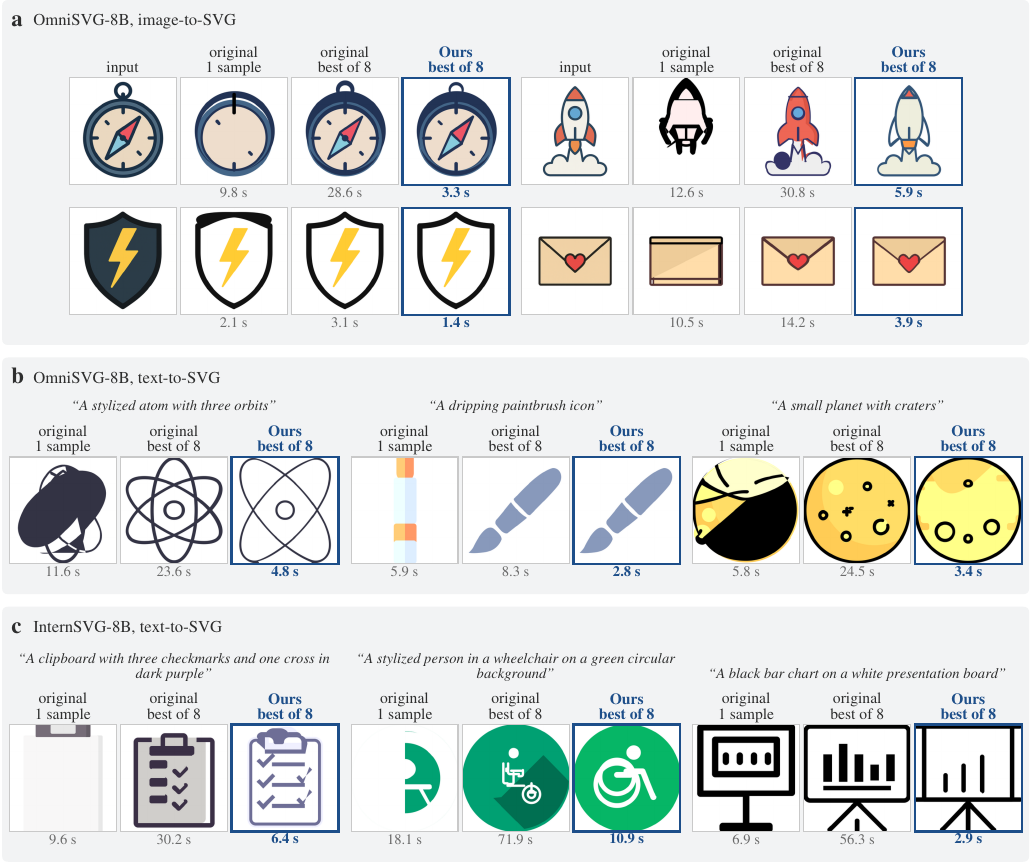}
\caption{\textbf{The 10 prompts of Figure~\ref{fig:cases} in full.} Each prompt shows one sample of the original model, its best of eight and our best of eight (blue), with the seconds per prompt under each, and image-to-SVG adds the input. Both best-of-eight rows use the same loop control and rendered selection.}
\label{app:ex-full-fig}
\end{figure*}

Figure~\ref{app:ex-full-fig} adds the original model's own best of eight to the 10 prompts of Figure~\ref{fig:cases}.
The two best-of-eight rows are of similar quality, as the judges of Figure~\ref{fig:ex2}j find over all 300 prompts, and ours takes a fraction of the time.

\subsection{More Prompts under the Same Rule}
\label{app:ex-more}
\begin{figure*}[t]
\centering
\includegraphics{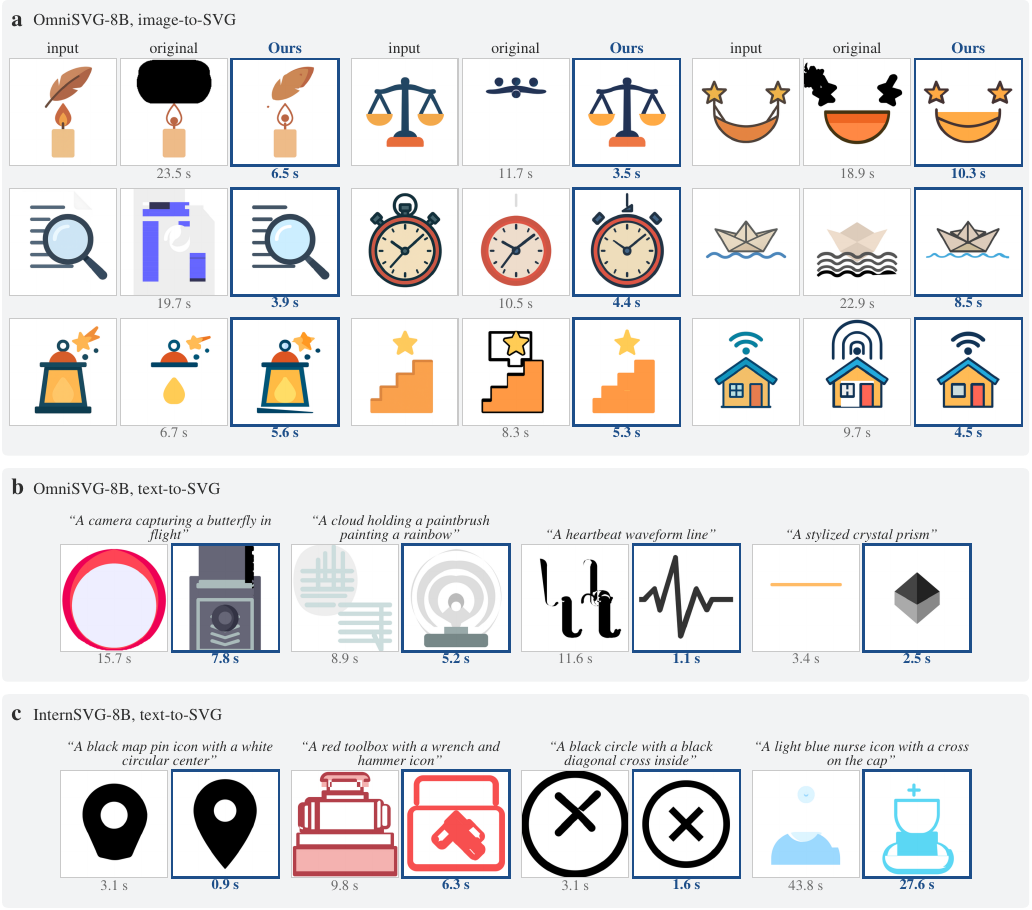}
\caption{\textbf{More prompts under the rule of Figure~\ref{fig:cases}.} The prompts that follow those of Figure~\ref{fig:cases} in each ranking, laid out as in Figure~\ref{fig:cases}, with one sample of the original model and our best of eight (blue).}
\label{app:ex-more-fig}
\end{figure*}

Figure~\ref{app:ex-more-fig} continues each ranking past the prompts of Figure~\ref{fig:cases}, with nine image-to-SVG prompts and four prompts of each text-to-SVG setting.

\subsection{Prompts Drawn by a Fixed Seed}
\label{app:ex-random}
\begin{figure*}[t]
\centering
\includegraphics{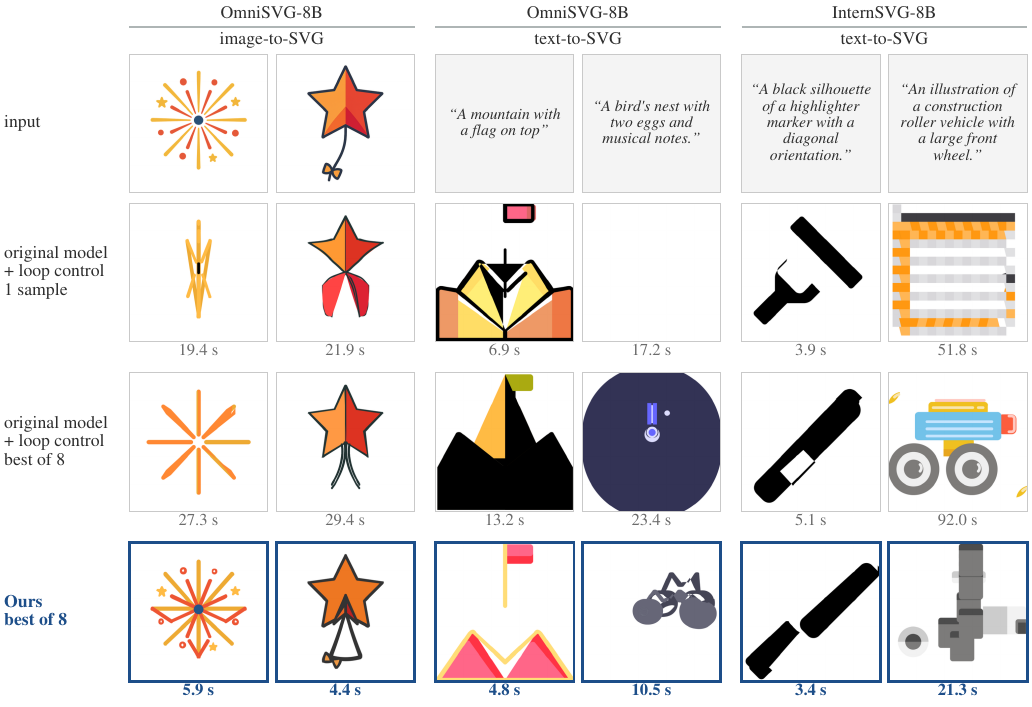}
\caption{\textbf{Qualitative results.} Two prompts in each of three settings, with the input, one sample and the best of eight of the original model with loop control, and our best of eight, and the seconds per prompt under each. Examples are drawn with a fixed seed among prompts where both best-of-eight rows beat one sample, by LPIPS for image-to-SVG and by two VLM judges in both orders for text-to-SVG, and where our best of eight finishes before one sample, as on 152 to 212 of the 300 prompts of each setting.}
\label{app:h4d}
\end{figure*}

\begin{figure*}[t]
\centering
\includegraphics{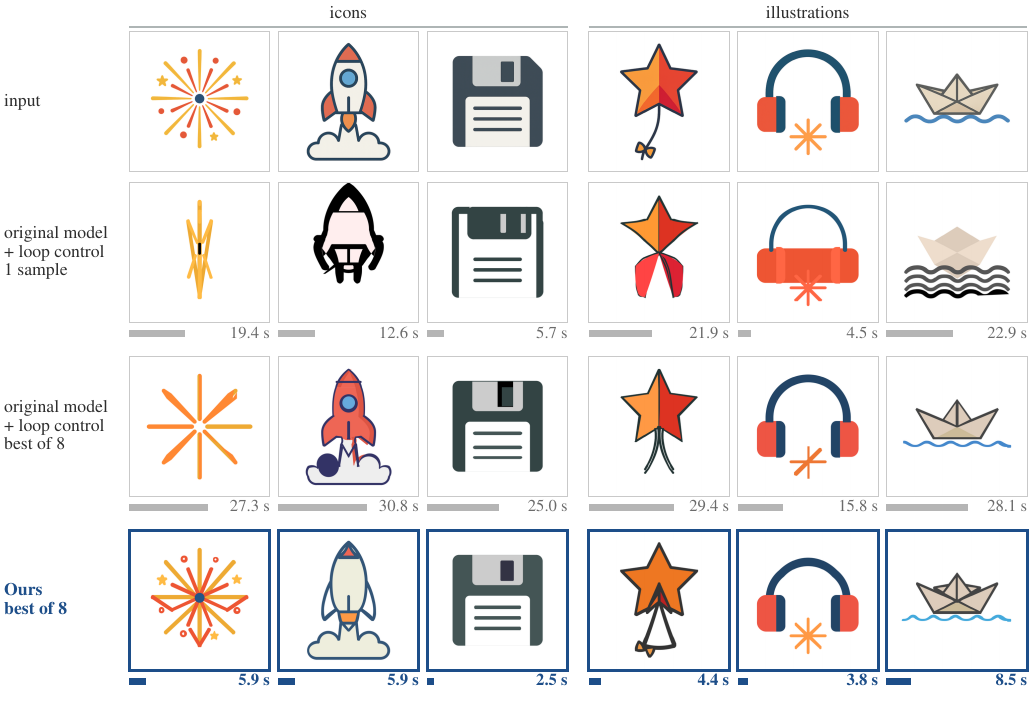}
\caption{\textbf{More qualitative results.} OmniSVG-8B image-to-SVG, three icons and three illustrations, rows as in Figure~\ref{app:h4d}, with the seconds drawn as bars on one scale. Examples follow the rule of Figure~\ref{app:h4d}, and our best of eight finishes first on 212 of the 300 prompts.}
\label{app:h4f}
\end{figure*}

\begin{figure*}[t]
\centering
\includegraphics{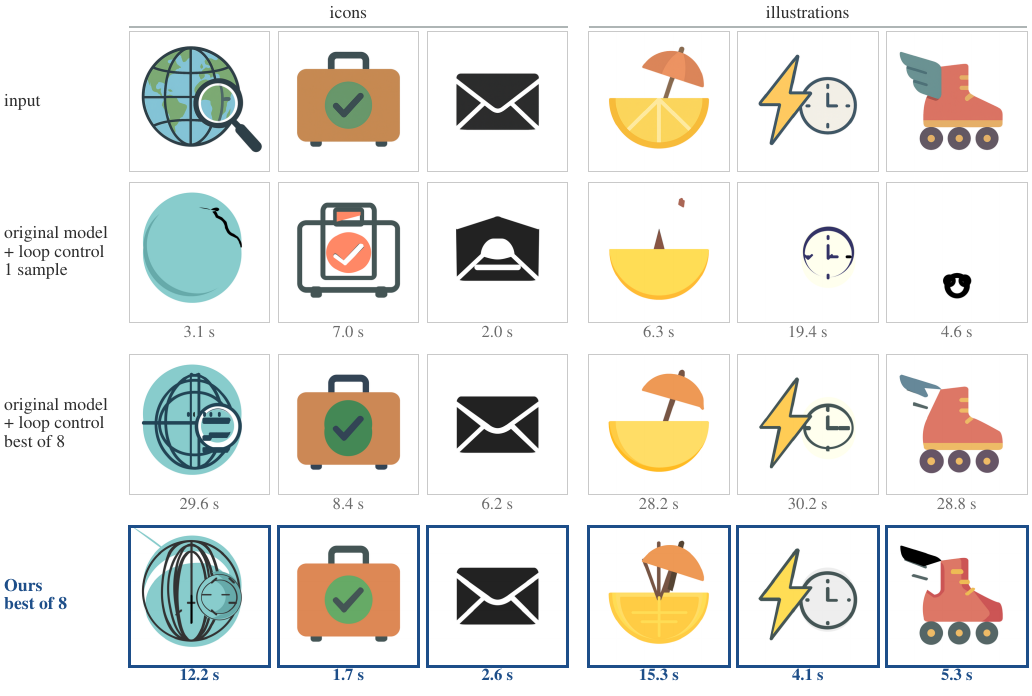}
\caption{\textbf{Qualitative results, any decoder first.} As Figure~\ref{app:h4f}, without the rule that our best of eight finishes first.}
\label{app:h4a}
\end{figure*}

Figures~\ref{app:h4d} and~\ref{app:h4f} draw their prompts with a fixed seed from a wider pool, the prompts where both best-of-eight rows beat one sample and ours finishes first, and also show the original model's best of eight.
Figure~\ref{app:h4a} repeats Figure~\ref{app:h4f} without asking that our best of eight finish first.


\subsection{Candidates and Decoding over Time}
\label{app:ex-time}
\begin{figure*}[t]
\centering
\includegraphics{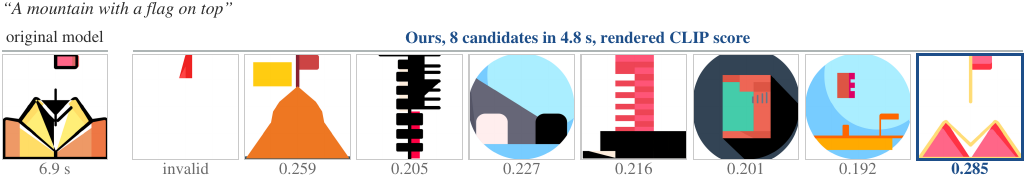}
\caption{\textbf{Eight candidates and their scores.} One sample of the original model for one prompt, and our eight candidates for the same prompt with their rendered CLIP scores. The kept candidate is framed.}
\label{app:h4c}
\end{figure*}

\begin{figure*}[t]
\centering
\includegraphics{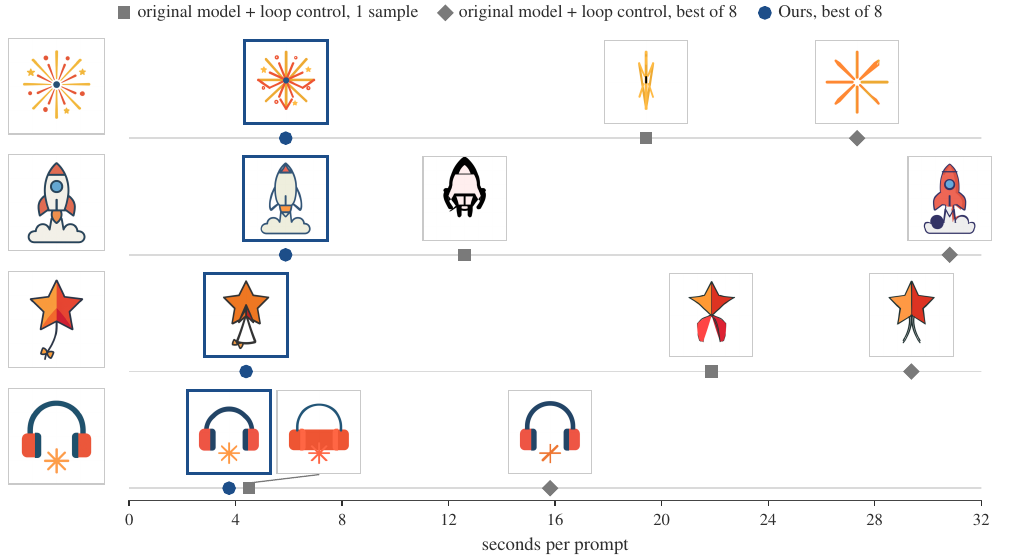}
\caption{\textbf{Decoding timelines.} Prompts of Figure~\ref{app:h4f}, one per row, with each output placed at the second its row finishes.}
\label{app:h4b}
\end{figure*}

\begin{figure*}[t]
\centering
\includegraphics{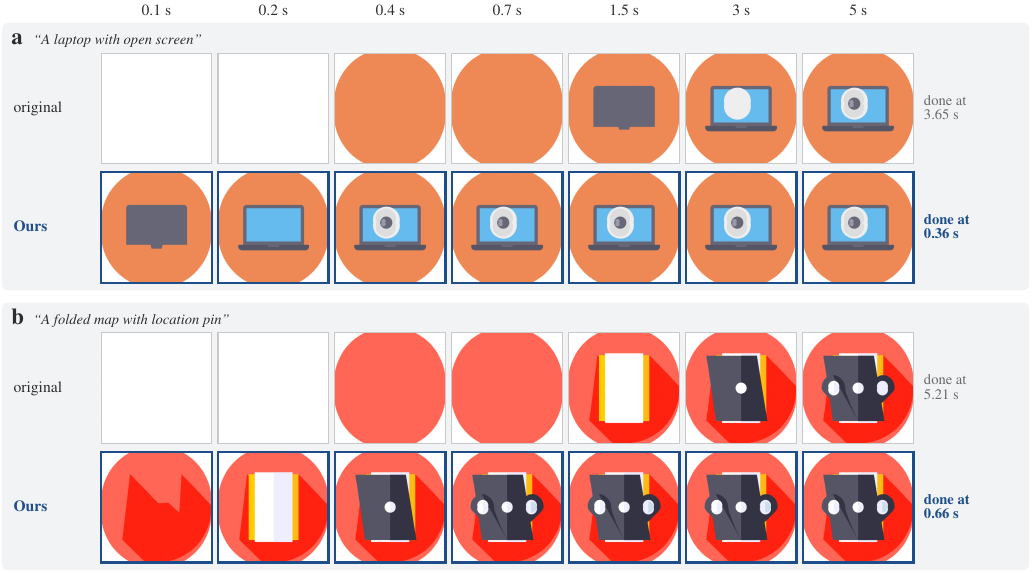}
\caption{\textbf{Decoding over time.} OmniSVG-8B text-to-SVG under greedy decoding on one H200, the drawings of the original model and of Sightline (blue) at the same instants. Both write the same tokens and draw the same SVG, and each frame shows the paths finished by its instant.}
\label{app:ex-film}
\end{figure*}

Figure~\ref{app:h4c} shows our eight candidates for one prompt, and Figure~\ref{app:h4b} places the outputs of Figure~\ref{app:h4f} on a time axis.
Figure~\ref{app:ex-film} follows two greedy decodings of OmniSVG-8B text-to-SVG as they draw.
Sightline writes the tokens of the original model, so both draw the same SVG, and by the time ours finishes, the original model has drawn little more than the background.
The prompts are MMSVG-Bench icons whose outputs end without a loop, whose tokens match those of the original model exactly, and whose speedup lies within 15\% of the median, and among them we show the drawings that read most clearly.
Both wall times come from the runs of Figure~\ref{fig:main}, and a frame spreads each decoder's time evenly over its forward passes.
A video in the code repository shows four such prompts in real time.


\FloatBarrier
\section{Sightline in Detail}
\label{app:detail}
This section spells out one round of Sightline (Algorithm~\ref{alg:round}), walks through the round of Figure~\ref{fig:method} and gives the details of the best of $N$.

\noindent\textbf{Draft tree.}
Let $C_j$ hold the $k$ most likely tokens of $q_j$, let $\mathcal{U}$ hold the continuations whose tokens come from these sets, and let $Q(u)=\prod_j q_j(u_j)$ be the drafter's probability of a continuation $u$.
The tree maximizes the drafter's estimate of $\mathbb{E}[A\mid x_{\le t}]=\sum_{u\in\mathcal{T}}p(u\mid x_{\le t})$ (Section~\ref{sec:prelim}), with $Q$ in place of $p$,
\begin{equation}
  \mathcal{T}=\operatorname*{arg\,max}_{\mathcal{S}\subseteq\mathcal{U},\ |\mathcal{S}|\le B}\ \sum_{u\in\mathcal{S}}Q(u).
  \label{eq:tree}
\end{equation}
A continuation's $Q$ is at most its parent's, so the maximizer consists of the $B$ continuations of highest $Q$ and is closed under prefixes.
We grow it best-first from the root, as DDTree does, with $k=16$ and $B=127$.

\noindent\textbf{Draft head.}
The head reads the hidden states of target layers 1, 7, 13, 19 and 25 for $x_{<t}$ and the embedding of $x_t$, and returns $q_1,\dots,q_{W-1}$ in one forward pass.
It sees none of the tokens it drafts, so each $q_j$ is a marginal for position $t+j$, the property that Proposition~\ref{prop:horizon} analyzes.
Appendix~\ref{app:training} gives its training.

\noindent\textbf{Loop control.}
At every position the rule looks for the shortest cycle of 8 to 128 tokens whose last three copies end the output, and it allows 256 tokens on InternSVG-8B image-to-SVG, whose outputs run up to 4,000 tokens.
At the first $m$ firings it bans the token that would start the next copy, and at the next firing it stops the output, as in~\eqref{eq:loop}.
The lossless decoders use no loop control, the setting with loop control uses $m=1$, and each sample of the best of $N$ stops at its first firing ($m=0$).

\noindent\textbf{Best of $N$.}
The $N$ samples decode in one batch, each by rounds of Algorithm~\ref{alg:round} with $k=1$.
Among the samples that render, the one with the highest render score is kept, by CLIP score against the prompt or DINOv2 similarity to the input image.

\noindent\textbf{A round in detail.}
Figure~\ref{fig:method} draws the first round of OmniSVG-8B text-to-SVG on one MMSVG prompt.
The round accepts 21 drafted tokens, six commands and 15 coordinates, and adds a bonus token.
At positions 11 and 15, the target takes the drafter's second candidate, a command in both cases.
A chain of first candidates would stop at position 11.
A window of 16 would stop after position 15 and write 16 tokens, while this round writes 22.
The tree places 81 of its 127 nodes beyond position 15, where a window of 16 drafts nothing.

\begin{algorithm}[t]
\caption{One round of Sightline}
\label{alg:round}
\begin{algorithmic}[1]
\Require prefix $x_{\le t}$, draft head $f$, window $W$, candidates per position $k$, node budget $B$, loop setting $m$
\State $q_1,\dots,q_{W-1}\gets f(x_{\le t})$ \Comment{one pass of the head}
\State $C_j\gets$ the $k$ most likely tokens of $q_j$ for $j=1,\dots,W-1$
\State $\mathcal{T}\gets$ the $B$ continuations over $C_1,\dots,C_{W-1}$ of highest $\prod_j q_j(u_j)$ \Comment{closed under prefixes}
\State score every node of $\mathcal{T}$ in one forward pass of the target
\For{$d=1,\dots,W$}
  \State draw $y_d$ from $\tilde p_m(\cdot\mid x_{\le t},y_{<d})$ \Comment{the target's own rule, \eqref{eq:loop}}
  \If{$d=W$ \textbf{or} $(y_1,\dots,y_d)\notin\mathcal{T}$} \Return $y_1,\dots,y_d$ \Comment{$A=d-1$ drafted tokens and a bonus token}
  \EndIf
\EndFor
\end{algorithmic}
\end{algorithm}

\FloatBarrier
\section{Proofs}
\label{app:proofs}
This section proves the statements of Section~\ref{sec:theory}.
A round starts at the first position not yet written, drafts the next $W-1$ positions, accepts the longest prefix of the draft that the target confirms, and then writes the target's own next token as the bonus token.

\subsection{Proof of Lemma~\ref{lem:runlength}}
\label{app:proof-lemma}
Under the condition of the lemma, each position of the output is either correct or a miss, whichever round drafts it.

\noindent\textbf{One round.}
Let a round start at position $s$.
If the first miss among $s,\dots,s+W-2$ is at $m$, the round accepts $s,\dots,m-1$, and its bonus token writes $m$, since the bonus token is the target's own.
The round then writes $m-s+1$ tokens, and the next round starts at $m+1$.
If none of $s,\dots,s+W-2$ is a miss, the round accepts all $W-1$ drafts, its bonus token writes $s+W-1$, and it writes $W$ tokens.

\noindent\textbf{Rounds and runs.}
A run ends with a miss, and a miss ends the round that writes it.
The first round starts at the first position, so every run starts with a new round.

\noindent\textbf{Rounds per run.}
In a run of length $n$, every position before the last one is correct.
Each round that starts inside the run advances $W$ positions until at most $W$ positions remain, and one more round writes the rest, ending with the miss as its bonus token.
The run thus takes $\lceil n/W\rceil$ rounds.
When $n$ is a multiple of $W$, the last round drafts $W-1$ correct positions and writes the miss as its bonus token.
An unclosed last run ends with the output and takes $\lceil n/W\rceil$ rounds in the same way.

\noindent\textbf{Sum.}
Adding over the runs, the decoder spends $\sum_i\lceil n_i/W\rceil$ rounds on $\sum_i n_i$ tokens, which gives \eqref{eq:runlength}.\hfill$\square$

\noindent\textbf{Long outputs.}
Let the runs of a long output be independent copies of $n=G+1$, where $G$ counts the correct positions before a miss.
By the renewal--reward theorem, $\tau_W$ tends to $\mathbb{E}[n]/\mathbb{E}\lceil n/W\rceil$ as the output grows.
Since $\lceil (G+1)/W\rceil=1+\lfloor G/W\rfloor$ and $\mathbb{E}\lfloor G/W\rfloor=\sum_{m\ge1}\Pr(G\ge mW)$,
\begin{equation}
  \tau_W\longrightarrow\frac{1+\mathbb{E}[G]}{1+\sum_{m\ge 1}\Pr(G\ge mW)}.
  \label{eq:stationary}
\end{equation}
Independent misses at rate $\lambda$ give $\Pr(G\ge g)=(1-\lambda)^g$, and \eqref{eq:stationary} becomes $(1-(1-\lambda)^W)/\lambda$, the expression in Section~\ref{sec:theory}.
This form needs outputs much longer than their mean run, so Section~\ref{sec:theory} applies \eqref{eq:runlength} to each output.

\subsection{Proof of Proposition~\ref{prop:share}}
\label{app:proof-share}
\noindent\textbf{Share.}
Let the wider rule offer, at every position, a candidate set that contains the narrower one.
A miss under the wider rule is then a miss under the narrower one, so every run under the wider rule joins adjacent runs under the narrower one.
Write $L=\sum_i n_i$ for the length of the output, $P$ for the number of runs and $R=\sum_i\lceil n_i/W\rceil$ for the number of rounds.
Then $\tau_\infty=L/P$, $\tau_W=L/R$ and $\tau_W/\tau_\infty=P/R$.
Joining runs lowers $P$, so $\tau'_\infty\ge\tau_\infty$.
For the share it suffices to join two adjacent runs of lengths $a$ and $b$.
This removes one run and $\delta=\lceil a/W\rceil+\lceil b/W\rceil-\lceil (a+b)/W\rceil$ rounds, and $\delta\in\{0,1\}$ because $\lceil x\rceil+\lceil y\rceil-1\le\lceil x+y\rceil\le\lceil x\rceil+\lceil y\rceil$.
If $\delta=0$, the share falls to $(P-1)/R$.
If $\delta=1$, it becomes $(P-1)/(R-1)$, which is at most $P/R$ because every run takes at least one round and so $P\le R$.
Joining runs one pair at a time gives $\tau'_W/\tau'_\infty\le\tau_W/\tau_\infty$.\hfill$\square$

\noindent\textbf{Clustering tax.}
The second part rests on the convexity of the rounds per token.

\begin{lemma}[Convexity]
\label{lem:convex}
For a window $W\ge 1$, the rounds per token $r_W(\lambda)=\lambda/(1-(1-\lambda)^W)$ are convex in $\lambda\in(0,1)$, and strictly convex for $W\ge 2$.
\end{lemma}

\noindent\textit{Proof.}
Write $q=1-\lambda$ and $h(q)=\sum_{j=0}^{W-1}q^j$, so that $r_W=1/h$.
Since $(1/h)''=(2h'^2-hh'')/h^3$ and $h>0$, it suffices to show that $2h'^2-hh''\ge0$ on $(0,1)$.
Expanding the products,
\begin{equation}
  2h'^2-hh''=\sum_{j,k=0}^{W-1}\bigl(2jk-k(k-1)\bigr)\,q^{j+k-2}.
\end{equation}
Averaging the coefficients of $(j,k)$ and $(k,j)$, and writing $m=j+k$ and $d=j-k$, each ordered pair contributes $c(m,d)=\bigl(m^2/2+m-3d^2/2\bigr)/2$ to the coefficient of $q^{m-2}$.
For fixed $m$, $c(m,d)$ falls as $|d|$ grows.
Its sum over $d=-m,-m+2,\dots,m$ is zero, since $\sum_{i=0}^{m}(2i-m)^2=m(m+1)(m+2)/3$.
If $m\le W-1$, all these pairs occur, and the coefficient is zero.
If $m\ge W$, the bound $j,k\le W-1$ removes the pairs with the largest $|d|$, which carry the smallest $c(m,d)$, so the pairs that remain give a nonnegative coefficient.
Every coefficient is therefore nonnegative, and $(1/h)''\ge0$.
For $W\ge2$ the top power has $m=2(W-1)\ge W$ and the single pair $j=k=W-1$, whose coefficient is positive, so $r_W$ is strictly convex.
Convexity in $q$ is convexity in $\lambda=1-q$.\hfill$\square$

Inside part $i$, misses are independent at rate $\lambda_i$, so by \eqref{eq:stationary} the part needs $r_W(\lambda_i)$ rounds per token.
A run that crosses a boundary between two parts changes the count by at most one round, which vanishes per token as the parts grow.
The rounds per token of the whole output thus tend to $\sum_i w_i\,r_W(\lambda_i)$.
By Jensen's inequality and Lemma~\ref{lem:convex}, this is at least $r_W(\bar\lambda)$, with equality for $W\ge2$ only when all $\lambda_i$ are equal.\hfill$\square$

\subsection{Proof of Proposition~\ref{prop:horizon}}
\label{app:proof-horizon}
The set $C_j$ depends on $x_{\le t}$ alone.
The probability that it contains the target's token at $t+j$ therefore averages over the ways $b$ in which the positions before $t+j$ are resolved, and equals $\sum_b p(b\mid x_{\le t})\,p_b(C_j)=\bar p(C_j)$.
The first inequality of \eqref{eq:horizon} holds because $C_j$ has $k$ tokens.
For every set $S$ of $k$ tokens, $\bar p(S)=\sum_b p(b\mid x_{\le t})\,p_b(S)\le\sum_b p(b\mid x_{\le t})\max_{|S'|=k}p_b(S')$, and the maximum over $S$ gives the second inequality.
If one set $S^\star$ is among the $k$ most likely tokens of every $p_b$ with $p(b\mid x_{\le t})>0$, then $\bar p(S^\star)$ equals the right side, and the last two terms are equal.
Conversely, if they are equal, a maximizer $S^\circ$ of $\bar p$ satisfies
\begin{equation}
  \sum_b p(b\mid x_{\le t})\Bigl(\max_{|S'|=k}p_b(S')-p_b(S^\circ)\Bigr)=0.
\end{equation}
Every term is nonnegative, so $p_b(S^\circ)=\max_{|S'|=k}p_b(S')$ for every $b$ with $p(b\mid x_{\le t})>0$, and $S^\circ$ is such a set.\hfill$\square$

Under sample-and-compare a candidate set is accepted with the target's probability on it (Section~\ref{sec:prelim}), so \eqref{eq:horizon} also bounds the chance that the round accepts the draft at $t+j$.

\subsection{Proof of Proposition~\ref{prop:loop}}
\label{app:proof-loop}
Let an output of $L$ tokens carry loops on a share $\varphi$ of its tokens.
Without loop control, the decoder spends $(1-\varphi)L/\tilde\tau_W$ rounds on the rest.
The drafter does not miss inside the loops, so the decoder writes $W$ tokens per round there and spends $\varphi L/W$ rounds on them.
Loop control removes the loops and leaves the rest unchanged, so the decoder then spends $(1-\varphi)L/\tilde\tau_W$ rounds in all.
The ratio of the two counts is \eqref{eq:loopgain}.
The original model writes one token per round, so $\tilde\tau_1=1$ and $\gamma_1=1/(1-\varphi)$, and $\gamma_W\le\gamma_1$ because $\tilde\tau_W\le W$.
The lead of $W'$ over $W$ is the ratio of their rounds, and loop control divides the rounds of each decoder by its own $\gamma$.
It therefore multiplies the lead by $\gamma_{W'}/\gamma_W$, and by \eqref{eq:loopgain} this factor is below one exactly when $\tilde\tau_{W'}/W'<\tilde\tau_W/W$.\hfill$\square$

\FloatBarrier
\section{More Measurements for the Theory}
\label{app:theory-more}
This section gives the measurements behind Section~\ref{sec:theory}.

\subsection{Measurement Setup}
\label{app:theory-setup}
Unless a caption says otherwise, the measurements use OmniSVG-8B text-to-SVG under greedy decoding and count loop-free content.
A position is labeled against the head's candidates at offset 1, and it is a decision when the target's own next-token entropy there exceeds 0.2 nats.
Token classes follow the vocabulary of each target.
OmniSVG writes SVG with tokens of its own, and the id of a token tells whether it is a path command, a coordinate on a $200\times200$ grid, a color, or another token such as an arc parameter.
InternSVG-8B writes SVG as text, and the decoded string tells whether a token is a number, a path command, a color, markup, whitespace or punctuation, or other text.
Natural language tokens are digits, punctuation, whitespace, words or other tokens (Table~\ref{app:t8}).
Figure~\ref{fig:ex1}c and Appendix~\ref{app:classes} count coordinates, arc parameters and numbers as numbers, and commands, markup and the remaining text as structure.
They leave out colors, which make up under 2\% of the tokens.

\subsection{Runs and the Window}
\label{app:runs}
\begin{figure*}[t]
\centering
\includegraphics{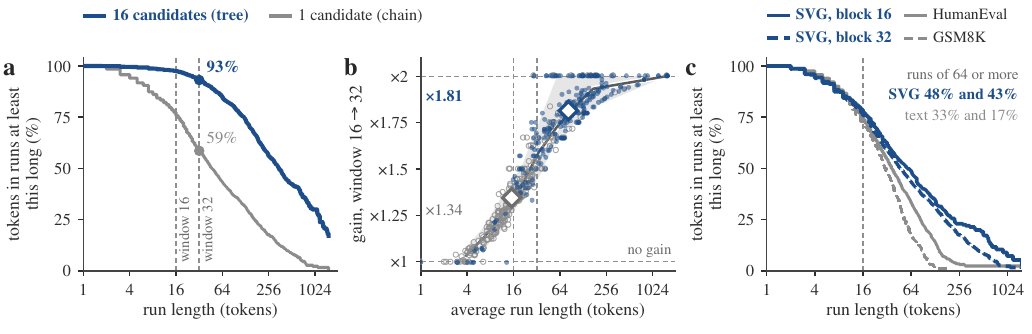}
\caption{\textbf{Run lengths.} (a)~Share of tokens in runs at least as long as the horizontal axis, for the head trained on blocks of 32 on 277 prompts, with one candidate (chain) or 16 (tree) per position. (b)~Gain from doubling the window from 16 to 32 against the average run length of each prompt, with the median prompt as a diamond. (c)~The share of (a) at one candidate per position, vector graphics against natural language. The head trained on blocks of 16 is paired with HumanEval and the head trained on blocks of 32 with GSM8K, at miss rates within 2\%. Natural language uses Qwen3-8B with the released DFlash head.}
\label{app:row-runs}
\end{figure*}

Figure~\ref{app:row-runs}a and~b show why a long window pays once the candidates are wide.
With 16 candidates per position, most tokens lie in runs longer than a window of 32, and the gain from doubling the window follows the average run length of a prompt for chains and trees alike.
Figure~\ref{app:row-runs}c pairs vector graphics with natural language at matched miss rates.
The two agree on short runs, and vector graphics hold more of their tokens in runs of 64 or more.
Table~\ref{app:t7} fits a model that keeps the accuracy of every position and draws positions independently.
It matches short windows and falls behind as the window grows, because it rarely produces rounds that fill the window.
At a window of 32 and 16 candidates per position, it gives about four fifths of the tokens per round of the real labels.
Accuracy per position alone thus does not fix the tokens per round of a long window.
The rest depends on how the misses fall along the output, which Lemma~\ref{lem:runlength} counts exactly.
The identity also prices a shorter window for the same head, to a median error of 2.7\% under official sampling (Table~\ref{app:t9}).
One head can therefore serve shorter windows at a predictable rate, as it does at high load in the serving experiment (Appendix~\ref{app:serving-load}).

\begin{table}[htbp]\centering
\begingroup
\renewenvironment{table}[1][]{\begin{minipage}[t]{0.62\textwidth}\centering}{\end{minipage}}%
\begin{table}[t]
\caption{\textbf{A conditionally independent model.} The model keeps the measured accuracy of every position, given its class, offset and the decisions crossed, and draws positions independently. It is fitted on half of the prompts and evaluated on the other half. Entries give its tokens per round as a share of the ideal value under the real labels. OmniSVG-8B text-to-SVG, greedy decoding, loop-free content. The share falls as the window grows, because rounds that fill the window are rarer in the model than in the real stream.}
\label{app:t7}
\centering
\footnotesize
\begin{tabular*}{\columnwidth}{@{\extracolsep{\fill}}cccccc@{}}
\toprule
Block & Candidates & $W=8$ & $W=16$ & $W=32$ & $W=64$ \\
\midrule
64 & 1 & 101\% & 98\% & 92\% & 89\% \\
 & 16 & 99\% & 93\% & 81\% & 72\% \\
32 & 1 & 100\% & 97\% & 91\% & -- \\
 & 16 & 100\% & 94\% & 82\% & -- \\
16 & 1 & 100\% & 97\% & -- & -- \\
 & 16 & 99\% & 93\% & -- & -- \\
\bottomrule
\end{tabular*}
\end{table}
\endgroup
\end{table}

\subsection{Decisions and the Horizon}
\label{app:decisions}
\begin{figure*}[t]
\centering
\includegraphics{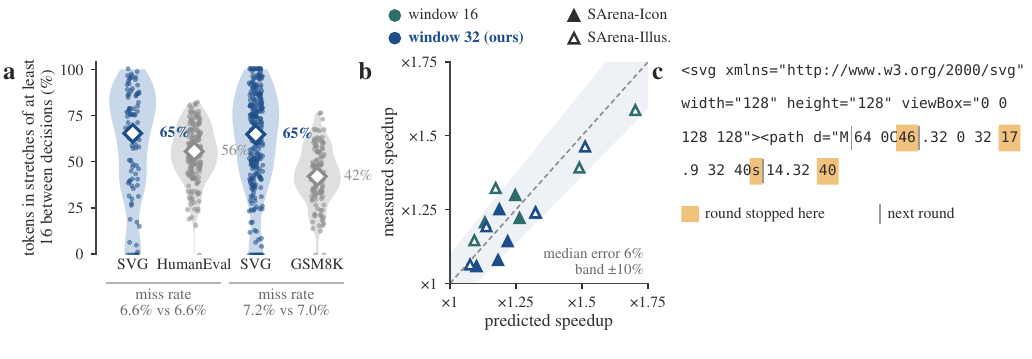}
\caption{\textbf{Decisions, loops and rounds.} (a)~Per prompt, the share of tokens in stretches of at least 16 between two decisions, for the two pairs of Figure~\ref{app:row-runs}c, with medians as diamonds. (b)~Predicted against measured gains of loop control on SArena, for windows of 16 and 32 (ours), icon (filled) and illustration (hollow). Each prediction puts the tokens per round measured inside loops in place of $W$ in \eqref{eq:loopgain}, and $W$ itself gives a median error of 5\%. The dashed line marks equality and the band $\pm10\%$. (c)~The opening of one InternSVG-8B text-to-SVG output under Sightline, chosen by a fixed rule. The first round writes the 32-token header at once. Each shaded token ends a round because the draft was wrong there, and bars mark where the next round starts.}
\label{app:row-theory}
\end{figure*}

Table~\ref{app:t3} breaks the drafter's accuracy down by token class, offset and the decisions crossed.
Commands keep most of their accuracy past two decisions, and coordinates keep less than half.
The head trained on blocks of 64 also loses accuracy beyond offset 32 without crossing any decision, which is where the drafter's horizon ends.
Figure~\ref{app:row-theory}a counts, per prompt, the tokens that lie in stretches of at least 16 between two decisions, and the median prompt of vector graphics keeps more of them.
Table~\ref{app:t8} compares the class that best keeps its accuracy in each corpus.
Under greedy decoding, the prefix fixes the target's next token, but the head cannot compute it.
The decisions and the mixture of Proposition~\ref{prop:horizon} therefore still come from the target's own distribution.
The measurements agree under greedy decoding and official sampling (Table~\ref{app:t5}).

\begin{table*}[t]
\caption{\textbf{Accuracy by class, offset and decisions crossed.} Top-1 accuracy of the draft head at positions that are not decisions themselves, by token class, offset in the block and the number $c$ of decisions crossed before the position. A decision is a position where the target's own next-token entropy exceeds 0.2 nats. OmniSVG-8B text-to-SVG, greedy decoding, loop-free content. Cells with fewer than 30 positions are left out, and thresholds of 0.05 and 0.5 nats give the same picture.}
\label{app:t3}
\centering
\footnotesize
\begin{tabular*}{\textwidth}{@{\extracolsep{\fill}}llcccccc@{}}
\toprule
 & & \multicolumn{3}{c}{Head trained on blocks of 32} & \multicolumn{3}{c}{Head trained on blocks of 64} \\
\cmidrule(lr){3-5}\cmidrule(lr){6-8}
Class & Offset & $c=0$ & $c=1$ & $c\ge2$ & $c=0$ & $c=1$ & $c\ge2$ \\
\midrule
Commands & 1--8 & 0.995 & 0.955 & 0.921 & 0.994 & 0.954 & 0.924 \\
 & 9--16 & 0.989 & 0.943 & 0.864 & 0.982 & 0.935 & 0.862 \\
 & 17--31 & 0.968 & 0.933 & 0.795 & 0.951 & 0.912 & 0.795 \\
 & 32--47 &  &  &  & 0.894 & 0.849 & 0.743 \\
 & 48--63 &  &  &  & 0.812 & 0.730 & 0.664 \\
\addlinespace[2pt]
Coordinates & 1--8 & 0.970 & 0.750 & 0.455 & 0.966 & 0.736 & 0.449 \\
 & 9--16 & 0.954 & 0.761 & 0.418 & 0.926 & 0.735 & 0.410 \\
 & 17--31 & 0.861 & 0.711 & 0.370 & 0.771 & 0.654 & 0.357 \\
 & 32--47 &  &  &  & 0.508 & 0.487 & 0.262 \\
 & 48--63 &  &  &  & 0.208 & 0.222 & 0.146 \\
\addlinespace[2pt]
Colors & 1--8 & 0.968 & 0.847 & -- & 0.957 & 0.800 & -- \\
 & 9--16 & 0.956 & 0.784 & 0.615 & 0.927 & 0.747 & 0.542 \\
 & 17--31 & 0.916 & 0.853 & 0.559 & 0.867 & 0.765 & 0.517 \\
 & 32--47 &  &  &  & 0.694 & 0.644 & 0.437 \\
 & 48--63 &  &  &  & 0.398 & 0.349 & 0.208 \\
\addlinespace[2pt]
Other & 1--8 & 0.991 & 0.924 & 0.730 & 0.989 & 0.918 & 0.784 \\
 & 9--16 & 0.963 & 0.935 & 0.741 & 0.960 & 0.907 & 0.766 \\
 & 17--31 & 0.939 & 0.925 & 0.732 & 0.893 & 0.852 & 0.720 \\
 & 32--47 &  &  &  & 0.726 & 0.749 & 0.628 \\
 & 48--63 &  &  &  & 0.531 & 0.604 & 0.511 \\
\bottomrule
\end{tabular*}
\end{table*}

\begin{table}[htbp]\centering
\begingroup
\renewenvironment{table}[1][]{\begin{minipage}[t]{0.48\textwidth}\centering}{\end{minipage}}%
\begin{table}[t]
\caption{\textbf{Vector graphics against natural language.} For each target and task, the token class that keeps the most of its accuracy past two decisions. Share is the class's share of the stream, and retention its accuracy after crossing two or more decisions divided by its accuracy when none is crossed. Qwen3-8B uses the released DFlash head under official decoding. The SVG rows are loop-free content.}
\label{app:t8}
\centering
\footnotesize
\setlength{\tabcolsep}{3pt}
\begin{tabular*}{\columnwidth}{@{\extracolsep{\fill}}lllcc@{}}
\toprule
Target and head & Task & Class & Share & Retention \\
\midrule
OmniSVG-8B, block 16 & text-to-SVG & Commands & 0.36 & 0.91 \\
OmniSVG-8B, block 32 & text-to-SVG & Commands & 0.36 & 0.89 \\
OmniSVG-8B, block 64 & text-to-SVG & Commands & 0.36 & 0.89 \\
OmniSVG-4B, block 16 & text-to-SVG & Commands & 0.43 & 0.93 \\
\addlinespace[2pt]
Qwen3-8B, block 16 & GSM8K & Punctuation & 0.34 & 0.34 \\
Qwen3-8B, block 16 & MATH-500 & Punctuation & 0.40 & 0.48 \\
Qwen3-8B, block 16 & HumanEval & Words & 0.45 & 0.34 \\
Qwen3-8B, block 16 & MBPP & Words & 0.49 & 0.34 \\
Qwen3-8B, block 16 & MT-Bench & Punctuation & 0.32 & 0.40 \\
\bottomrule
\end{tabular*}
\end{table}
\endgroup
\end{table}

\subsection{Loops}
\label{app:loops-theory}
Figure~\ref{app:row-theory}b compares, setting by setting, the speedup from loop control that \eqref{eq:loopgain} predicts with the measured one.
The original model writes one token per round, so its gain $\gamma_1=1/(1-\varphi)$ needs no estimate.
This curve predicts its eight measured gains in Figure~\ref{fig:theory}f to a median error of 6\% and at most 13.7\%, on the node-scaled seconds of Table~\ref{tab:quality} (Appendix~\ref{app:timing}).
The 16 points of DDTree and Sightline pair a lossless row with a loop-control row from the same job, so no node factor enters them.

\subsection{Other Decodings and Content}
\label{app:robust}
Table~\ref{app:t5} repeats the measurements under official sampling, and Table~\ref{app:t6} on strict loop-free content, which also leaves out outputs that repeat without an exact cycle.
The conclusions of Section~\ref{sec:theory} hold in both.

\begin{table*}[t]
\caption{\textbf{The theory under official sampling.} OmniSVG-8B text-to-SVG, loop-free content, three heads trained on blocks of 64, 32 and 16. Miss rate and run-length CV are for one candidate per position. Ceiling to Budget are the tokens per round of a 127-node tree as the model gains each loss, the window at the head's own block length, the horizon and the node budget, and Measured is the decoder itself. The last three columns give the share of positions that are decisions and the accuracy of commands and coordinates at offsets 1 to 8 after two or more decisions. Lemma~\ref{lem:runlength} matches a round-by-round simulation exactly under both decodings.}
\label{app:t5}
\centering
\footnotesize
\setlength{\tabcolsep}{3pt}
\begin{tabular*}{\textwidth}{@{\extracolsep{\fill}}llcccccccccc@{}}
\toprule
Block & Decoding & Miss rate & Run CV & Ceiling & Window & Horizon & Budget & Measured & Decisions & Commands & Coordinates \\
\midrule
64 & greedy & 7.5\% & 3.16 & 79.00 & 41.13 & 20.58 & 12.23 & 12.24 & 14\% & 0.92 & 0.45 \\
 & official sampling & 9.4\% & 3.31 & 62.12 & 36.62 & 17.45 & 10.71 & 10.71 & 17\% & 0.91 & 0.40 \\
\addlinespace[2pt]
32 & greedy & 7.2\% & 2.97 & 82.91 & 25.83 & 18.50 & 12.35 & 12.37 & 15\% & 0.92 & 0.46 \\
 & official sampling & 7.9\% & 3.29 & 64.61 & 24.16 & 16.87 & 11.58 & 11.61 & 15\% & 0.91 & 0.43 \\
\addlinespace[2pt]
16 & greedy & 6.7\% & 3.63 & 86.04 & 14.45 & 12.73 & 10.37 & 10.36 & 15\% & 0.93 & 0.50 \\
 & official sampling & 8.1\% & 3.03 & 64.30 & 13.86 & 11.75 & 9.63 & 9.63 & 17\% & 0.92 & 0.54 \\
\bottomrule
\end{tabular*}
\end{table*}

\begin{table*}[t]
\caption{\textbf{Robustness on strict loop-free content.} Strict loop-free content also leaves out outputs that repeat without an exact cycle. Columns as in Table~\ref{app:t5}, greedy decoding. Every direction holds. The ceiling falls most, since many of the removed outputs have no miss at all.}
\label{app:t6}
\centering
\footnotesize
\setlength{\tabcolsep}{3pt}
\begin{tabular*}{\textwidth}{@{\extracolsep{\fill}}llcccccccc@{}}
\toprule
Block & Content & Ceiling & Window & Horizon & Budget & Measured & Run CV & Commands & Coordinates \\
\midrule
64 & loop-free & 79.00 & 41.13 & 20.58 & 12.23 & 12.24 & 3.16 & 0.92 & 0.45 \\
 & strict & 52.03 & 34.39 & 16.92 & 11.42 & 11.43 & 2.53 & 0.88 & 0.40 \\
\addlinespace[2pt]
32 & loop-free & 82.91 & 25.83 & 18.50 & 12.35 & 12.37 & 2.97 & 0.92 & 0.46 \\
 & strict & 55.82 & 23.49 & 16.12 & 11.60 & 11.62 & 2.57 & 0.88 & 0.41 \\
\addlinespace[2pt]
16 & loop-free & 86.04 & 14.45 & 12.73 & 10.37 & 10.36 & 3.63 & 0.93 & 0.50 \\
 & strict & 56.44 & 13.72 & 11.66 & 9.68 & 9.66 & 2.96 & 0.90 & 0.41 \\
\bottomrule
\end{tabular*}
\end{table*}

\subsection{Registered Predictions}
\label{app:predictions}
Table~\ref{app:t9} lists every prediction we registered before reading the corresponding results, with its outcome.

\begin{table*}[t]
\caption{\textbf{Predictions and outcomes.} Every prediction we registered before reading the corresponding results, grouped by the statement it tests. Greedy rows use loop-free content of OmniSVG-8B text-to-SVG, and sampled rows the official sampling of the same pair. Predictions that did not hold or remain undecided are listed with the rest, and no claim of the paper rests on them.}
\label{app:t9}
\centering
\footnotesize
\renewcommand{\arraystretch}{1.1}
\begin{tabular}{@{}p{0.10\textwidth}p{0.09\textwidth}p{0.34\textwidth}p{0.28\textwidth}p{0.08\textwidth}@{}}
\toprule
Statement & Decoding & Prediction & Outcome & Verdict \\
\midrule
Lemma~\ref{lem:runlength} & Greedy & The identity on full-window labels bounds the tokens per round of a shorter window from below & 5 of 6 cases on OmniSVG-8B, 5 of 5 on InternSVG-8B & held \\
 & Sampled & The identity lies within 3\% below the measured value in at least 4 of 5 cases per head & 8 of 10 and 3 of 3 cases, every shorter window, median error 2.7\% over 13 cases & held \\
 & Greedy & An offline simulation on cached drafts matches a tree decoder that verifies a prefix of its block to about 1\% & 2.5\% to 3.5\% & not met \\
 & Greedy & Predictions per case for a chain decoder, from its measured full-window run & errors from $-11.4\%$ to $+19.0\%$ over 7 cases & -- \\
 & Sampled & Main predictions per case from cached drafts & median error 6.9\% over 16 cases, 10 of 16 inside the registered band & -- \\
 & Both & Drafting fewer positions is no better than verifying a prefix of the block & 5 of 6 cases greedy, 6 of 6 sampled, two sampled magnitudes outside the registered range & held in 11 of 12 \\
\midrule
Proposition~\ref{prop:share} & Greedy & The return from doubling the window from 16 to 32 grows as the candidates widen & $\times1.27\to\times1.78$ and $\times1.28\to\times1.79$ for the two heads, one to 16 candidates & held \\
 & Greedy & Window and node budget have increasing differences & every pair for the three heads & held \\
 & Sampled & Window and design have increasing differences & 5 of 5 & held \\
 & Sampled & The cost of sampling relative to greedy does not depend on the window (80\% of cases within 6\%) & 5 of 12; 9 of 12 confidence intervals still contain the registered value & did not hold \\
 & Sampled & A tree pays more for sampling than a chain & 4 of 7, every confidence interval overlaps & undecided \\
\midrule
Proposition~\ref{prop:horizon} & Greedy & At shared positions the head trained on blocks of 64 pays its extra loss on coordinates & offset 15: commands $0.999\to0.994$, coordinates $0.991\to0.958$ & held \\
 & Greedy & The head trained on blocks of 64, cut to a window of 32, is weaker than the head trained on blocks of 32 & tree $-7.9\%$ (predicted $-5.6\%$), chain $-1.2\%$ & held \\
 & Sampled & The same under official sampling & tree $-9.8\%$, chain $-2.3\%$ & held \\
\midrule
Proposition~\ref{prop:loop} & Both & Loop control speeds up the original model most and Sightline least & 10 of 10 settings & held \\
 & Both & Point predictions of the speedup from loop control & median error 6.1\% over 16 settings & -- \\
 & Both & With loop control on both, the lead of a window of 32 over 16 narrows and remains & lead kept in 8 of 8 settings, narrowed in 7 of 8 & held \\
\bottomrule
\end{tabular}
\end{table*}

\FloatBarrier
\section{Experimental Setup}
\label{app:setup}

\subsection{Models, Benchmarks and Decoding}
\label{app:models}
We accelerate the released OmniSVG1.1\_8B and OmniSVG1.1\_4B, which we call OmniSVG-8B and OmniSVG-4B, and InternSVG-8B.
OmniSVG-3B of Figure~\ref{fig:ex1}a is the earlier release of OmniSVG.
Each setting decodes every prompt of its benchmark set, 150 per set for MMSVG, 300 for MMSVG-2M and 450 for SArena.
The draft heads train on MMSVG-2M, and its two benchmark sets are held out from that training.
MMSVG synthesizes its prompts and images with GPT-4o, and SArena is the benchmark of InternSVG, so neither is drawn from MMSVG-2M.
InternSVG-8B reads an image together with a one-sentence description of it.
MMSVG ships its images without text, so a DeepSeek model (deepseek-flash, temperature 0) wrote a description for each of its 300 images, in the style of the MMSVG captions.

Greedy decoding writes the most likely token at every step.
Sampling uses the settings that each model releases (Table~\ref{app:decoding}).
InternSVG-8B releases greedy decoding only, and its sampled runs use temperature 0.6 and top-$p$ 0.9.
Loop control and the best of $N$ use the rule and the scores of Section~\ref{sec:sightline} for every decoder.
Loop control escapes once and then stops ($m=1$), and the best of $N$ draws $N=8$ samples and stops each at its first firing.
On InternSVG-8B image-to-SVG, whose outputs run up to 4,000 tokens, the cycle for loop control may span 8 to 256 tokens.

\begin{table}[t]
\centering
\begin{minipage}{0.62\textwidth}
\caption{\textbf{Sampling settings.} OmniSVG's come from its released configuration, per task, for both OmniSVG-8B and OmniSVG-4B, and Penalty is its repetition penalty. InternSVG-8B releases greedy decoding only.}
\label{app:decoding}
\centering
\footnotesize
\setlength{\tabcolsep}{3.0pt}
\begin{tabular*}{\columnwidth}{@{\extracolsep{\fill}}llcccc@{}}
\toprule
Model & Task & Temp. & top-$p$ & top-$k$ & Penalty \\
\midrule
OmniSVG & text-to-SVG, icon & 0.5 & 0.88 & 50 & 1.05 \\
OmniSVG & text-to-SVG, illus. & 0.6 & 0.90 & 60 & 1.03 \\
OmniSVG-8B & image-to-SVG & 0.3 & 0.90 & 50 & 1.05 \\
InternSVG-8B & both & 0.6 & 0.90 & -- & -- \\
\bottomrule
\end{tabular*}
\end{minipage}
\end{table}

\subsection{Baselines}
\label{app:baselines}
Every decoder runs on the same target with the verifier of Section~\ref{sec:prelim}, and its speedup is timed against the original model in the same job.
PLD and REST draft without training.
PLD copies up to 10 tokens that follow the latest match of at most 3 context tokens.
REST retrieves continuations from a datastore and verifies them as a tree of 127 nodes.
SpS drafts OmniSVG-8B with the smaller model of its family.
EAGLE-3 and ViSpec train autoregressive heads.
EAGLE-3 verifies a dynamic tree of 127 nodes, at most 12 deep.
ViSpec runs on image-to-SVG, and on OmniSVG-8B its head ends with an RMSNorm.
DFlash, DDTree and DARTree train block drafters.
DFlash drafts a chain over a window of 16, and DDTree builds a tree of 127 nodes from the top 16 candidates of the same block.
DARTree grows a tree of 127 nodes with its autoregressive correction head.
Every trained drafter keeps the checkpoint with the highest tokens per round on held-out MMSVG-2M prompts.
Every compared drafter proposes its candidates deterministically, as the most likely tokens of its own distribution or as copied text, and SpS drafts greedily from the smaller model.
For such candidates, no exact rule can accept a position more often than the target's probability of its candidates, and the walk of Section~\ref{sec:prelim} reaches this bound.
Under greedy decoding it is every method's own rule, and under sampling it gives every decoder the most it can accept from its candidates.
Table~\ref{app:decoders} lists the window, the candidates per position and the node budget of every decoder.
Without training, the better of PLD and REST reaches a median speedup of $1.96\times$ over the 60 settings of Figure~\ref{fig:main}, between $1.55\times$ and $3.80\times$.
Sightline runs 2.7 to 4.2 times as fast as the better of the two in every setting.

\begin{table*}[t]
\caption{\textbf{Decoders compared.} The window $W$ is the most tokens a round can write, $k$ the candidates a drafter offers per position and $B$ the nodes the target verifies in one pass (Section~\ref{sec:prelim}). Every decoder runs on the same target with the same verifier, and the block drafters we compare all sit near a window of 16.}
\label{app:decoders}
\centering
\footnotesize
\begin{tabular}{@{}llccll@{}}
\toprule
Method & Drafter & $W$ & $k$ & $B$ & Design \\
\midrule
AR & -- & 1 & -- & -- & the original model \\
PLD & context copy & 11 & 1 & 11 & up to 10 tokens after the latest match \\
REST & datastore & 17 & varies & 127 & continuations retrieved from a datastore \\
SpS & smaller model & 11 & 1 & 11 & OmniSVG-4B drafts OmniSVG-8B \\
EAGLE-3 & autoregressive head & 13 & varies & 127 & a dynamic tree, at most 12 deep \\
ViSpec & autoregressive head & 13 & varies & 127 & image-to-SVG only \\
DFlash & block head, 16 & 16 & 1 & 15 & a chain over the block \\
DDTree & block head, 16 & 16 & 16 & 127 & a tree over the same block \\
DARTree & block head, 12 + correction & 17 & varies & 127 & an autoregressive correction head \\
\textbf{Ours} & block head, 32 & 32 & 16 & 127 & a tree over a block of 32 \\
\bottomrule
\end{tabular}
\end{table*}

\subsection{Training}
\label{app:training}
Each draft head trains on outputs of its own target for MMSVG-2M prompts (Table~\ref{app:hparams}).
Ours follows the design and recipe of DFlash at block length 32, with the data, epochs, batch size and learning rate of the DFlash head of the same target.
All block heads read the hidden states of target layers 1, 7, 13, 19 and 25.
The image-to-SVG heads start from the text-to-SVG head of the same target, and the InternSVG-8B text-to-SVG heads continue a first round of training.
Table~\ref{app:hparams} gives the epochs of the kept checkpoints, so the DARTree head of OmniSVG-8B text-to-SVG is its second epoch, which beat its sixth on the held-out prompts.

\begin{table*}[t]
\caption{\textbf{Training budget and hyper-parameters of every draft head.} \emph{Data} is the number of generated training prompts, \emph{Ep.} the passes over them, and \emph{Layers} the target layers whose hidden states the head reads. \emph{Warm start from} marks a head initialized from an earlier head. Our heads and those of DFlash and DDTree share every setting except the block length, 32 against 16, and DFlash and DDTree share one head per target.}
\label{app:hparams}
\centering
\footnotesize
\renewcommand{\arraystretch}{0.85}
\setlength{\tabcolsep}{2.0pt}
\begin{tabular*}{\textwidth}{@{\extracolsep{\fill}}ll*{7}{c}@{}}
\toprule
Model and task & Method & Data & Ep. & Batch & LR & Len. & Layers & Warm start from \\
\midrule
\multirow{5}{*}{OmniSVG-8B text} & EAGLE-3 & 300k & 6 & 16 & $1\!\times\!10^{-4}$ & 2048 & -- & -- \\
 & DFlash & 300k & 6 & 12 & $6\!\times\!10^{-4}$ & 2048 & 1,7,13,19,25 & -- \\
 & DARTree & 300k & 2 & 12 & $6\!\times\!10^{-4}$ & 2048 & 1,7,13,19,25 & -- \\
 & DDTree & 300k & 6 & 12 & $6\!\times\!10^{-4}$ & 2048 & 1,7,13,19,25 & -- \\
 & \textbf{Ours} & 300k & 6 & 12 & $6\!\times\!10^{-4}$ & 2048 & 1,7,13,19,25 & -- \\
\midrule
\multirow{5}{*}{OmniSVG-4B text} & EAGLE-3 & 300k & 6 & 16 & $1\!\times\!10^{-4}$ & 2048 & -- & -- \\
 & DFlash & 300k & 6 & 16 & $6\!\times\!10^{-4}$ & 2048 & 1,7,13,19,25 & -- \\
 & DARTree & 300k & 6 & 16 & $6\!\times\!10^{-4}$ & 2048 & 1,7,13,19,25 & -- \\
 & DDTree & 300k & 6 & 16 & $6\!\times\!10^{-4}$ & 2048 & 1,7,13,19,25 & -- \\
 & \textbf{Ours} & 300k & 6 & 16 & $6\!\times\!10^{-4}$ & 2048 & 1,7,13,19,25 & -- \\
\midrule
\multirow{5}{*}{InternSVG-8B text} & EAGLE-3 & 600k & 1 & 16 & $5\!\times\!10^{-5}$ & 4096 & -- & 300k, 2 ep \\
 & DFlash & 600k & 1 & 16 & $3\!\times\!10^{-4}$ & 4096 & 1,7,13,19,25 & 300k, 2 ep \\
 & DARTree & 600k & 1 & 16 & $3\!\times\!10^{-4}$ & 4096 & 1,7,13,19,25 & 300k, 2 ep \\
 & DDTree & 600k & 1 & 16 & $3\!\times\!10^{-4}$ & 4096 & 1,7,13,19,25 & 300k, 2 ep \\
 & \textbf{Ours} & 600k & 1 & 16 & $3\!\times\!10^{-4}$ & 4096 & 1,7,13,19,25 & 300k, 2 ep \\
\midrule
\multirow{6}{*}{OmniSVG-8B image} & EAGLE-3 & 100k & 3 & 16 & $1\!\times\!10^{-4}$ & 2048 & -- & 300k, 6 ep \\
 & ViSpec & 100k & -- & 16 & $3\!\times\!10^{-6}$ & 2048 & -- & 300k \\
 & DFlash & 100k & 3 & 16 & $6\!\times\!10^{-4}$ & 2048 & 1,7,13,19,25 & 300k, 6 ep \\
 & DARTree & 100k & 3 & 16 & $6\!\times\!10^{-4}$ & 2048 & 1,7,13,19,25 & 300k, 4 ep \\
 & DDTree & 100k & 3 & 16 & $6\!\times\!10^{-4}$ & 2048 & 1,7,13,19,25 & 300k, 6 ep \\
 & \textbf{Ours} & 100k & 3 & 16 & $6\!\times\!10^{-4}$ & 2048 & 1,7,13,19,25 & 300k, 6 ep \\
\midrule
\multirow{6}{*}{InternSVG-8B image} & EAGLE-3 & 100k & 3 & 16 & $1\!\times\!10^{-4}$ & 4096 & -- & 600k, 1 ep \\
 & ViSpec & 100k & -- & 16 & $3\!\times\!10^{-6}$ & 2048 & -- & 300k \\
 & DFlash & 100k & 3 & 16 & $6\!\times\!10^{-4}$ & 4096 & 1,7,13,19,25 & 600k, 1 ep \\
 & DARTree & 100k & 3 & 16 & $6\!\times\!10^{-4}$ & 4096 & 1,7,13,19,25 & 600k, 1 ep \\
 & DDTree & 100k & 3 & 16 & $6\!\times\!10^{-4}$ & 4096 & 1,7,13,19,25 & 600k, 1 ep \\
 & \textbf{Ours} & 100k & 3 & 16 & $6\!\times\!10^{-4}$ & 4096 & 1,7,13,19,25 & 600k, 1 ep \\
\midrule
\multirow{5}{*}{SVGen-3B} & EAGLE-3 & 100k & 2 & 16 & $1\!\times\!10^{-4}$ & 2048 & -- & -- \\
 & DFlash & 100k & 2 & 16 & $6\!\times\!10^{-4}$ & 2048 & 1,7,13,19,25 & -- \\
 & DARTree & 100k & 2 & 16 & $6\!\times\!10^{-4}$ & 2048 & 1,7,13,19,25 & -- \\
 & DDTree & 100k & 2 & 16 & $6\!\times\!10^{-4}$ & 2048 & 1,7,13,19,25 & -- \\
 & \textbf{Ours} & 100k & 2 & 16 & $6\!\times\!10^{-4}$ & 2048 & 1,7,13,19,25 & -- \\
\midrule
\multirow{5}{*}{SVGen-7B} & EAGLE-3 & 100k & 2 & 16 & $1\!\times\!10^{-4}$ & 2048 & -- & -- \\
 & DFlash & 100k & 2 & 16 & $6\!\times\!10^{-4}$ & 2048 & 1,7,13,19,25 & -- \\
 & DARTree & 100k & 2 & 16 & $6\!\times\!10^{-4}$ & 2048 & 1,7,13,19,25 & -- \\
 & DDTree & 100k & 2 & 16 & $6\!\times\!10^{-4}$ & 2048 & 1,7,13,19,25 & -- \\
 & \textbf{Ours} & 100k & 2 & 16 & $6\!\times\!10^{-4}$ & 2048 & 1,7,13,19,25 & -- \\
\midrule
\multirow{5}{*}{HiVG-3B} & EAGLE-3 & 100k & 2 & 16 & $1\!\times\!10^{-4}$ & 2048 & -- & -- \\
 & DFlash & 100k & 2 & 16 & $6\!\times\!10^{-4}$ & 2048 & 1,9,17,25,33 & -- \\
 & DARTree & 100k & 2 & 16 & $6\!\times\!10^{-4}$ & 2048 & 1,9,17,25,33 & -- \\
 & DDTree & 100k & 2 & 16 & $6\!\times\!10^{-4}$ & 2048 & 1,9,17,25,33 & -- \\
 & \textbf{Ours} & 100k & 2 & 16 & $6\!\times\!10^{-4}$ & 2048 & 1,9,17,25,33 & -- \\
\midrule
\multirow{5}{*}{OmniLottie-3B} & EAGLE-3 & 100k & 2 & 16 & $1\!\times\!10^{-4}$ & 4608 & -- & -- \\
 & DFlash & 100k & 2 & 16 & $6\!\times\!10^{-4}$ & 4608 & 1,9,17,25,33 & -- \\
 & DARTree & 100k & 2 & 16 & $6\!\times\!10^{-4}$ & 4608 & 1,9,17,25,33 & -- \\
 & DDTree & 100k & 2 & 16 & $6\!\times\!10^{-4}$ & 4608 & 1,9,17,25,33 & -- \\
 & \textbf{Ours} & 100k & 2 & 16 & $6\!\times\!10^{-4}$ & 4608 & 1,9,17,25,33 & -- \\
\midrule
\multirow{5}{*}{StarVector-1B} & EAGLE-3 & 100k & 2 & 16 & $1\!\times\!10^{-4}$ & 4096 & -- & -- \\
 & DFlash & 100k & 2 & 16 & $6\!\times\!10^{-4}$ & 4096 & 1,6,11,16,21 & -- \\
 & DARTree & 100k & 2 & 16 & $6\!\times\!10^{-4}$ & 4096 & 1,6,11,16,21 & -- \\
 & DDTree & 100k & 2 & 16 & $6\!\times\!10^{-4}$ & 4096 & 1,6,11,16,21 & -- \\
 & \textbf{Ours} & 100k & 2 & 16 & $6\!\times\!10^{-4}$ & 4096 & 1,6,11,16,21 & -- \\
\midrule
\multirow{5}{*}{StarVector-8B} & EAGLE-3 & 100k & 2 & 16 & $1\!\times\!10^{-4}$ & 4096 & -- & -- \\
 & DFlash & 100k & 2 & 16 & $6\!\times\!10^{-4}$ & 4096 & 1,8,15,22,29 & -- \\
 & DARTree & 100k & 2 & 16 & $6\!\times\!10^{-4}$ & 4096 & 1,8,15,22,29 & -- \\
 & DDTree & 100k & 2 & 16 & $6\!\times\!10^{-4}$ & 4096 & 1,8,15,22,29 & -- \\
 & \textbf{Ours} & 100k & 2 & 16 & $6\!\times\!10^{-4}$ & 4096 & 1,8,15,22,29 & -- \\
\bottomrule
\end{tabular*}
\end{table*}

\subsection{Timing}
\label{app:timing}
Apart from the serving experiment, every timing runs on one H200 in bf16 with eager attention.
A speedup divides the original model's time per token, measured in the same job on the same prompts, by the decoder's own.
The time of the best of $N$ includes drawing, rendering and scoring all samples.
Losslessness holds in exact arithmetic, and two bf16 forward passes of different shapes can break a near-tie differently.
The fp32 audits of Table~\ref{app:fp32} separate these ties from errors.
Table~\ref{tab:quality} combines runs from several nodes.
Each job that added rows after the main runs also reran an existing row as a stopwatch.
The ratio of the two timings of that row scales the added rows to the node of the main runs (Table~\ref{app:node-factors}).
The factors lie between 0.90 and 1.26.
The original model's stopwatch varies most, with factors of 1.15 to 1.26 on four shards.
Its outputs there match the existing ones token for token, so the spread reflects the speed of the nodes.
Without the factors, no ranking in Table~\ref{tab:quality} changes, and the released data list every factor.

\begin{table*}[t]
\caption{\textbf{Node factors.} Each factor divides the existing row's mean seconds by the stopwatch's, per job and subset, over the prompts whose tokens the stopwatch reproduces exactly. Ranges are over jobs.}
\label{app:node-factors}
\centering
\footnotesize
\begin{tabular}{@{}lllccc@{}}
\toprule
Model and task & Stopwatch & Rows it scales & Jobs & Icon & Illustration \\
\midrule
\multirow{5}{*}{\shortstack[l]{OmniSVG-8B\\text-to-SVG}} & DDTree & DFlash, same outputs and loop control & 1 & 1.02 & 1.03 \\
 & EAGLE-3 & EAGLE-3, loop control & 1 & 1.06 & 1.05 \\
 & DARTree & DARTree, loop control & 1 & 1.04 & 1.03 \\
 & AR & AR, loop control & 2 & 1.14--1.25 & 1.08--1.26 \\
 & Ours, best of 8 & AR and DFlash, best of 8 & 5 & 0.96--1.06 & 0.96--1.07 \\
\midrule
\multirow{5}{*}{\shortstack[l]{InternSVG-8B\\text-to-SVG}} & DDTree & DFlash, same outputs and loop control & 2 & 0.96--1.02 & 0.99--1.00 \\
 & EAGLE-3 & EAGLE-3, loop control & 3 & 1.07--1.08 & 1.08--1.09 \\
 & DARTree & DARTree, loop control & 2 & 1.02--1.03 & 1.01 \\
 & AR & AR, loop control & 5 & 1.03--1.19 & 1.02--1.17 \\
 & Ours, best of 8 & AR and DFlash, best of 8 & 14 & 0.97--1.05 & 0.94--1.02 \\
\midrule
\multirow{5}{*}{\shortstack[l]{OmniSVG-8B\\image-to-SVG}} & DDTree & DFlash, same outputs and loop control & 2 & 0.94--0.97 & 0.95--0.98 \\
 & EAGLE-3 & EAGLE-3, loop control & 2 & 1.08--1.09 & 1.08--1.09 \\
 & DARTree & DARTree, loop control & 1 & 0.99 & 1.00 \\
 & AR & AR, loop control & 3 & 0.99--1.13 & 1.00--1.14 \\
 & Ours, best of 8 & AR and DFlash, best of 8 & 7 & 1.01--1.12 & 1.01--1.12 \\
\midrule
\multirow{5}{*}{\shortstack[l]{InternSVG-8B\\image-to-SVG}} & DDTree & DFlash, same outputs and loop control & 3 & 0.96--1.02 & 0.97--1.02 \\
 & EAGLE-3 & EAGLE-3, loop control & 3 & 1.01--1.04 & 1.02--1.04 \\
 & DARTree & DARTree, loop control & 2 & 1.00--1.03 & 1.01--1.04 \\
 & AR & AR, loop control & 5 & 0.97--1.14 & 0.95--1.19 \\
 & Ours, best of 8 & AR and DFlash, best of 8 & 16 & 0.90--1.01 & 0.91--1.01 \\
\bottomrule
\end{tabular}
\end{table*}

\subsection{Seeds and Temperatures}
\label{app:variation}
Sampled runs use one seed per prompt.
OmniSVG-8B image-to-SVG reruns at temperatures 0.6 and 1.0 besides its released 0.3 (Table~\ref{app:temperature}), and two model--task pairs rerun at three seeds (Table~\ref{app:seeds}).
The order of the methods holds at every temperature and seed.

\begin{table}[htbp]\centering
\begingroup
\renewenvironment{table}[1][]{\begin{minipage}[t]{0.485\textwidth}\vspace{0pt}\centering}{\end{minipage}}%
\renewenvironment{table*}[1][]{\begin{minipage}[t]{0.485\textwidth}\vspace{0pt}\centering}{\end{minipage}}%
\begin{table}[t]
\caption{\textbf{Sampling temperature.} OmniSVG-8B image-to-SVG, MMSVG-Bench, official top-$p$, the prompts that every method finished at that temperature. Speedup is the mean over prompts against the original model timed in the same job. \textbf{Best} in bold, \underline{second}{} underlined, per row.}
\label{app:temperature}
\centering
\footnotesize
\setlength{\tabcolsep}{1.4pt}
\begin{tabular*}{\columnwidth}{@{\extracolsep{\fill}}lr*{6}{c}@{}}
\toprule
 & & \multicolumn{3}{c}{$\tau$} & \multicolumn{3}{c}{speedup} \\
\cmidrule(lr){3-5}\cmidrule(lr){6-8}
Temperature & prompts & \textbf{Ours} & DDTree & DFlash & \textbf{Ours} & DDTree & DFlash \\
\midrule
$T$ 0.3 (official) & 300 & \textbf{13.12} & \underline{10.09} & 7.49 & \textbf{6.49}$\times$ & \underline{5.42}$\times$ & 4.58$\times$ \\
$T$ 0.6 & 300 & \textbf{11.85} & \underline{9.64} & 7.21 & \textbf{5.99}$\times$ & \underline{5.19}$\times$ & 4.31$\times$ \\
$T$ 1.0 & 300 & \textbf{10.26} & \underline{8.74} & 6.16 & \textbf{5.48}$\times$ & \underline{4.90}$\times$ & 3.99$\times$ \\
\bottomrule
\end{tabular*}
\end{table}
\hfill\begin{table}[t]
\caption{\textbf{Sampling seed.} Each model's official sampling, three seeds, the same 300 MMSVG-Bench prompts. DDTree runs in the same job as ours and shares its timing of the original model. Our speedup varies by 9\% to 14\% across seeds and stays ahead of DDTree at every seed.}
\label{app:seeds}
\centering
\footnotesize
\setlength{\tabcolsep}{2.0pt}
\begin{tabular*}{\columnwidth}{@{\extracolsep{\fill}}lr*{5}{c}@{}}
\toprule
 & & \multicolumn{2}{c}{$\tau$} & \multicolumn{2}{c}{speedup} & \\
\cmidrule(lr){3-4}\cmidrule(lr){5-6}
Seed & prompts & \textbf{Ours} & DDTree & \textbf{Ours} & DDTree & Ours / DDTree \\
\midrule
\multicolumn{7}{l}{\textit{OmniSVG-8B, image-to-SVG}} \\
1 & 300 & 13.12 & 10.09 & 6.49$\times$ & 5.42$\times$ & 1.20 \\
2 & 300 & 12.56 & 10.13 & 6.16$\times$ & 4.95$\times$ & 1.24 \\
3 & 300 & 12.86 & 10.22 & 5.95$\times$ & 5.12$\times$ & 1.16 \\
\addlinespace[2pt]
\multicolumn{7}{l}{\textit{InternSVG-8B, text-to-SVG}} \\
1 & 300 & 9.18 & 8.53 & 6.44$\times$ & 5.40$\times$ & 1.19 \\
2 & 300 & 9.64 & 8.53 & 6.99$\times$ & 5.89$\times$ & 1.19 \\
3 & 300 & 9.13 & 8.55 & 6.09$\times$ & 5.35$\times$ & 1.14 \\
\bottomrule
\end{tabular*}
\end{table}
\endgroup
\end{table}

\subsection{Metrics, Judges and Scores}
\label{app:metrics}
Quality follows the official metrics of SArena (Appendix~\ref{app:full}).
For text-to-SVG, FID and FID-C compare the renderings with the reference renderings on Inception-v3 and CLIP ViT-B/32 features.
CLIP-T2I and CLIP-I2I score each rendering against its prompt and against its reference rendering with CLIP ViT-L/14.
For image-to-SVG, DINO similarity, LPIPS, SSIM and PSNR compare each rendering with the input image.
For these metrics, every output is rendered by the official rasterizer of SArena, and an output that does not render counts as a black image.
Before extraction, the open tags of an InternSVG-8B output are closed for every decoder alike, since loop control and the length limit can end an output inside a tag.
Rendered selection scores a candidate by the CLIP ViT-B/32 similarity of its rendering to the prompt, or by the DINOv2 similarity of its rendering to the input image.
The aesthetic score is the LAION predictor on CLIP ViT-L/14 features, and HPS is HPS v2, both as in the official metrics of OmniSVG.

The three VLM judges of Figure~\ref{fig:ex2}j are claude-opus-5-5 (Anthropic), gpt-6.1-sol (OpenAI) and gemini-3.8-flash (Google).
Each judge sees the renderings of two outputs for one prompt.
It picks the better one, first by how well it matches the prompt and then by visual quality, or calls a tie.
Every pair is judged in both orders, a tie counts half, and intervals are 95\% bootstrap intervals over prompts.

\FloatBarrier
\section{Main Results in Full}
\label{app:full}
This section gives the numbers behind Section~\ref{sec:main}.

\subsection{Speed in All 60 Settings}
\label{app:speed-all}
\begin{figure*}[t]
\centering
\includegraphics{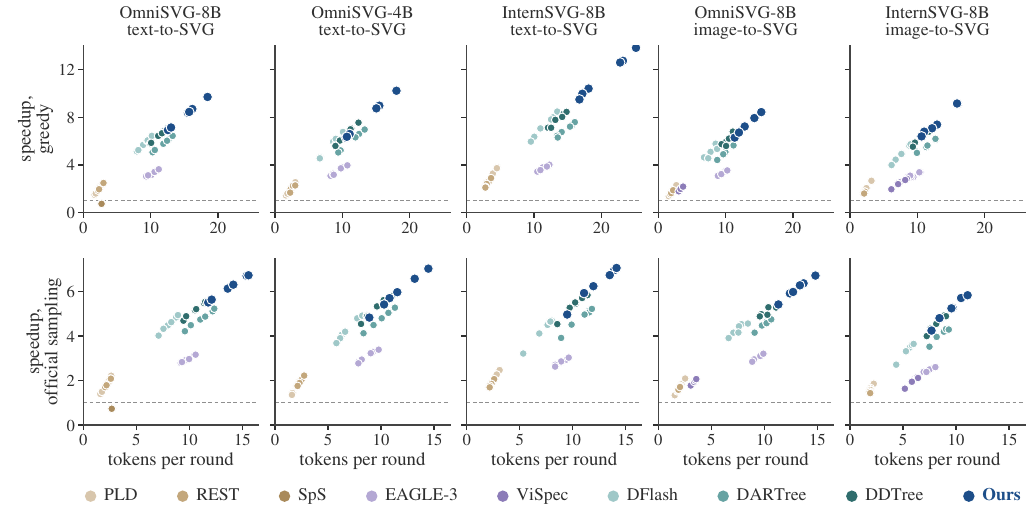}
\caption{\textbf{Tokens per round against speedup, pair by pair.} Figure~\ref{fig:main}b split by model--task pair, under greedy decoding (top) and official sampling (bottom), one point per decoder and benchmark. The dotted line marks the original model.}
\label{app:tau-pairs}
\end{figure*}

Tables~\ref{app:t1-greedy} and~\ref{app:t1-sample} give the speedup and the tokens per round of every decoder in each of the 60 settings of Figure~\ref{fig:main}, and Figure~\ref{app:tau-pairs} plots the two against each other pair by pair.
On loop-free content alone, Sightline also writes the most tokens per round in every setting (Table~\ref{app:t1-healthy}).
On OmniSVG-8B, the slowest tenth of our outputs still decodes 2.6 to 2.8 times as fast as the original model (Table~\ref{app:hard-rows}).

\begin{table*}[t]
\caption{\textbf{Figure~\ref{fig:main} in full, greedy decoding.} For every benchmark, the wall-clock speedup over the original model timed on the same H200 (Sp.) and the tokens per round $\tau$, counting all tokens. Mean is over the six benchmarks. $^\dagger$Held-out captions of the drafts' training source. SpS needs a smaller model of the same family, which only OmniSVG-8B has. ViSpec is a vision language method and runs on image-to-SVG. \textbf{Best} in bold, \underline{second}{} underlined, within each model.}
\label{app:t1-greedy}
\centering
\footnotesize
\setlength{\tabcolsep}{1.5pt}
\begin{tabular*}{\textwidth}{@{\extracolsep{\fill}}l*{14}{c}@{}}
\toprule
 & \multicolumn{4}{c}{MMSVG} & \multicolumn{4}{c}{MMSVG-2M$^\dagger$} & \multicolumn{4}{c}{SArena} & \multicolumn{2}{c}{Mean} \\
\cmidrule(lr){2-5}\cmidrule(lr){6-9}\cmidrule(lr){10-13}\cmidrule(lr){14-15}
 & \multicolumn{2}{c}{Icon} & \multicolumn{2}{c}{Illus.} & \multicolumn{2}{c}{Icon} & \multicolumn{2}{c}{Illus.} & \multicolumn{2}{c}{Icon} & \multicolumn{2}{c}{Illus.} & \multicolumn{2}{c}{} \\
\cmidrule(lr){2-3}\cmidrule(lr){4-5}\cmidrule(lr){6-7}\cmidrule(lr){8-9}\cmidrule(lr){10-11}\cmidrule(lr){12-13}
Method & Sp. & $\tau$ & Sp. & $\tau$ & Sp. & $\tau$ & Sp. & $\tau$ & Sp. & $\tau$ & Sp. & $\tau$ & Sp. & $\tau$ \\
\midrule
\multicolumn{15}{l}{\textit{OmniSVG-8B, text-to-SVG}} \\
PLD & 1.97$\times$ & 2.26 & 2.15$\times$ & 2.46 & 1.49$\times$ & 1.68 & 2.07$\times$ & 2.37 & 1.60$\times$ & 1.85 & 2.32$\times$ & 2.69 & 1.93$\times$ & 2.22 \\
REST & 2.03$\times$ & 2.43 & 2.14$\times$ & 2.56 & 1.85$\times$ & 2.20 & 2.07$\times$ & 2.47 & 1.95$\times$ & 2.34 & 2.48$\times$ & 2.99 & 2.09$\times$ & 2.50 \\
SpS & 0.74$\times$ & 2.73 & 0.72$\times$ & 2.69 & 0.73$\times$ & 2.70 & 0.73$\times$ & 2.71 & 0.74$\times$ & 2.72 & 0.73$\times$ & 2.71 & 0.73$\times$ & 2.71 \\
EAGLE-3 & 3.17$\times$ & 10.12 & 3.51$\times$ & 10.89 & 3.04$\times$ & 9.34 & 3.41$\times$ & 10.58 & 3.12$\times$ & 9.62 & 3.63$\times$ & 11.21 & 3.31$\times$ & 10.29 \\
DFlash & 5.67$\times$ & 8.88 & 6.02$\times$ & 9.46 & 5.14$\times$ & 7.96 & 6.06$\times$ & 9.55 & 5.24$\times$ & 8.17 & 6.44$\times$ & 10.18 & 5.76$\times$ & 9.03 \\
DARTree & 5.77$\times$ & \underline{11.88} & 6.13$\times$ & \underline{12.67} & 5.06$\times$ & \underline{10.29} & 6.03$\times$ & \underline{12.44} & 5.25$\times$ & \underline{10.62} & 6.44$\times$ & \underline{13.27} & 5.78$\times$ & \underline{11.86} \\
DDTree & \underline{6.44}$\times$ & 11.12 & \underline{6.82}$\times$ & 11.96 & \underline{5.81}$\times$ & 9.94 & \underline{6.66}$\times$ & 11.68 & \underline{5.85}$\times$ & 10.07 & \underline{7.12}$\times$ & 12.59 & \underline{6.45}$\times$ & 11.23 \\
\textbf{Ours} & \textbf{8.37}$\times$ & \textbf{15.53} & \textbf{8.70}$\times$ & \textbf{16.19} & \textbf{6.92}$\times$ & \textbf{12.57} & \textbf{8.45}$\times$ & \textbf{15.73} & \textbf{7.14}$\times$ & \textbf{13.03} & \textbf{9.71}$\times$ & \textbf{18.46} & \textbf{8.21}$\times$ & \textbf{15.25} \\
\midrule
\multicolumn{15}{l}{\textit{OmniSVG-4B, text-to-SVG}} \\
PLD & 1.50$\times$ & 1.71 & 2.10$\times$ & 2.41 & 1.44$\times$ & 1.63 & 2.14$\times$ & 2.45 & 1.59$\times$ & 1.85 & 2.54$\times$ & 3.02 & 1.88$\times$ & 2.18 \\
REST & 2.13$\times$ & 2.55 & 2.26$\times$ & 2.72 & 1.97$\times$ & 2.34 & 2.23$\times$ & 2.67 & 1.67$\times$ & 2.25 & 2.27$\times$ & 3.00 & 2.09$\times$ & 2.59 \\
EAGLE-3 & 3.08$\times$ & 8.26 & 3.80$\times$ & 10.04 & 3.25$\times$ & 8.62 & 3.71$\times$ & 9.79 & 3.17$\times$ & 8.72 & 3.96$\times$ & 10.75 & 3.50$\times$ & 9.36 \\
DFlash & 4.65$\times$ & 6.76 & 5.95$\times$ & 8.63 & 4.63$\times$ & 6.67 & 6.19$\times$ & 9.03 & 4.55$\times$ & 6.66 & 6.76$\times$ & 10.10 & 5.45$\times$ & 7.98 \\
DARTree & 5.24$\times$ & \underline{9.77} & 6.60$\times$ & \underline{12.49} & 5.22$\times$ & \underline{9.74} & 6.31$\times$ & \underline{11.96} & 5.05$\times$ & \underline{9.40} & 6.97$\times$ & \underline{13.32} & 5.90$\times$ & \underline{11.11} \\
DDTree & \underline{6.04}$\times$ & 9.64 & \underline{7.06}$\times$ & 11.34 & \underline{5.61}$\times$ & 8.90 & \underline{6.99}$\times$ & 11.22 & \underline{5.59}$\times$ & 8.98 & \underline{7.55}$\times$ & 12.40 & \underline{6.47}$\times$ & 10.41 \\
\textbf{Ours} & \textbf{6.59}$\times$ & \textbf{11.09} & \textbf{8.97}$\times$ & \textbf{15.50} & \textbf{6.28}$\times$ & \textbf{10.54} & \textbf{8.75}$\times$ & \textbf{15.06} & \textbf{6.36}$\times$ & \textbf{10.68} & \textbf{10.23}$\times$ & \textbf{18.04} & \textbf{7.86}$\times$ & \textbf{13.49} \\
\midrule
\multicolumn{15}{l}{\textit{InternSVG-8B, text-to-SVG}} \\
PLD & 3.27$\times$ & 3.89 & 3.80$\times$ & 4.53 & 2.51$\times$ & 2.90 & 3.72$\times$ & 4.49 & 2.52$\times$ & 3.04 & 2.78$\times$ & 3.49 & 3.10$\times$ & 3.72 \\
REST & 2.62$\times$ & 3.31 & 2.89$\times$ & 3.62 & 2.28$\times$ & 2.81 & 2.39$\times$ & 2.96 & 2.16$\times$ & 2.79 & 2.10$\times$ & 2.78 & 2.40$\times$ & 3.05 \\
EAGLE-3 & 3.86$\times$ & 12.15 & 4.01$\times$ & 12.29 & 3.78$\times$ & 11.28 & 3.86$\times$ & 11.92 & 3.44$\times$ & 10.50 & 3.56$\times$ & 10.95 & 3.75$\times$ & 11.52 \\
DFlash & 8.00$\times$ & 12.84 & \underline{8.48}$\times$ & 13.46 & 7.07$\times$ & 10.95 & 7.81$\times$ & 12.54 & 6.36$\times$ & 10.04 & 5.95$\times$ & 9.54 & 7.28$\times$ & 11.56 \\
DARTree & 7.35$\times$ & \underline{15.69} & 7.59$\times$ & \underline{16.06} & 6.76$\times$ & \underline{14.11} & 7.22$\times$ & \underline{15.36} & 6.44$\times$ & \underline{13.43} & 6.30$\times$ & \underline{13.53} & 6.94$\times$ & \underline{14.70} \\
DDTree & \underline{8.28}$\times$ & 14.39 & 8.45$\times$ & 14.84 & \underline{7.78}$\times$ & 13.18 & \underline{8.03}$\times$ & 14.19 & \underline{7.12}$\times$ & 12.08 & \underline{7.12}$\times$ & 12.54 & \underline{7.80}$\times$ & 13.54 \\
\textbf{Ours} & \textbf{12.74}$\times$ & \textbf{23.22} & \textbf{13.83}$\times$ & \textbf{25.16} & \textbf{10.41}$\times$ & \textbf{18.11} & \textbf{12.60}$\times$ & \textbf{22.80} & \textbf{9.98}$\times$ & \textbf{17.19} & \textbf{9.50}$\times$ & \textbf{16.75} & \textbf{11.51}$\times$ & \textbf{20.54} \\
\midrule
\multicolumn{15}{l}{\textit{OmniSVG-8B, image-to-SVG}} \\
PLD & 1.61$\times$ & 1.85 & 1.94$\times$ & 2.23 & 1.37$\times$ & 1.56 & 1.76$\times$ & 2.02 & 1.73$\times$ & 1.99 & 2.30$\times$ & 2.68 & 1.79$\times$ & 2.05 \\
REST & 1.53$\times$ & 1.79 & 1.72$\times$ & 2.01 & 1.59$\times$ & 1.87 & 1.53$\times$ & 1.79 & 1.63$\times$ & 1.92 & 1.87$\times$ & 2.20 & 1.65$\times$ & 1.93 \\
EAGLE-3 & 3.15$\times$ & 9.21 & 3.46$\times$ & 9.92 & 3.08$\times$ & 8.85 & 3.26$\times$ & 9.51 & 3.23$\times$ & 9.36 & 3.53$\times$ & 10.24 & 3.29$\times$ & 9.52 \\
ViSpec & 1.93$\times$ & 3.22 & 2.08$\times$ & 3.52 & 1.78$\times$ & 3.03 & 2.01$\times$ & 3.36 & 1.98$\times$ & 3.31 & 2.18$\times$ & 3.68 & 1.99$\times$ & 3.36 \\
DFlash & 4.65$\times$ & 7.10 & 5.78$\times$ & 8.51 & 4.63$\times$ & 6.79 & 5.09$\times$ & 7.76 & 4.56$\times$ & 7.41 & 5.34$\times$ & 8.74 & 5.01$\times$ & 7.72 \\
DARTree & 4.97$\times$ & 9.80 & 5.61$\times$ & \underline{11.07} & 4.41$\times$ & 8.75 & 5.08$\times$ & 9.96 & 4.89$\times$ & 9.65 & 5.64$\times$ & 11.17 & 5.10$\times$ & 10.07 \\
DDTree & \underline{5.89}$\times$ & \underline{9.90} & \underline{6.79}$\times$ & \underline{11.07} & \underline{5.74}$\times$ & \underline{9.39} & \underline{6.21}$\times$ & \underline{10.49} & \underline{5.59}$\times$ & \underline{10.07} & \underline{6.32}$\times$ & \underline{11.42} & \underline{6.09}$\times$ & \underline{10.39} \\
\textbf{Ours} & \textbf{6.79}$\times$ & \textbf{12.09} & \textbf{7.94}$\times$ & \textbf{14.27} & \textbf{6.29}$\times$ & \textbf{11.34} & \textbf{7.23}$\times$ & \textbf{12.83} & \textbf{6.72}$\times$ & \textbf{11.98} & \textbf{8.44}$\times$ & \textbf{15.30} & \textbf{7.23}$\times$ & \textbf{12.97} \\
\midrule
\multicolumn{15}{l}{\textit{InternSVG-8B, image-to-SVG}} \\
PLD & 2.07$\times$ & 2.43 & 2.64$\times$ & 3.15 & 1.86$\times$ & 2.16 & 2.67$\times$ & 3.18 & 1.98$\times$ & 2.30 & 2.05$\times$ & 2.44 & 2.21$\times$ & 2.61 \\
REST & 1.65$\times$ & 2.07 & 1.68$\times$ & 2.14 & 1.75$\times$ & 2.18 & 1.74$\times$ & 2.22 & 1.59$\times$ & 1.98 & 1.59$\times$ & 2.10 & 1.67$\times$ & 2.12 \\
EAGLE-3 & 2.97$\times$ & 9.41 & 3.38$\times$ & 10.45 & 2.98$\times$ & 9.10 & 3.38$\times$ & 10.30 & 3.00$\times$ & 8.61 & 3.09$\times$ & 8.94 & 3.13$\times$ & 9.47 \\
ViSpec & 2.53$\times$ & 7.37 & 2.83$\times$ & 8.45 & 2.44$\times$ & 7.06 & 2.73$\times$ & 8.20 & 2.38$\times$ & 7.15 & 1.95$\times$ & 6.14 & 2.48$\times$ & 7.40 \\
DFlash & 4.96$\times$ & 7.69 & 5.62$\times$ & 8.81 & 4.92$\times$ & 7.63 & 5.73$\times$ & 8.99 & 4.44$\times$ & 6.78 & 3.99$\times$ & 6.18 & 4.94$\times$ & 7.68 \\
DARTree & 5.49$\times$ & \underline{11.35} & 6.11$\times$ & \underline{12.74} & 5.62$\times$ & \underline{11.51} & 6.19$\times$ & \underline{12.70} & 5.12$\times$ & \underline{9.88} & 5.01$\times$ & \underline{9.94} & 5.59$\times$ & \underline{11.35} \\
DDTree & \underline{6.44}$\times$ & 10.93 & \underline{6.75}$\times$ & 11.66 & \underline{6.32}$\times$ & 10.64 & \underline{6.77}$\times$ & 11.67 & \underline{5.87}$\times$ & 9.65 & \underline{5.53}$\times$ & 9.34 & \underline{6.28}$\times$ & 10.65 \\
\textbf{Ours} & \textbf{7.38}$\times$ & \textbf{12.93} & \textbf{9.12}$\times$ & \textbf{16.02} & \textbf{7.07}$\times$ & \textbf{12.19} & \textbf{9.16}$\times$ & \textbf{15.90} & \textbf{6.79}$\times$ & \textbf{11.03} & \textbf{6.39}$\times$ & \textbf{10.63} & \textbf{7.65}$\times$ & \textbf{13.12} \\
\bottomrule
\end{tabular*}
\end{table*}

\begin{table*}[t]
\caption{\textbf{Figure~\ref{fig:main} in full, each model's official sampling.} For every benchmark, the wall-clock speedup over the original model timed on the same H200 (Sp.) and the tokens per round $\tau$, counting all tokens. Mean is over the six benchmarks. $^\dagger$Held-out captions of the drafts' training source. SpS needs a smaller model of the same family, which only OmniSVG-8B has. ViSpec is a vision language method and runs on image-to-SVG. \textbf{Best} in bold, \underline{second}{} underlined, within each model.}
\label{app:t1-sample}
\centering
\footnotesize
\setlength{\tabcolsep}{1.5pt}
\begin{tabular*}{\textwidth}{@{\extracolsep{\fill}}l*{14}{c}@{}}
\toprule
 & \multicolumn{4}{c}{MMSVG} & \multicolumn{4}{c}{MMSVG-2M$^\dagger$} & \multicolumn{4}{c}{SArena} & \multicolumn{2}{c}{Mean} \\
\cmidrule(lr){2-5}\cmidrule(lr){6-9}\cmidrule(lr){10-13}\cmidrule(lr){14-15}
 & \multicolumn{2}{c}{Icon} & \multicolumn{2}{c}{Illus.} & \multicolumn{2}{c}{Icon} & \multicolumn{2}{c}{Illus.} & \multicolumn{2}{c}{Icon} & \multicolumn{2}{c}{Illus.} & \multicolumn{2}{c}{} \\
\cmidrule(lr){2-3}\cmidrule(lr){4-5}\cmidrule(lr){6-7}\cmidrule(lr){8-9}\cmidrule(lr){10-11}\cmidrule(lr){12-13}
Method & Sp. & $\tau$ & Sp. & $\tau$ & Sp. & $\tau$ & Sp. & $\tau$ & Sp. & $\tau$ & Sp. & $\tau$ & Sp. & $\tau$ \\
\midrule
\multicolumn{15}{l}{\textit{OmniSVG-8B, text-to-SVG}} \\
PLD & 1.81$\times$ & 2.14 & 2.22$\times$ & 2.63 & 1.40$\times$ & 1.60 & 1.88$\times$ & 2.22 & 1.50$\times$ & 1.75 & 2.08$\times$ & 2.54 & 1.81$\times$ & 2.15 \\
REST & 1.77$\times$ & 2.15 & 1.97$\times$ & 2.43 & 1.71$\times$ & 2.06 & 1.85$\times$ & 2.26 & 1.79$\times$ & 2.19 & 2.08$\times$ & 2.60 & 1.86$\times$ & 2.28 \\
SpS & 0.74$\times$ & 2.71 & 0.73$\times$ & 2.66 & 0.73$\times$ & 2.66 & 0.75$\times$ & 2.74 & 0.74$\times$ & 2.69 & 0.73$\times$ & 2.68 & 0.74$\times$ & 2.69 \\
EAGLE-3 & 2.92$\times$ & 9.65 & 3.16$\times$ & 10.48 & 2.79$\times$ & 9.19 & 2.97$\times$ & 9.97 & 2.83$\times$ & 9.30 & 3.16$\times$ & 10.59 & 2.97$\times$ & 9.86 \\
DFlash & 4.49$\times$ & 7.97 & 4.84$\times$ & 8.75 & 4.02$\times$ & 7.10 & 4.63$\times$ & 8.29 & 4.33$\times$ & 7.55 & 4.95$\times$ & 8.92 & 4.54$\times$ & 8.10 \\
DARTree & 4.75$\times$ & \underline{11.03} & 5.12$\times$ & \underline{12.19} & 4.22$\times$ & \underline{9.58} & 4.88$\times$ & \underline{11.49} & 4.49$\times$ & \underline{10.14} & 5.23$\times$ & \underline{12.35} & 4.78$\times$ & \underline{11.13} \\
DDTree & \underline{5.17}$\times$ & 10.53 & \underline{5.49}$\times$ & 11.38 & \underline{4.69}$\times$ & 9.44 & \underline{5.21}$\times$ & 10.64 & \underline{4.90}$\times$ & 9.70 & \underline{5.55}$\times$ & 11.49 & \underline{5.17}$\times$ & 10.53 \\
\textbf{Ours} & \textbf{6.14}$\times$ & \textbf{13.62} & \textbf{6.71}$\times$ & \textbf{15.40} & \textbf{5.51}$\times$ & \textbf{11.78} & \textbf{6.32}$\times$ & \textbf{14.14} & \textbf{5.64}$\times$ & \textbf{12.10} & \textbf{6.74}$\times$ & \textbf{15.57} & \textbf{6.18}$\times$ & \textbf{13.77} \\
\midrule
\multicolumn{15}{l}{\textit{OmniSVG-4B, text-to-SVG}} \\
PLD & 1.46$\times$ & 1.64 & 1.93$\times$ & 2.24 & 1.44$\times$ & 1.64 & 1.77$\times$ & 2.04 & 1.36$\times$ & 1.60 & 2.03$\times$ & 2.48 & 1.66$\times$ & 1.94 \\
REST & 1.73$\times$ & 2.10 & 2.04$\times$ & 2.51 & 1.81$\times$ & 2.21 & 1.89$\times$ & 2.31 & 1.76$\times$ & 2.13 & 2.22$\times$ & 2.75 & 1.91$\times$ & 2.33 \\
EAGLE-3 & 2.84$\times$ & 7.85 & 3.29$\times$ & 9.22 & 2.94$\times$ & 8.18 & 3.23$\times$ & 9.03 & 2.78$\times$ & 7.86 & 3.39$\times$ & 9.78 & 3.08$\times$ & 8.65 \\
DFlash & 4.06$\times$ & 6.31 & 4.80$\times$ & 7.75 & 3.91$\times$ & 6.14 & 4.20$\times$ & 6.61 & 3.68$\times$ & 5.79 & 4.92$\times$ & 8.22 & 4.26$\times$ & 6.80 \\
DARTree & 4.52$\times$ & \underline{9.27} & 5.05$\times$ & \underline{10.69} & 4.50$\times$ & \underline{9.29} & 4.80$\times$ & \underline{10.06} & 4.13$\times$ & \underline{8.35} & 5.28$\times$ & \underline{11.31} & 4.71$\times$ & \underline{9.83} \\
DDTree & \underline{4.92}$\times$ & 8.64 & \underline{5.62}$\times$ & 10.25 & \underline{4.89}$\times$ & 8.67 & \underline{5.34}$\times$ & 9.62 & \underline{4.55}$\times$ & 8.11 & \underline{5.82}$\times$ & 11.08 & \underline{5.19}$\times$ & 9.39 \\
\textbf{Ours} & \textbf{5.71}$\times$ & \textbf{10.80} & \textbf{6.58}$\times$ & \textbf{13.17} & \textbf{5.42}$\times$ & \textbf{10.28} & \textbf{5.98}$\times$ & \textbf{11.53} & \textbf{4.83}$\times$ & \textbf{8.89} & \textbf{7.04}$\times$ & \textbf{14.46} & \textbf{5.93}$\times$ & \textbf{11.52} \\
\midrule
\multicolumn{15}{l}{\textit{InternSVG-8B, text-to-SVG}} \\
PLD & 2.15$\times$ & 2.68 & 2.47$\times$ & 3.11 & 1.70$\times$ & 2.09 & 2.27$\times$ & 2.83 & 1.89$\times$ & 2.31 & 1.73$\times$ & 2.22 & 2.04$\times$ & 2.54 \\
REST & 1.77$\times$ & 2.19 & 1.80$\times$ & 2.26 & 2.06$\times$ & 2.59 & 1.85$\times$ & 2.33 & 1.76$\times$ & 2.19 & 1.69$\times$ & 2.17 & 1.82$\times$ & 2.29 \\
EAGLE-3 & 2.83$\times$ & 8.93 & 2.95$\times$ & 9.40 & 2.86$\times$ & 8.91 & 3.03$\times$ & 9.62 & 2.71$\times$ & 8.34 & 2.63$\times$ & 8.34 & 2.83$\times$ & 8.92 \\
DFlash & 4.51$\times$ & 7.63 & 4.69$\times$ & 8.03 & 4.66$\times$ & 7.83 & 4.66$\times$ & 7.90 & 4.12$\times$ & 6.85 & 3.22$\times$ & 5.33 & 4.31$\times$ & 7.26 \\
DARTree & 5.14$\times$ & \underline{11.53} & 5.07$\times$ & \underline{11.57} & 4.97$\times$ & \underline{11.12} & 5.22$\times$ & \underline{11.81} & 4.52$\times$ & \underline{9.89} & 3.92$\times$ & \underline{8.90} & 4.81$\times$ & \underline{10.80} \\
DDTree & \underline{5.45}$\times$ & 10.28 & \underline{5.85}$\times$ & 11.39 & \underline{5.51}$\times$ & 10.23 & \underline{5.71}$\times$ & 10.88 & \underline{5.28}$\times$ & 9.71 & \underline{4.54}$\times$ & 8.55 & \underline{5.39}$\times$ & 10.17 \\
\textbf{Ours} & \textbf{6.96}$\times$ & \textbf{13.90} & \textbf{6.75}$\times$ & \textbf{13.49} & \textbf{6.25}$\times$ & \textbf{11.95} & \textbf{7.07}$\times$ & \textbf{14.13} & \textbf{5.94}$\times$ & \textbf{11.07} & \textbf{4.97}$\times$ & \textbf{9.48} & \textbf{6.32}$\times$ & \textbf{12.34} \\
\midrule
\multicolumn{15}{l}{\textit{OmniSVG-8B, image-to-SVG}} \\
PLD & 1.60$\times$ & 1.87 & 1.83$\times$ & 2.17 & 1.34$\times$ & 1.55 & 1.76$\times$ & 2.07 & 1.73$\times$ & 2.05 & 2.09$\times$ & 2.53 & 1.72$\times$ & 2.04 \\
REST & 1.52$\times$ & 1.80 & 1.59$\times$ & 1.88 & 1.55$\times$ & 1.85 & 1.57$\times$ & 1.87 & 1.59$\times$ & 1.90 & 1.71$\times$ & 2.05 & 1.59$\times$ & 1.89 \\
EAGLE-3 & 2.95$\times$ & 9.02 & 3.18$\times$ & 9.77 & 2.85$\times$ & 8.85 & 3.13$\times$ & 9.70 & 3.11$\times$ & 9.62 & 3.21$\times$ & 9.90 & 3.07$\times$ & 9.48 \\
ViSpec & 1.86$\times$ & 3.23 & 1.98$\times$ & 3.46 & 1.78$\times$ & 3.06 & 1.94$\times$ & 3.37 & 2.01$\times$ & 3.49 & 2.07$\times$ & 3.58 & 1.94$\times$ & 3.36 \\
DFlash & 4.15$\times$ & 7.08 & 4.54$\times$ & 7.83 & 3.91$\times$ & 6.63 & 4.44$\times$ & 7.60 & 4.16$\times$ & 7.55 & 4.57$\times$ & 8.45 & 4.30$\times$ & 7.52 \\
DARTree & 4.48$\times$ & \underline{9.77} & 4.71$\times$ & 10.37 & 4.15$\times$ & 9.08 & 4.65$\times$ & 10.19 & 4.58$\times$ & 10.21 & 4.75$\times$ & 10.65 & 4.55$\times$ & 10.04 \\
DDTree & \underline{4.99}$\times$ & 9.70 & \underline{5.30}$\times$ & \underline{10.38} & \underline{4.89}$\times$ & \underline{9.57} & \underline{5.30}$\times$ & \underline{10.38} & \underline{4.96}$\times$ & \underline{10.28} & \underline{5.29}$\times$ & \underline{11.10} & \underline{5.12}$\times$ & \underline{10.24} \\
\textbf{Ours} & \textbf{5.92}$\times$ & \textbf{12.40} & \textbf{6.38}$\times$ & \textbf{13.69} & \textbf{5.42}$\times$ & \textbf{11.32} & \textbf{6.29}$\times$ & \textbf{13.35} & \textbf{5.99}$\times$ & \textbf{12.70} & \textbf{6.72}$\times$ & \textbf{14.81} & \textbf{6.12}$\times$ & \textbf{13.05} \\
\midrule
\multicolumn{15}{l}{\textit{InternSVG-8B, image-to-SVG}} \\
PLD & 1.65$\times$ & 1.96 & 1.86$\times$ & 2.23 & 1.54$\times$ & 1.81 & 1.86$\times$ & 2.24 & 1.66$\times$ & 1.94 & 1.60$\times$ & 1.93 & 1.70$\times$ & 2.02 \\
REST & 1.50$\times$ & 1.89 & 1.47$\times$ & 1.86 & 1.56$\times$ & 1.95 & 1.49$\times$ & 1.92 & 1.49$\times$ & 1.85 & 1.44$\times$ & 1.90 & 1.49$\times$ & 1.90 \\
EAGLE-3 & 2.56$\times$ & 7.85 & 2.50$\times$ & 7.56 & 2.50$\times$ & 7.50 & 2.61$\times$ & 8.05 & 2.39$\times$ & 6.95 & 2.39$\times$ & 7.21 & 2.49$\times$ & 7.52 \\
ViSpec & 2.06$\times$ & 6.09 & 1.96$\times$ & 5.87 & 2.15$\times$ & 6.34 & 2.12$\times$ & 6.42 & 1.95$\times$ & 5.83 & 1.63$\times$ & 5.17 & 1.98$\times$ & 5.95 \\
DFlash & 3.58$\times$ & 5.81 & 3.49$\times$ & 5.64 & 3.60$\times$ & 5.84 & 3.65$\times$ & 6.00 & 3.32$\times$ & 5.28 & 2.71$\times$ & 4.36 & 3.39$\times$ & 5.49 \\
DARTree & 4.25$\times$ & \underline{9.03} & 4.20$\times$ & \underline{8.85} & 4.32$\times$ & \underline{9.06} & 4.30$\times$ & \underline{9.31} & 3.96$\times$ & \underline{8.17} & 3.52$\times$ & \underline{7.49} & 4.09$\times$ & \underline{8.65} \\
DDTree & \underline{4.77}$\times$ & 8.57 & \underline{4.55}$\times$ & 8.13 & \underline{4.94}$\times$ & 8.91 & \underline{4.90}$\times$ & 9.05 & \underline{4.33}$\times$ & 7.57 & \underline{4.00}$\times$ & 7.24 & \underline{4.58}$\times$ & 8.24 \\
\textbf{Ours} & \textbf{5.27}$\times$ & \textbf{9.68} & \textbf{5.25}$\times$ & \textbf{9.56} & \textbf{5.71}$\times$ & \textbf{10.47} & \textbf{5.84}$\times$ & \textbf{11.09} & \textbf{4.81}$\times$ & \textbf{8.44} & \textbf{4.25}$\times$ & \textbf{7.69} & \textbf{5.19}$\times$ & \textbf{9.49} \\
\bottomrule
\end{tabular*}
\end{table*}

\begin{table*}[t]
\caption{\textbf{Tokens per round on loop-free content.} $\tau$ over the rounds that start before the output enters a loop, on the outputs of Figure~\ref{fig:main}. Loops hold 13\% to 75\% of the generated tokens, depending on model and benchmark. \textbf{Best} in bold, \underline{second}{} underlined, within each model.}
\label{app:t1-healthy}
\centering
\footnotesize
\setlength{\tabcolsep}{2.2pt}
\begin{tabular*}{\textwidth}{@{\extracolsep{\fill}}l*{14}{c}@{}}
\toprule
 & \multicolumn{7}{c}{Greedy} & \multicolumn{7}{c}{Official sampling} \\
\cmidrule(lr){2-8}\cmidrule(lr){9-15}
 & \multicolumn{2}{c}{MMSVG} & \multicolumn{2}{c}{MMSVG-2M$^\dagger$} & \multicolumn{2}{c}{SArena} & & \multicolumn{2}{c}{MMSVG} & \multicolumn{2}{c}{MMSVG-2M$^\dagger$} & \multicolumn{2}{c}{SArena} & \\
\cmidrule(lr){2-3}\cmidrule(lr){4-5}\cmidrule(lr){6-7}\cmidrule(lr){9-10}\cmidrule(lr){11-12}\cmidrule(lr){13-14}
Method & Icon & Illus. & Icon & Illus. & Icon & Illus. & Mean & Icon & Illus. & Icon & Illus. & Icon & Illus. & Mean \\
\midrule
\multicolumn{15}{l}{\textit{OmniSVG-8B, text-to-SVG}} \\
PLD & 1.54 & 1.70 & 1.31 & 1.59 & 1.45 & 1.60 & 1.53 & 1.58 & 1.81 & 1.33 & 1.62 & 1.41 & 1.69 & 1.57 \\
REST & 2.20 & 2.16 & 2.07 & 2.20 & 2.16 & 2.18 & 2.16 & 1.87 & 2.00 & 2.03 & 1.96 & 2.00 & 2.00 & 1.97 \\
SpS & 2.70 & 2.65 & 2.71 & 2.70 & 2.71 & 2.70 & 2.69 & 2.64 & 2.63 & 2.66 & 2.70 & 2.65 & 2.63 & 2.65 \\
EAGLE-3 & 9.06 & 9.79 & 8.52 & 9.39 & 8.72 & 9.67 & 9.19 & 8.74 & 9.23 & 8.49 & 9.02 & 8.47 & 9.21 & 8.86 \\
DFlash & 6.96 & 7.49 & 6.80 & 7.54 & 6.82 & 7.28 & 7.15 & 6.91 & 7.02 & 6.36 & 6.86 & 6.41 & 6.76 & 6.72 \\
DARTree & \underline{9.86} & \underline{10.86} & \underline{8.91} & \underline{10.70} & \underline{9.00} & \underline{10.63} & \underline{9.99} & \underline{9.60} & \underline{10.30} & \underline{8.68} & \underline{9.91} & \underline{8.68} & \underline{10.15} & \underline{9.55} \\
DDTree & 9.32 & 10.35 & 8.83 & 10.07 & 8.90 & 10.24 & 9.62 & 8.91 & 9.65 & 8.53 & 9.30 & 8.59 & 9.47 & 9.07 \\
\textbf{Ours} & \textbf{11.76} & \textbf{12.72} & \textbf{10.32} & \textbf{12.27} & \textbf{10.65} & \textbf{12.72} & \textbf{11.74} & \textbf{11.17} & \textbf{11.76} & \textbf{10.00} & \textbf{11.41} & \textbf{9.91} & \textbf{11.32} & \textbf{10.93} \\
\midrule
\multicolumn{15}{l}{\textit{OmniSVG-4B, text-to-SVG}} \\
PLD & 1.32 & 1.56 & 1.32 & 1.52 & 1.35 & 1.49 & 1.43 & 1.35 & 1.40 & 1.34 & 1.40 & 1.32 & 1.52 & 1.39 \\
REST & 2.08 & 2.25 & 2.25 & 2.18 & 2.08 & 2.02 & 2.14 & 1.89 & 1.94 & 2.06 & 1.98 & 1.99 & 2.03 & 1.98 \\
EAGLE-3 & 6.80 & 8.37 & 7.63 & 8.13 & 7.25 & 7.96 & 7.69 & 6.66 & 7.31 & 7.37 & 7.27 & 6.82 & 7.71 & 7.19 \\
DFlash & 5.15 & 5.90 & 5.53 & 6.38 & 5.25 & 5.86 & 5.68 & 4.71 & 5.15 & 5.29 & 5.02 & 4.82 & 5.23 & 5.04 \\
DARTree & \underline{8.27} & \underline{9.57} & \underline{8.06} & \underline{9.18} & \underline{7.61} & \underline{9.17} & \underline{8.64} & \underline{7.47} & \underline{7.98} & \underline{7.93} & \underline{7.86} & 6.97 & \underline{8.12} & \underline{7.72} \\
DDTree & 7.84 & 9.03 & 7.80 & 8.71 & 7.30 & 8.79 & 8.25 & 7.42 & 7.81 & 7.90 & 7.71 & \underline{7.20} & 8.05 & 7.68 \\
\textbf{Ours} & \textbf{8.44} & \textbf{9.92} & \textbf{8.38} & \textbf{9.96} & \textbf{8.14} & \textbf{10.04} & \textbf{9.15} & \textbf{7.94} & \textbf{8.77} & \textbf{8.24} & \textbf{8.39} & \textbf{7.46} & \textbf{8.82} & \textbf{8.27} \\
\midrule
\multicolumn{15}{l}{\textit{InternSVG-8B, text-to-SVG}} \\
PLD & 2.39 & 2.81 & 1.90 & 2.61 & 2.03 & 2.43 & 2.36 & 2.28 & 2.54 & 1.79 & 2.37 & 1.98 & 1.97 & 2.15 \\
REST & 2.74 & 3.46 & 2.88 & 2.84 & 2.72 & 2.65 & 2.88 & 2.16 & 2.26 & 2.53 & 2.34 & 2.24 & 2.18 & 2.29 \\
EAGLE-3 & 10.97 & 11.35 & 9.97 & 10.40 & 8.89 & 9.51 & 10.18 & 8.27 & 8.29 & 8.25 & 9.00 & 7.66 & 7.66 & 8.19 \\
DFlash & 9.76 & 10.67 & 8.40 & 9.25 & 7.57 & 7.16 & 8.80 & 5.76 & 6.57 & 6.69 & 7.01 & 6.08 & 4.77 & 6.15 \\
DARTree & \underline{13.75} & \underline{15.00} & \underline{12.09} & \underline{13.09} & \underline{11.30} & \underline{11.29} & \underline{12.75} & \underline{10.18} & \underline{10.28} & \underline{10.25} & \underline{10.98} & \underline{9.07} & \underline{8.10} & \underline{9.81} \\
DDTree & 12.43 & 13.50 & 11.23 & 12.03 & 10.15 & 10.27 & 11.60 & 9.20 & 10.21 & 9.52 & 9.91 & 8.85 & 7.87 & 9.26 \\
\textbf{Ours} & \textbf{16.33} & \textbf{19.29} & \textbf{13.52} & \textbf{16.41} & \textbf{12.08} & \textbf{12.14} & \textbf{14.96} & \textbf{11.19} & \textbf{10.99} & \textbf{10.55} & \textbf{11.92} & \textbf{10.11} & \textbf{8.49} & \textbf{10.54} \\
\midrule
\multicolumn{15}{l}{\textit{OmniSVG-8B, image-to-SVG}} \\
PLD & 1.34 & 1.53 & 1.32 & 1.45 & 1.51 & 1.60 & 1.46 & 1.40 & 1.60 & 1.30 & 1.50 & 1.49 & 1.60 & 1.48 \\
REST & 1.52 & 1.57 & 1.72 & 1.53 & 1.68 & 1.64 & 1.61 & 1.54 & 1.61 & 1.72 & 1.57 & 1.68 & 1.62 & 1.62 \\
EAGLE-3 & 7.97 & 8.31 & 8.19 & 8.26 & 8.40 & 8.53 & 8.28 & 8.15 & 8.38 & 8.31 & 8.44 & 8.55 & 8.28 & 8.35 \\
ViSpec & 2.83 & 2.99 & 2.83 & 2.91 & 2.95 & 3.01 & 2.92 & 2.88 & 3.03 & 2.82 & 2.93 & 3.05 & 2.97 & 2.95 \\
DFlash & 5.80 & 6.26 & 6.00 & 6.02 & 6.02 & 6.14 & 6.04 & 5.84 & 6.18 & 5.92 & 6.12 & 6.03 & 6.05 & 6.02 \\
DARTree & 8.15 & 8.64 & 8.14 & 8.20 & 8.27 & 8.43 & 8.31 & 8.48 & 8.68 & 8.15 & 8.44 & 8.61 & 8.17 & 8.42 \\
DDTree & \underline{8.52} & \underline{9.20} & \underline{8.57} & \underline{8.88} & \underline{8.76} & \underline{9.05} & \underline{8.83} & \underline{8.51} & \underline{8.82} & \underline{8.67} & \underline{8.87} & \underline{8.72} & \underline{8.75} & \underline{8.72} \\
\textbf{Ours} & \textbf{9.54} & \textbf{10.52} & \textbf{10.32} & \textbf{9.84} & \textbf{9.69} & \textbf{10.25} & \textbf{10.03} & \textbf{10.05} & \textbf{10.51} & \textbf{9.94} & \textbf{10.43} & \textbf{10.13} & \textbf{9.96} & \textbf{10.17} \\
\midrule
\multicolumn{15}{l}{\textit{InternSVG-8B, image-to-SVG}} \\
PLD & 1.72 & 2.04 & 1.62 & 2.08 & 1.69 & 1.90 & 1.84 & 1.79 & 1.89 & 1.56 & 1.96 & 1.70 & 1.73 & 1.77 \\
REST & 2.07 & 2.15 & 2.26 & 2.22 & 2.07 & 2.09 & 2.14 & 1.87 & 1.87 & 2.02 & 1.94 & 1.90 & 1.92 & 1.92 \\
EAGLE-3 & 8.32 & 8.94 & 7.69 & 8.57 & 7.77 & 7.82 & 8.19 & 7.17 & 7.14 & 6.89 & 7.42 & 6.90 & 6.80 & 7.05 \\
ViSpec & 6.35 & 6.91 & 6.13 & 6.79 & 5.97 & 5.26 & 6.23 & 5.60 & 5.46 & 5.72 & 5.76 & 5.41 & 4.85 & 5.47 \\
DFlash & 6.35 & 6.49 & 5.99 & 6.69 & 5.78 & 4.92 & 6.04 & 5.23 & 4.98 & 5.26 & 5.35 & 5.07 & 4.00 & 4.98 \\
DARTree & \underline{9.62} & \underline{10.54} & \underline{9.49} & \underline{10.72} & \underline{9.30} & \underline{8.51} & \underline{9.70} & \underline{8.32} & \underline{8.16} & \underline{8.31} & \underline{8.34} & \underline{7.72} & \underline{7.01} & \underline{7.98} \\
DDTree & 9.02 & 9.42 & 8.93 & 9.62 & 8.69 & 7.91 & 8.93 & 8.05 & 7.51 & 7.98 & 8.08 & 7.44 & 6.77 & 7.64 \\
\textbf{Ours} & \textbf{10.67} & \textbf{11.14} & \textbf{9.82} & \textbf{11.45} & \textbf{9.62} & \textbf{8.56} & \textbf{10.21} & \textbf{8.70} & \textbf{8.68} & \textbf{8.73} & \textbf{9.31} & \textbf{8.22} & \textbf{7.02} & \textbf{8.44} \\
\bottomrule
\end{tabular*}
\end{table*}

\begin{table*}[t]
\caption{\textbf{The hardest and the easiest prompts.} Our outputs of Figure~\ref{fig:main} under greedy decoding, ranked by their own speedup over the original model and split into the slowest 10\%, the middle 80\% and the fastest 10\%. The slowest tenth is still 2.6$\times$ to 2.8$\times$ faster.}
\label{app:hard-rows}
\centering
\footnotesize
\setlength{\tabcolsep}{3.0pt}
\begin{tabular*}{\textwidth}{@{\extracolsep{\fill}}l*{7}{c}@{}}
\toprule
Rows & $n$ & speedup & $\tau$ & tokens & rounds filling the window (\%) & ended normally (\%) & original model hit the cap (\%) \\
\midrule
\multicolumn{8}{l}{\textit{OmniSVG-8B, text-to-SVG}} \\
slowest 10\% & 90 & 2.61$\times$ & 8.25 & 599 & 8 & 76 & 1 \\
middle 80\% & 720 & 8.72$\times$ & 16.79 & 762 & 25 & 63 & 37 \\
fastest 10\% & 90 & 16.72$\times$ & 23.40 & 888 & 56 & 58 & 63 \\
\midrule
\multicolumn{8}{l}{\textit{OmniSVG-8B, image-to-SVG}} \\
slowest 10\% & 90 & 2.84$\times$ & 6.85 & 533 & 3 & 84 & 1 \\
middle 80\% & 720 & 7.10$\times$ & 13.10 & 773 & 13 & 69 & 33 \\
fastest 10\% & 90 & 13.97$\times$ & 22.57 & 1273 & 52 & 26 & 91 \\
\bottomrule
\end{tabular*}
\end{table*}

\subsection{Speed and Quality on the Full SArena Test Sets}
\label{app:quality-full}
Tables~\ref{app:quality-t2s} and~\ref{app:quality-i2s} give every official metric of SArena for the rows of Table~\ref{tab:quality}, and a third group in each draws eight samples per prompt.
Its rows keep the sample whose rendering scores highest, except the authors' pipeline of OmniSVG, which keeps the first valid one.
The original model appears in this group with and without loop control.
Rendered selection improves every metric over one sample of the original model, including FID and LPIPS, which it does not use.
Our best of eight takes less time on average than one sample of the original model as officially decoded, and less than half the time of the original model's own best of eight.
In a batch, every sample verifies a chain, so the chain head of DFlash draws its eight about as fast.

\begin{table*}[t]
\caption{\textbf{Text-to-SVG on the full SArena test sets, every official metric.} The lossless and loop-control groups are those of Table~\ref{tab:quality}, and a third group draws eight samples per prompt and keeps the best by its rendering (Appendix~\ref{app:quality}). That group also lists the authors' pipeline of OmniSVG, which samples eight and keeps the first valid one, and the original model without loop control, decoded as officially released. Seconds per prompt on one H200, with \textbf{fastest} in bold and \underline{second}{} underlined within a group.}
\label{app:quality-t2s}
\centering
\small
\setlength{\tabcolsep}{2.0pt}
\begin{tabular*}{\textwidth}{@{\extracolsep{\fill}}ll*{10}{c}@{}}
\toprule
 &  & \multicolumn{5}{c}{SArena-Icon (text-to-SVG, 6,013)} & \multicolumn{5}{c}{SArena-Illus. (text-to-SVG, 2,001)} \\
\cmidrule(lr){3-7}\cmidrule(lr){8-12}
 & Method & FID$\downarrow$ & FID-C$\downarrow$ & CLIP-T2I$\uparrow$ & CLIP-I2I$\uparrow$ & s$\downarrow$ & FID$\downarrow$ & FID-C$\downarrow$ & CLIP-T2I$\uparrow$ & CLIP-I2I$\uparrow$ & s$\downarrow$ \\
\midrule
\multicolumn{12}{l}{\textbf{OmniSVG-8B}, its released sampling settings} \\
\multirow{6}{1.05cm}{\centering\itshape\shortstack{same\\outputs}} & AR & 10.81 & 2.91 & 22.76 & 78.73 & 5.72 & 38.93 & 26.33 & 17.40 & 65.19 & 15.92 \\
 & EAGLE-3 & 10.93 & 2.97 & 22.79 & 78.76 & 1.91 & 39.09 & 26.58 & 17.31 & 65.23 & 4.76 \\
 & DFlash & 10.76 & 2.93 & 22.80 & 78.77 & 1.29 & 39.30 & 26.64 & 17.39 & 65.17 & 3.08 \\
 & DDTree & 10.95 & 2.99 & 22.79 & 78.67 & \underline{1.14} & 39.19 & 26.81 & 17.33 & 65.27 & \underline{2.79} \\
 & DARTree & 10.89 & 2.99 & 22.79 & 78.70 & 1.27 & 39.16 & 26.86 & 17.36 & 65.19 & 3.06 \\
 & \textbf{Ours} & 10.86 & 2.95 & 22.77 & 78.71 & \textbf{1.01} & 39.24 & 26.81 & 17.32 & 65.04 & \textbf{2.28} \\
\cmidrule(lr){1-12}
\multirow{6}{1.05cm}{\centering\itshape\shortstack{loop\\control}} & AR & 10.89 & 2.94 & 22.76 & 78.73 & 4.46 & 39.07 & 27.04 & 17.37 & 65.08 & 8.46 \\
 & EAGLE-3 & 10.99 & 3.01 & 22.81 & 78.83 & 1.54 & 39.22 & 27.39 & 17.33 & 65.03 & 2.81 \\
 & DFlash & 10.87 & 2.98 & 22.80 & 78.80 & 1.08 & 39.34 & 27.32 & 17.38 & 65.20 & 2.05 \\
 & DDTree & 10.99 & 3.02 & 22.79 & 78.67 & \underline{0.94} & 39.27 & 27.37 & 17.33 & 65.23 & \underline{1.76} \\
 & DARTree & 11.02 & 3.02 & 22.79 & 78.70 & 1.08 & 38.99 & 27.35 & 17.35 & 65.09 & 1.96 \\
 & \textbf{Ours} & 10.94 & 3.00 & 22.76 & 78.71 & \textbf{0.80} & 39.18 & 27.24 & 17.33 & 65.06 & \textbf{1.56} \\
\cmidrule(lr){1-12}
\multirow{5}{1.05cm}{\centering\itshape\shortstack{best\\of 8}} & Authors' pipeline & 9.96 & 2.55 & 22.83 & 78.86 & 12.81 & 31.11 & 19.08 & 17.45 & 65.65 & 24.95 \\
 & AR, no loop stop & 9.73 & 2.48 & 23.98 & 80.27 & 12.81 & 28.49 & 15.92 & 18.66 & 67.09 & 24.95 \\
 & AR & 9.85 & 2.51 & 23.96 & 80.23 & 10.40 & 28.93 & 16.23 & 18.58 & 66.74 & 21.13 \\
 & DFlash & 9.82 & 2.50 & 24.01 & 80.15 & \textbf{2.74} & 29.13 & 16.32 & 18.70 & 66.98 & \textbf{5.32} \\
 & \textbf{Ours} & 9.85 & 2.48 & 23.96 & 80.25 & \underline{2.76} & 29.02 & 16.45 & 18.75 & 67.12 & \underline{5.37} \\
\midrule
\multicolumn{12}{l}{\textbf{InternSVG-8B}, greedy; best of 8 at $T{=}0.6$, top-$p$ 0.9} \\
\multirow{6}{1.05cm}{\centering\itshape\shortstack{same\\outputs}} & AR & 11.83 & 3.17 & 23.71 & 80.53 & 11.21 & 52.91 & 29.65 & 17.95 & 66.83 & 43.58 \\
 & EAGLE-3 & 11.75 & 3.16 & 23.76 & 80.52 & 3.28 & 52.84 & 29.35 & 17.92 & 66.99 & 12.62 \\
 & DFlash & 11.88 & 3.23 & 23.74 & 80.53 & 1.81 & 54.32 & 30.36 & 17.90 & 66.80 & 7.31 \\
 & DDTree & 11.94 & 3.24 & 23.71 & 80.50 & \underline{1.60} & 53.83 & 30.60 & 17.89 & 66.66 & \underline{6.18} \\
 & DARTree & 11.89 & 3.26 & 23.74 & 80.56 & 1.77 & 54.62 & 30.77 & 17.88 & 66.74 & 6.81 \\
 & \textbf{Ours} & 11.84 & 3.18 & 23.73 & 80.58 & \textbf{1.14} & 54.37 & 30.56 & 17.88 & 66.80 & \textbf{4.70} \\
\cmidrule(lr){1-12}
\multirow{6}{1.05cm}{\centering\itshape\shortstack{loop\\control}} & AR & 12.22 & 3.32 & 23.69 & 80.45 & 7.91 & 53.42 & 29.87 & 17.96 & 66.79 & 29.88 \\
 & EAGLE-3 & 12.09 & 3.24 & 23.72 & 80.43 & 2.48 & 54.18 & 29.99 & 17.95 & 66.85 & 9.33 \\
 & DFlash & 12.07 & 3.29 & 23.74 & 80.47 & 1.47 & 54.93 & 30.58 & 17.86 & 66.71 & 5.81 \\
 & DDTree & 12.38 & 3.39 & 23.69 & 80.42 & \underline{1.23} & 54.15 & 30.88 & 17.87 & 66.59 & \underline{4.67} \\
 & DARTree & 12.22 & 3.39 & 23.74 & 80.51 & 1.38 & 54.87 & 30.82 & 17.88 & 66.71 & 5.14 \\
 & \textbf{Ours} & 12.10 & 3.25 & 23.72 & 80.52 & \textbf{1.05} & 54.93 & 30.96 & 17.87 & 66.75 & \textbf{3.94} \\
\cmidrule(lr){1-12}
\multirow{4}{1.05cm}{\centering\itshape\shortstack{best\\of 8}} & AR, no loop stop & 8.75 & 2.12 & 25.21 & 82.59 & 26.28 & 23.98 & 6.68 & 21.51 & 74.04 & 86.43 \\
 & AR & 8.75 & 2.14 & 25.19 & 82.55 & 21.51 & 24.15 & 6.85 & 21.47 & 73.87 & 85.97 \\
 & DFlash & 8.80 & 2.14 & 25.22 & 82.50 & \textbf{5.22} & 23.90 & 6.62 & 21.48 & 74.03 & \textbf{29.53} \\
 & \textbf{Ours} & 8.80 & 2.13 & 25.22 & 82.53 & \underline{5.25} & 24.25 & 6.81 & 21.44 & 74.05 & \underline{31.63} \\
\bottomrule
\end{tabular*}
\end{table*}

\begin{table*}[t]
\caption{\textbf{Image-to-SVG on the full SArena test sets, every official metric.} Layout as in Table~\ref{app:quality-t2s}, with candidates scored by DINO similarity to the input image.}
\label{app:quality-i2s}
\centering
\small
\setlength{\tabcolsep}{2.0pt}
\begin{tabular*}{\textwidth}{@{\extracolsep{\fill}}ll*{10}{c}@{}}
\toprule
 &  & \multicolumn{5}{c}{SArena-Icon (image-to-SVG, 6,013)} & \multicolumn{5}{c}{SArena-Illus. (image-to-SVG, 2,001)} \\
\cmidrule(lr){3-7}\cmidrule(lr){8-12}
 & Method & DINO$\uparrow$ & LPIPS$\downarrow$ & SSIM$\uparrow$ & PSNR$\uparrow$ & s$\downarrow$ & DINO$\uparrow$ & LPIPS$\downarrow$ & SSIM$\uparrow$ & PSNR$\uparrow$ & s$\downarrow$ \\
\midrule
\multicolumn{12}{l}{\textbf{OmniSVG-8B}, its released sampling settings} \\
\multirow{6}{1.05cm}{\centering\itshape\shortstack{same\\outputs}} & AR & 0.903 & 0.178 & 0.779 & 14.52 & 8.85 & 0.803 & 0.272 & 0.715 & 11.58 & 19.82 \\
 & EAGLE-3 & 0.902 & 0.178 & 0.778 & 14.49 & 2.96 & 0.805 & 0.270 & 0.717 & 11.62 & 6.14 \\
 & DFlash & 0.901 & 0.179 & 0.779 & 14.48 & 2.11 & 0.803 & 0.270 & 0.720 & 11.74 & 4.24 \\
 & DDTree & 0.902 & 0.178 & 0.779 & 14.49 & \underline{1.76} & 0.804 & 0.271 & 0.717 & 11.60 & \underline{3.68} \\
 & DARTree & 0.902 & 0.177 & 0.780 & 14.54 & 1.96 & 0.803 & 0.272 & 0.716 & 11.60 & 4.08 \\
 & \textbf{Ours} & 0.902 & 0.178 & 0.779 & 14.54 & \textbf{1.48} & 0.805 & 0.272 & 0.715 & 11.56 & \textbf{2.94} \\
\cmidrule(lr){1-12}
\multirow{6}{1.05cm}{\centering\itshape\shortstack{loop\\control}} & AR & 0.903 & 0.177 & 0.780 & 14.55 & 6.05 & 0.805 & 0.272 & 0.713 & 11.63 & 11.43 \\
 & EAGLE-3 & 0.903 & 0.177 & 0.779 & 14.52 & 2.33 & 0.807 & 0.270 & 0.714 & 11.64 & 4.26 \\
 & DFlash & 0.902 & 0.178 & 0.780 & 14.51 & 1.77 & 0.805 & 0.270 & 0.716 & 11.73 & 3.25 \\
 & DDTree & 0.903 & 0.177 & 0.780 & 14.52 & \underline{1.42} & 0.806 & 0.273 & 0.709 & 11.52 & \underline{2.64} \\
 & DARTree & 0.903 & 0.177 & 0.781 & 14.57 & 1.57 & 0.805 & 0.272 & 0.712 & 11.60 & 2.91 \\
 & \textbf{Ours} & 0.903 & 0.177 & 0.781 & 14.58 & \textbf{1.30} & 0.806 & 0.272 & 0.712 & 11.57 & \textbf{2.37} \\
\cmidrule(lr){1-12}
\multirow{5}{1.05cm}{\centering\itshape\shortstack{best\\of 8}} & Authors' pipeline & 0.908 & 0.178 & 0.774 & 14.42 & 17.80 & 0.818 & 0.272 & 0.700 & 11.25 & 30.15 \\
 & AR, no loop stop & 0.952 & 0.125 & 0.839 & 17.14 & 17.80 & 0.875 & 0.226 & 0.753 & 13.02 & 30.15 \\
 & AR & 0.950 & 0.126 & 0.838 & 17.08 & 14.93 & 0.870 & 0.226 & 0.754 & 13.06 & 25.61 \\
 & DFlash & 0.950 & 0.126 & 0.838 & 17.10 & \textbf{3.87} & 0.869 & 0.229 & 0.752 & 12.96 & \textbf{6.69} \\
 & \textbf{Ours} & 0.950 & 0.126 & 0.838 & 17.07 & \underline{4.05} & 0.870 & 0.229 & 0.752 & 12.93 & \underline{7.11} \\
\midrule
\multicolumn{12}{l}{\textbf{InternSVG-8B}, greedy; best of 8 at $T{=}0.6$, top-$p$ 0.9} \\
\multirow{6}{1.05cm}{\centering\itshape\shortstack{same\\outputs}} & AR & 0.965 & 0.077 & 0.898 & 23.80 & 10.98 & 0.864 & 0.204 & 0.786 & 14.89 & 43.93 \\
 & EAGLE-3 & 0.963 & 0.078 & 0.898 & 23.82 & 3.66 & 0.869 & 0.202 & 0.787 & 14.95 & 14.65 \\
 & DFlash & 0.965 & 0.077 & 0.898 & 23.76 & 2.37 & 0.870 & 0.201 & 0.788 & 14.99 & 10.95 \\
 & DDTree & 0.963 & 0.078 & 0.898 & 23.80 & \underline{1.88} & 0.864 & 0.204 & 0.786 & 14.89 & \underline{7.86} \\
 & DARTree & 0.965 & 0.077 & 0.898 & 23.81 & 2.03 & 0.868 & 0.202 & 0.787 & 15.01 & 8.81 \\
 & \textbf{Ours} & 0.963 & 0.078 & 0.897 & 23.76 & \textbf{1.54} & 0.866 & 0.203 & 0.786 & 14.93 & \textbf{6.96} \\
\cmidrule(lr){1-12}
\multirow{6}{1.05cm}{\centering\itshape\shortstack{loop\\control}} & AR & 0.965 & 0.077 & 0.897 & 23.75 & 8.49 & 0.866 & 0.203 & 0.785 & 14.89 & 34.55 \\
 & EAGLE-3 & 0.964 & 0.077 & 0.897 & 23.78 & 3.01 & 0.868 & 0.204 & 0.783 & 14.86 & 12.54 \\
 & DFlash & 0.965 & 0.077 & 0.897 & 23.73 & 2.13 & 0.871 & 0.202 & 0.786 & 14.91 & 10.12 \\
 & DDTree & 0.964 & 0.078 & 0.896 & 23.76 & \underline{1.56} & 0.866 & 0.204 & 0.784 & 14.86 & \underline{6.86} \\
 & DARTree & 0.965 & 0.077 & 0.898 & 23.78 & 1.78 & 0.870 & 0.202 & 0.785 & 14.94 & 7.98 \\
 & \textbf{Ours} & 0.963 & 0.079 & 0.896 & 23.72 & \textbf{1.46} & 0.866 & 0.204 & 0.784 & 14.86 & \textbf{6.54} \\
\cmidrule(lr){1-12}
\multirow{4}{1.05cm}{\centering\itshape\shortstack{best\\of 8}} & AR, no loop stop & 0.992 & 0.041 & 0.934 & 28.67 & 24.44 & 0.966 & 0.132 & 0.821 & 17.47 & 89.87 \\
 & AR & 0.991 & 0.042 & 0.932 & 28.54 & 21.18 & 0.964 & 0.134 & 0.819 & 17.40 & 92.40 \\
 & DFlash & 0.991 & 0.042 & 0.932 & 28.51 & \textbf{6.32} & 0.964 & 0.135 & 0.819 & 17.38 & \textbf{37.75} \\
 & \textbf{Ours} & 0.991 & 0.043 & 0.931 & 28.42 & \underline{6.64} & 0.965 & 0.131 & 0.822 & 17.52 & \underline{41.20} \\
\bottomrule
\end{tabular*}
\end{table*}

\subsection{Comparison with SVG Generators}
\label{app:generators}
Table~\ref{app:generators-tab} lists every generator of Figure~\ref{fig:ex1}a.
InternSVG reports five of them, the quality in its Tables 2 and 3 and the seconds per SVG in its Table 19, which covers icons only.
We measure SVGen-7B, SVGen-3B and HiVG-3B with their released sampling settings and length limits.
Their quality comes from the full test set with the official metrics, and their seconds from 300 prompts on one H200.
The rows of our two targets are those of Table~\ref{app:quality-t2s}.
InternSVG-8B thus appears twice, and the two measurements differ in hardware and in how outputs are extracted (Appendix~\ref{app:metrics}).
InternSVG reports 6.9\,s and an FID of 8.71 for it, and we measure 11.2\,s and 11.83 for its original decoding.

\begin{table}[htbp]\centering
\begingroup
\renewenvironment{table}[1][]{\begin{minipage}[t]{0.74\textwidth}\centering}{\end{minipage}}%
\begin{table}[t]
\caption{\textbf{SVG generators on SArena-Icon text-to-SVG, Figure~\ref{fig:ex1}a in full.} Reported rows are quoted from InternSVG, and measured rows come from our runs. Seconds are per SVG, and the rows of our two targets are those of Table~\ref{app:quality-t2s}.}
\label{app:generators-tab}
\centering
\footnotesize
\setlength{\tabcolsep}{3pt}
\begin{tabular*}{\columnwidth}{@{\extracolsep{\fill}}llccccc@{}}
\toprule
Generator & Source & FID$\downarrow$ & FID-C$\downarrow$ & CLIP-T2I$\uparrow$ & CLIP-I2I$\uparrow$ & s$\downarrow$ \\
\midrule
\multicolumn{7}{@{}l}{\textit{Optimization-based}} \\
VectorFusion~\cite{10204913} & reported & 16.59 & 17.39 & 22.99 & 70.31 & 71.0 \\
SVGDreamer~\cite{10658565} & reported & 26.61 & 33.31 & 20.33 & 69.97 & 289.0 \\
\addlinespace[2pt]
\multicolumn{7}{@{}l}{\textit{Other SVG generation models}} \\
LLM4SVG-7B~\cite{Xing_2025_CVPR} & reported & 21.94 & 8.61 & 19.46 & 70.73 & 9.7 \\
OmniSVG-3B & reported & 28.29 & 11.32 & 21.68 & 74.83 & 12.3 \\
InternSVG-8B & reported & 8.71 & 1.88 & 23.92 & 80.91 & 6.9 \\
SVGen-7B~\cite{10.1145/3746027.3755011} & measured & 24.68 & 7.52 & 22.05 & 76.05 & 21.07 \\
SVGen-3B & measured & 28.16 & 9.89 & 21.66 & 73.81 & 17.29 \\
HiVG-3B~\cite{xing2026hierarchical} & measured & 33.97 & 11.61 & 23.31 & 76.03 & 26.29 \\
\addlinespace[2pt]
\multicolumn{7}{@{}l}{\textit{Our targets}} \\
OmniSVG-8B & measured & 10.81 & 2.91 & 22.76 & 78.73 & 5.72 \\
\quad + Sightline & measured & 10.86 & 2.95 & 22.77 & 78.71 & 1.01 \\
\quad + Sightline, best of 8 & measured & 9.85 & 2.48 & 23.96 & 80.25 & 2.76 \\
InternSVG-8B & measured & 11.83 & 3.17 & 23.71 & 80.53 & 11.21 \\
\quad + Sightline & measured & 11.84 & 3.18 & 23.73 & 80.58 & 1.14 \\
\quad + Sightline, best of 8 & measured & 8.80 & 2.13 & 25.22 & 82.53 & 5.25 \\
\bottomrule
\end{tabular*}
\end{table}
\endgroup
\end{table}

\subsection{Other Model Families}
\label{app:families}
Tables~\ref{app:t3-speed} to~\ref{app:t3-settings} give the six models of Figure~\ref{fig:ex1}b setting by setting, with the sampling setting of each.

\begin{table*}[t]
\caption{\textbf{Figure~\ref{fig:ex1}b in full, speedup.} Wall-clock speedup over the original model timed on the same node. StarVector's last row samples at $T$ 1.0 and top-$p$ 0.9, the defaults of its released generation code, and its official row uses its released evaluation configuration. \textbf{Best} in bold, \underline{second}{} underlined, per row and subset.}
\label{app:t3-speed}
\centering
\footnotesize
\setlength{\tabcolsep}{2.0pt}
\begin{tabular*}{\textwidth}{@{\extracolsep{\fill}}ll*{12}{c}@{}}
\toprule
 & & \multicolumn{6}{c}{Icon (OmniLottie-3B: real)} & \multicolumn{6}{c}{Illustration (OmniLottie-3B: synthetic)} \\
\cmidrule(lr){3-8}\cmidrule(lr){9-14}
Model & Decoding & PLD & EAGLE-3 & DFlash & DARTree & DDTree & \textbf{Ours} & PLD & EAGLE-3 & DFlash & DARTree & DDTree & \textbf{Ours} \\
\midrule
SVGen-3B & greedy & 1.39$\times$ & 2.93$\times$ & 4.96$\times$ & 5.47$\times$ & \underline{6.61}$\times$ & \textbf{7.37}$\times$ & 1.48$\times$ & 3.16$\times$ & 5.34$\times$ & 5.95$\times$ & \underline{6.95}$\times$ & \textbf{8.10}$\times$ \\
 & official sampling & 1.42$\times$ & 2.88$\times$ & 4.32$\times$ & 5.14$\times$ & \underline{5.65}$\times$ & \textbf{6.34}$\times$ & 1.46$\times$ & 3.00$\times$ & 4.52$\times$ & 5.43$\times$ & \underline{5.88}$\times$ & \textbf{6.86}$\times$ \\
\addlinespace[2pt]
SVGen-7B & greedy & 1.47$\times$ & 3.41$\times$ & 4.94$\times$ & 6.44$\times$ & \underline{6.98}$\times$ & \textbf{8.25}$\times$ & 1.49$\times$ & 3.39$\times$ & 4.74$\times$ & 6.26$\times$ & \underline{6.72}$\times$ & \textbf{7.81}$\times$ \\
 & official sampling & 1.46$\times$ & 3.28$\times$ & 4.42$\times$ & 5.68$\times$ & \underline{5.98}$\times$ & \textbf{7.07}$\times$ & 1.49$\times$ & 3.28$\times$ & 4.27$\times$ & 5.61$\times$ & \underline{5.86}$\times$ & \textbf{6.74}$\times$ \\
\addlinespace[2pt]
HiVG-3B & greedy & 3.93$\times$ & 4.46$\times$ & 8.48$\times$ & 8.29$\times$ & \underline{9.26}$\times$ & \textbf{13.78}$\times$ & 5.14$\times$ & 4.76$\times$ & 9.65$\times$ & 8.70$\times$ & \underline{9.90}$\times$ & \textbf{15.97}$\times$ \\
 & official sampling & 2.74$\times$ & 3.77$\times$ & 5.57$\times$ & 6.04$\times$ & \underline{6.72}$\times$ & \textbf{7.99}$\times$ & 3.00$\times$ & 3.89$\times$ & 5.78$\times$ & 6.33$\times$ & \underline{7.04}$\times$ & \textbf{8.83}$\times$ \\
\addlinespace[2pt]
OmniLottie-3B & greedy & 2.07$\times$ & 3.33$\times$ & 5.62$\times$ & 6.53$\times$ & \underline{7.00}$\times$ & \textbf{8.61}$\times$ & 1.92$\times$ & 3.44$\times$ & 4.64$\times$ & 6.19$\times$ & \underline{6.71}$\times$ & \textbf{7.73}$\times$ \\
 & official sampling & 2.01$\times$ & 3.16$\times$ & 4.95$\times$ & 5.73$\times$ & \underline{6.15}$\times$ & \textbf{7.18}$\times$ & 1.94$\times$ & 3.07$\times$ & 4.73$\times$ & 5.41$\times$ & \underline{6.19}$\times$ & \textbf{6.83}$\times$ \\
\addlinespace[2pt]
StarVector-1B & greedy & 1.99$\times$ & 1.90$\times$ & 4.14$\times$ & 4.23$\times$ & \underline{5.18}$\times$ & \textbf{6.35}$\times$ & 2.78$\times$ & 2.11$\times$ & 5.20$\times$ & 4.72$\times$ & \underline{5.87}$\times$ & \textbf{7.98}$\times$ \\
 & official sampling & 1.75$\times$ & 1.82$\times$ & 3.37$\times$ & 3.59$\times$ & \underline{4.23}$\times$ & \textbf{4.84}$\times$ & 2.35$\times$ & 1.99$\times$ & 4.05$\times$ & 3.92$\times$ & \underline{4.62}$\times$ & \textbf{5.63}$\times$ \\
 & sampling, $T$ 1.0 & 1.51$\times$ & 1.43$\times$ & 2.30$\times$ & 2.72$\times$ & \underline{3.44}$\times$ & \textbf{3.65}$\times$ & 1.52$\times$ & 1.32$\times$ & 1.99$\times$ & 2.39$\times$ & \underline{3.15}$\times$ & \textbf{3.25}$\times$ \\
\addlinespace[2pt]
StarVector-8B & greedy & 2.14$\times$ & 2.82$\times$ & 4.69$\times$ & 4.75$\times$ & \underline{5.73}$\times$ & \textbf{6.71}$\times$ & 2.86$\times$ & 2.86$\times$ & 4.93$\times$ & 4.94$\times$ & \underline{5.16}$\times$ & \textbf{6.54}$\times$ \\
 & official sampling & 1.72$\times$ & 1.95$\times$ & 2.43$\times$ & 2.81$\times$ & \textbf{3.41}$\times$ & \underline{3.35}$\times$ & 1.96$\times$ & 1.91$\times$ & 2.46$\times$ & 3.05$\times$ & \underline{3.20}$\times$ & \textbf{3.71}$\times$ \\
 & sampling, $T$ 1.0 & 1.47$\times$ & 1.23$\times$ & 1.72$\times$ & 1.84$\times$ & \underline{2.33}$\times$ & \textbf{2.34}$\times$ & \textbf{1.56}$\times$ & 0.85$\times$ & 1.19$\times$ & 1.25$\times$ & \underline{1.48}$\times$ & 1.46$\times$ \\
\bottomrule
\end{tabular*}
\end{table*}

\begin{table*}[t]
\caption{\textbf{Figure~\ref{fig:ex1}b in full, tokens per round.} $\tau$ over all tokens. StarVector's last row samples at $T$ 1.0 and top-$p$ 0.9, the defaults of its released generation code, and its official row uses its released evaluation configuration. \textbf{Best} in bold, \underline{second}{} underlined, per row and subset.}
\label{app:t3-tau}
\centering
\footnotesize
\setlength{\tabcolsep}{2.0pt}
\begin{tabular*}{\textwidth}{@{\extracolsep{\fill}}ll*{12}{c}@{}}
\toprule
 & & \multicolumn{6}{c}{Icon (OmniLottie-3B: real)} & \multicolumn{6}{c}{Illustration (OmniLottie-3B: synthetic)} \\
\cmidrule(lr){3-8}\cmidrule(lr){9-14}
Model & Decoding & PLD & EAGLE-3 & DFlash & DARTree & DDTree & \textbf{Ours} & PLD & EAGLE-3 & DFlash & DARTree & DDTree & \textbf{Ours} \\
\midrule
SVGen-3B & greedy & 1.61 & 8.87 & 7.34 & \underline{10.72} & 10.27 & \textbf{12.00} & 1.69 & 9.40 & 7.95 & \underline{11.42} & 10.90 & \textbf{13.16} \\
 & official sampling & 1.63 & 8.90 & 7.16 & \underline{10.86} & 10.21 & \textbf{11.67} & 1.68 & 9.33 & 7.62 & \underline{11.54} & 10.86 & \textbf{12.87} \\
\addlinespace[2pt]
SVGen-7B & greedy & 1.69 & 10.12 & 7.51 & \underline{12.41} & 11.34 & \textbf{13.63} & 1.72 & 9.96 & 7.20 & \underline{12.16} & 11.04 & \textbf{12.95} \\
 & official sampling & 1.69 & 10.01 & 7.42 & \underline{12.29} & 11.20 & \textbf{13.66} & 1.72 & 10.05 & 7.16 & \underline{12.09} & 11.07 & \textbf{13.01} \\
\addlinespace[2pt]
HiVG-3B & greedy & 4.61 & 11.91 & 12.09 & \underline{15.17} & 13.85 & \textbf{21.66} & 6.05 & 12.47 & 13.77 & \underline{15.85} & 14.87 & \textbf{25.11} \\
 & official sampling & 3.32 & 10.50 & 8.90 & \underline{12.51} & 11.72 & \textbf{14.87} & 3.68 & 10.90 & 9.22 & \underline{13.26} & 12.28 & \textbf{16.87} \\
\addlinespace[2pt]
OmniLottie-3B & greedy & 2.41 & 8.70 & 8.14 & \underline{11.43} & 10.64 & \textbf{13.46} & 2.24 & 8.62 & 6.64 & \underline{10.70} & 10.16 & \textbf{11.94} \\
 & official sampling & 2.40 & 8.51 & 7.97 & \underline{11.44} & 10.90 & \textbf{13.39} & 2.32 & 8.45 & 7.59 & 10.94 & \underline{10.99} & \textbf{12.57} \\
\addlinespace[2pt]
StarVector-1B & greedy & 2.48 & 10.27 & 8.41 & \underline{12.98} & 11.53 & \textbf{14.60} & 3.48 & 11.31 & 10.59 & \underline{14.49} & 13.19 & \textbf{18.46} \\
 & official sampling & 2.28 & 9.85 & 7.61 & \underline{12.15} & 10.89 & \textbf{13.52} & 3.14 & 10.58 & 9.44 & \underline{13.27} & 12.28 & \textbf{16.32} \\
 & sampling, $T$ 1.0 & 1.88 & 7.38 & 4.72 & \underline{8.41} & 8.07 & \textbf{8.65} & 1.89 & 6.83 & 4.00 & \underline{7.33} & 7.26 & \textbf{7.62} \\
\addlinespace[2pt]
StarVector-8B & greedy & 2.86 & 9.70 & 8.10 & \underline{11.11} & 10.83 & \textbf{12.80} & 3.96 & 9.83 & 8.79 & \underline{11.70} & 10.04 & \textbf{12.91} \\
 & official sampling & 2.30 & 6.73 & 4.37 & 6.70 & \textbf{6.80} & \underline{6.75} & 2.71 & 6.54 & 4.49 & \underline{7.35} & 6.50 & \textbf{7.66} \\
 & sampling, $T$ 1.0 & 2.05 & 4.17 & 3.05 & 4.39 & \underline{4.58} & \textbf{4.60} & 2.20 & 2.86 & 2.12 & \underline{2.90} & \textbf{2.91} & 2.88 \\
\bottomrule
\end{tabular*}
\end{table*}

\begin{table*}[t]
\caption{\textbf{The six models of Figure~\ref{fig:ex1}b.} Official sampling is each model's released setting, and StarVector uses its released evaluation configuration with the library's default top-$k$. Prompts per subset are icon and illustration (OmniLottie-3B, real and synthetic). Every trained draft has the same budget, 100k prompts and 2 epochs.}
\label{app:t3-settings}
\centering
\footnotesize
\begin{tabular*}{\textwidth}{@{\extracolsep{\fill}}llllll@{}}
\toprule
Model & Backbone & Output tokens & Task & Official sampling & Prompts \\
\midrule
SVGen-3B & Llama-3.2-3B & raw SVG code & text-to-SVG & T 0.2, top-p 0.7 & 300 / 300 \\
SVGen-7B & Qwen2.5-Coder-7B & raw SVG code & text-to-SVG & T 0.2, top-p 0.7 & 300 / 300 \\
HiVG-3B & Qwen2.5-VL-3B & hierarchical compressed tokens & text-to-SVG & T 0.7, top-p 0.9, top-k 50 & 300 / 300 \\
OmniLottie-3B & Qwen2.5-VL-3B & Lottie JSON tokens & text-to-Lottie & T 0.9, top-p 0.25, top-k 5 & 150 / 150 \\
StarVector-1B & StarCoderBase-1B + CLIP & raw SVG code & image-to-SVG & T 0.2, top-p 0.95, top-k 50 & 300 / 300 \\
StarVector-8B & StarCoder2-7B + SigLIP & raw SVG code & image-to-SVG & T 0.7, top-p 0.95, top-k 50 & 300 / 300 \\
\bottomrule
\end{tabular*}
\end{table*}

\subsection{Losslessness}
\label{app:lossless}
Table~\ref{app:fp32} lists the audits behind Section~\ref{sec:setup}.
Every audited decoder writes the same tokens as the original model in fp32.
In bf16, the timed runs part from the original model's outputs at near-ties.
Within each model--task pair, every decoder keeps about the same share of its outputs identical to the original model's (Table~\ref{app:bf16}), including PLD and REST, which train no drafter and change only the shape of the verification pass.
The parting thus comes from the shape of the pass, and it causes the small quality differences within the lossless groups of Table~\ref{tab:quality}.

\begin{table*}[t]
\caption{\textbf{Token-for-token audits in fp32.} Prompts re-decoded in fp32 by each method under the verifier of Section~\ref{sec:prelim} and by the original model. A cell counts the prompt $\times$ draft design checks, and every one of them gave identical tokens. In all, 102 audits run 1,101 checks over 1,051,036 target tokens without a mismatch. A dash marks a decoder that ran but was not audited, and n/a one that does not run on that model. The verifier is the same code for every method and model.}
\label{app:fp32}
\centering
\footnotesize
\setlength{\tabcolsep}{2.2pt}
\begin{tabular*}{\textwidth}{@{\extracolsep{\fill}}l*{8}{c}@{}}
\toprule
 & OmniSVG-8B & OmniSVG-4B & InternSVG-8B & OmniSVG-8B & InternSVG-8B & StarVector-1B & StarVector-8B & OmniLottie-3B \\
Method & text & text & text & image & image & image & image & text \\
\midrule
PLD & 10 & 10 & 10 & 10 & 10 & 10 & 10 & 10 \\
REST & 10 & 10 & 10 & 10 & 10 & n/a & n/a & n/a \\
SpS & -- & n/a & n/a & n/a & n/a & n/a & n/a & n/a \\
EAGLE-3 & 30 & 10 & 30 & 12 & 12 & 10 & 10 & 10 \\
ViSpec & n/a & n/a & n/a & 30 & 10 & n/a & n/a & n/a \\
DFlash & 77 & 10 & 10 & 10 & 10 & 10 & 10 & 10 \\
DARTree & 20 & 12 & 12 & 34 & 12 & 10 & 10 & 10 \\
DDTree & 20 & 10 & 10 & 10 & 10 & 10 & 10 & 10 \\
\textbf{Ours} & 110 & 10 & 20 & 60 & 14 & 10 & 10 & 10 \\
Ours + loop control & 44 & n/a & 16 & 30 & 10 & n/a & 10 & n/a \\
\bottomrule
\end{tabular*}
\end{table*}

\begin{table*}[t]
\caption{\textbf{Token identity in bf16.} Share of greedy outputs, in the runs of Figure~\ref{fig:main}, whose tokens equal those of the original model decoded in bf16 in the same job, per decoder and model--task pair. PLD and REST train no drafter and change only the shape of the verification pass. A dash marks a decoder whose rows have no reference output of the original model. 50,400 outputs in all.}
\label{app:bf16}
\centering
\footnotesize
\setlength{\tabcolsep}{2.4pt}
\begin{tabular*}{\textwidth}{@{\extracolsep{\fill}}l*{9}{c}@{}}
\toprule
Model and task & PLD & REST & SpS & EAGLE-3 & ViSpec & DFlash & DDTree & DARTree & \textbf{Ours} \\
\midrule
OmniSVG-8B, text-to-SVG & 57\% & 58\% & 59\% & 57\% & -- & 60\% & 59\% & 59\% & 59\% \\
OmniSVG-4B, text-to-SVG & -- & 42\% & -- & -- & -- & -- & -- & 41\% & 41\% \\
InternSVG-8B, text-to-SVG & -- & 30\% & -- & -- & -- & -- & -- & 29\% & 32\% \\
OmniSVG-8B, image-to-SVG & 47\% & 47\% & -- & 48\% & 47\% & 47\% & 47\% & 46\% & 47\% \\
InternSVG-8B, image-to-SVG & 14\% & 14\% & -- & 15\% & 17\% & 16\% & 15\% & 14\% & 15\% \\
\bottomrule
\end{tabular*}
\end{table*}

\FloatBarrier
\section{Sources of the Lead}
\label{app:lead}

\subsection{Misses by Token Class}
\label{app:classes}
Figure~\ref{fig:ex1}c gives the miss rate of our draft head on structure and on numbers, the classes of Appendix~\ref{app:theory-setup}, under greedy decoding on loop-free content.
In every pair, numbers miss two to eight times as often as structure.
They make up 59\% to 66\% of the tokens but 78\% to 91\% of the misses, so most rounds that stop at a miss stop at a number.
Figure~\ref{app:row-theory}c marks the rounds on one real output.

\subsection{Rounds That Fill the Window}
\label{app:rounds}
\begin{figure*}[t]
\centering
\includegraphics{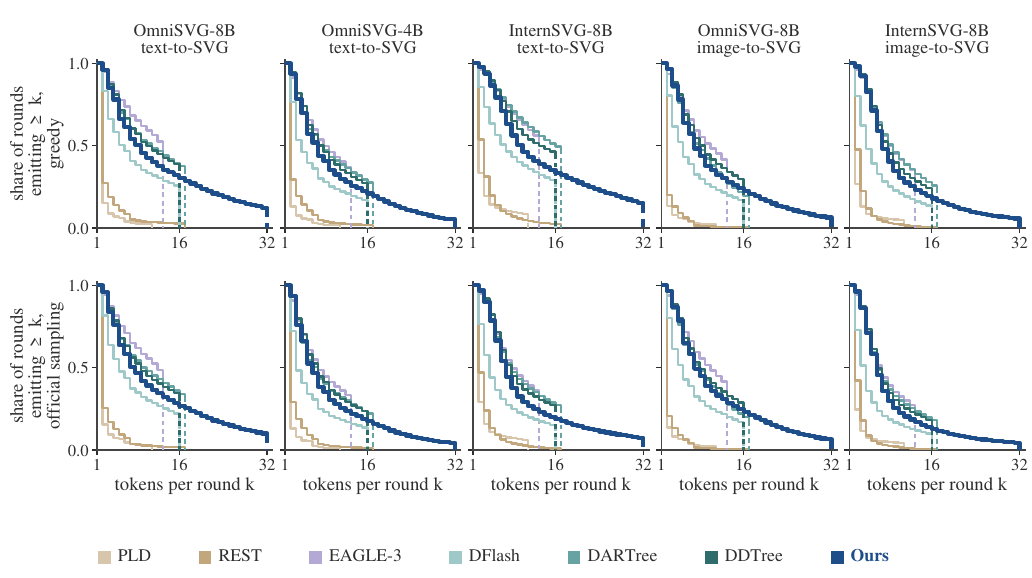}
\caption{\textbf{Survival of rounds over position.} Share of rounds that write at least $k$ tokens, for every decoder and model--task pair, under greedy decoding (top) and official sampling (bottom), loop-free content. Dashed lines mark each decoder's window.}
\label{app:m2b}
\end{figure*}

Figure~\ref{fig:ex1}d pools the rounds of the five model--task pairs under greedy decoding on loop-free content.
Figure~\ref{app:m2b} gives, for every decoder, the share of rounds that write at least $k$ tokens.
The curve of each existing decoder drops at its own window, and ours continues to 32.

Figure~\ref{fig:ex1}e splits the outputs of each target's own decoding into those that end, those that reach the length limit, and those that reach it inside a loop.
The offline check looks at the outputs that reach the length limit.
Such an output is in a loop when its last 384 tokens repeat with a period of at most 256 tokens at 70\% or more of their positions.
The loop starts where this repetition begins.
Under greedy decoding, 25\% to 47\% of the outputs of the SVG generation models end in a loop, against 6\% for OmniLottie-3B, which writes Lottie JSON.

\subsection{Tokens and Time per Round}
\label{app:round-cost}
Loop-free content counts the rounds that start before the output enters a loop, and the other rounds fall inside loops, with the start of a loop found as in Appendix~\ref{app:rounds}.
Averaged over the two decodings, our tokens per round are 1.1 to 1.2 times those of DDTree on loop-free content and 1.5 to 1.7 times inside loops (Figure~\ref{fig:ex2}a).
Table~\ref{app:timing-split} splits the time of one round by stage.
The verification pass costs about the same for every method, and drafting all positions of a block in one pass keeps the drafting time of a longer window small.
EAGLE-3 writes about as many tokens per round as the block drafters, but it drafts one token per pass, so its rounds are the slowest (Figure~\ref{fig:ex2}b).

\begin{table}[htbp]\centering
\begingroup
\renewenvironment{table}[1][]{\begin{minipage}[t]{0.62\textwidth}\centering}{\end{minipage}}%
\begin{table}[t]
\caption{\textbf{Time of one round, by stage.} OmniSVG-8B text-to-SVG, 20 icons and 20 illustrations, greedy decoding, one H200, where the original model takes 10.92\,ms per token on the same prompts. The timing is instrumented, so the speedups differ slightly from Figure~\ref{fig:main}. The verification pass costs about the same for every method, and EAGLE-3 spends more on drafting one token at a time. \textbf{Best} in bold, \underline{second}{} underlined, for $\tau$ and speedup.}
\label{app:timing-split}
\centering
\footnotesize
\setlength{\tabcolsep}{1.6pt}
\begin{tabular*}{\columnwidth}{@{\extracolsep{\fill}}l*{7}{c}@{}}
\toprule
 & & & \multicolumn{4}{c}{ms per round, split} & \\
\cmidrule(lr){4-7}
Method & $\tau$ & ms / round & draft & tree build & verify & other & speedup \\
\midrule
PLD & 2.56 & 13.1 & 0.0 & 0.0 & 11.9 & 1.2 & 2.14$\times$ \\
REST & 2.74 & 13.6 & 0.2 & 0.0 & 12.1 & 1.2 & 2.20$\times$ \\
EAGLE-3 & 11.33 & 35.9 & 20.9 & 0.0 & 12.3 & 2.8 & 3.44$\times$ \\
DFlash & 10.36 & 18.1 & 3.5 & 0.1 & 12.0 & 2.5 & 6.26$\times$ \\
DARTree & \underline{13.59} & 23.2 & 7.9 & 0.0 & 12.3 & 3.1 & 6.38$\times$ \\
DDTree & 12.85 & 20.2 & 3.5 & 1.6 & 12.2 & 3.0 & \underline{6.93}$\times$ \\
\textbf{Ours} & \textbf{19.02} & 21.5 & 3.6 & 1.7 & 12.3 & 3.9 & \textbf{9.66}$\times$ \\
\bottomrule
\end{tabular*}
\end{table}
\endgroup
\end{table}

\FloatBarrier
\section{Training and Design}
\label{app:design}

\subsection{Training Length and Warm Start}
\label{app:train-length}
Figure~\ref{fig:ex2}c compares tokens per round on loop-free content only, since the evaluations of epochs 2 to 5 reused outputs of the original model timed on other nodes.
On OmniSVG-8B text-to-SVG, our head trains on 300,000 prompts with the budget of DFlash's head, and its tokens per round grow by 33\% from the first epoch to the fourth.
The fifth and sixth epochs stay within 1\% of the best.
For image-to-SVG, three epochs on 100,000 images give 10.06 tokens per round with a warm start from the text head.
The text head as is gives 8.77 and training from scratch 8.65, so the warm start gains 15\%.

\subsection{Window and Tree}
\label{app:window-tree}
Figure~\ref{fig:ex2}d adds the two changes of Sightline to DFlash's recipe and compares tokens per round on loop-free content only, since its four variants ran on different nodes.
The tree alone gains 1.37 to 1.49 times and the window of 32 alone 1.07 to 1.13 times.
Together they gain 1.67 to 1.73 times, more than the product of their separate gains, 1.54 to 1.59 times.
The InternSVG-8B rows of Figure~\ref{fig:ex2}d use the heads of its first round of training (Table~\ref{app:hparams}), before the continued training of the heads in Figure~\ref{fig:main}.

Figure~\ref{fig:ex2}f simulates the rounds offline from cached drafts of OmniSVG-8B text-to-SVG, adding nodes in the order of \eqref{eq:tree}.
By its 87th node, our tree accepts as many tokens per round as DDTree does with all 127.
The other 40 nodes sit at positions 16 to 31, where a window of 16 never drafts, and they add 4.8 tokens per round.
Under greedy decoding on the three text-to-SVG pairs, our tree spends 6.7 to 8.7 nodes per token written and DDTree's 9.6 to 11.5.

\subsection{Block Length}
\label{app:block-length}
Figure~\ref{fig:ex2}e compares heads trained on blocks of 16, 32 and 64 for OmniSVG-8B text-to-SVG, with the head of block length 16 trained for six epochs and the others for three.
From a block length of 32 to 64, all tokens gain 5\% under greedy decoding and 1\% under official sampling, while loop-free content loses 1\% and 4\%.
The added tokens thus all come from inside loops, where a longer window writes more per round.

\subsection{Tree Shape}
\label{app:tree-shape}
\begin{figure*}[t]
\centering
\includegraphics{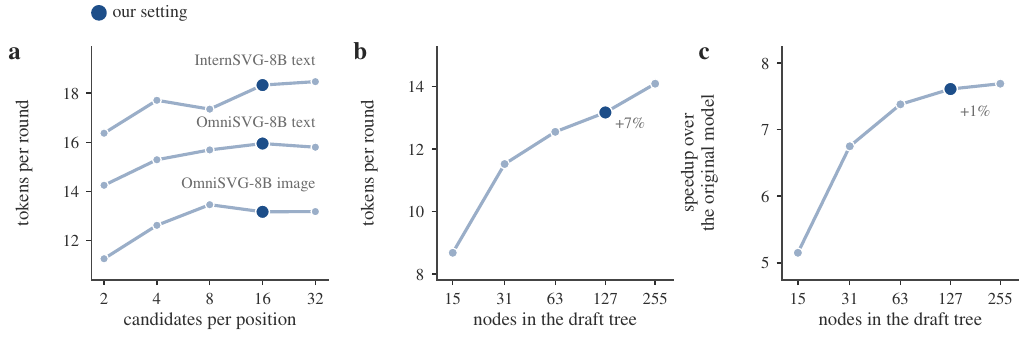}
\caption{\textbf{Shape of the draft tree.} (a)~Tokens per round of a 127-node tree as the candidates per position vary, for three model--task pairs. (b,~c)~Tokens per round and speedup over the original model as the node budget grows, OmniSVG-8B image-to-SVG. Greedy decoding on the 300 prompts of MMSVG, with our setting in dark blue.}
\label{app:tree-fig}
\end{figure*}

Figure~\ref{app:tree-fig}a varies the candidates per position of a 127-node tree.
From 16 on, every curve stays within 1\% of its better value at 16 or 32, so we take the narrowest width on this plateau.
Figure~\ref{app:tree-fig}b and c vary the node budget.
From 127 to 255 nodes, the tokens per round rise by 7\%, while the speedup rises by 1\%, since a larger tree takes longer to verify.

\subsection{Certainty of the Target}
\label{app:certainty}
Acceptance follows the target's own certainty, and most positions are near-deterministic (Table~\ref{app:entropy}).

\begin{table*}[t]
\caption{\textbf{Acceptance follows the target's own certainty.} For every generated position, the entropy of the target's next-token distribution and whether our draft's token was kept, under greedy decoding on the outputs of Figure~\ref{fig:main}. Between 65\% and 73\% of positions are near-deterministic, below 0.01 nats, and the draft is kept at 95\% to 98\% of them.}
\label{app:entropy}
\centering
\footnotesize
\setlength{\tabcolsep}{3.0pt}
\begin{tabular*}{\textwidth}{@{\extracolsep{\fill}}l*{6}{c}@{}}
\toprule
 & \multicolumn{2}{c}{OmniSVG-8B, text-to-SVG} & \multicolumn{2}{c}{OmniSVG-8B, image-to-SVG} & \multicolumn{2}{c}{InternSVG-8B, text-to-SVG} \\
\cmidrule(lr){2-3}\cmidrule(lr){4-5}\cmidrule(lr){6-7}
Target entropy (nats) & positions (\%) & accepted & positions (\%) & accepted & positions (\%) & accepted \\
\midrule
0--0.01 & 73.1 & 0.968 & 72.6 & 0.954 & 65.3 & 0.976 \\
0.01--0.05 & 10.1 & 0.956 & 9.0 & 0.930 & 12.5 & 0.968 \\
0.05--0.2 & 6.2 & 0.929 & 6.7 & 0.907 & 7.7 & 0.946 \\
0.2--0.5 & 3.4 & 0.892 & 3.8 & 0.855 & 4.2 & 0.922 \\
0.5--1 & 3.0 & 0.849 & 3.7 & 0.832 & 4.1 & 0.869 \\
1--2 & 1.5 & 0.865 & 1.5 & 0.823 & 4.6 & 0.858 \\
2--4 & 0.8 & 0.883 & 0.8 & 0.840 & 1.4 & 0.835 \\
$\geq$4 & 1.9 & 0.921 & 1.9 & 0.895 & 0.1 & 0.848 \\
\bottomrule
\end{tabular*}
\end{table*}

\FloatBarrier
\section{Loop Control}
\label{app:loop}

\subsection{Three Rules}
\label{app:loop-cmp}
Table~\ref{app:loop-rules} compares three rules on the same outputs, rolling back before the loop, escaping once and then stopping, which we use, and stopping at once.
Figure~\ref{fig:ex2}g sets them against the lossless run, which does nothing when the rule fires.
Rollback deletes all but one copy of the cycle and bans the token that would restart it, and it stops the output at the next firing.
Escape then stop bans that token in place at the first firing and stops the output at the next, as \eqref{eq:loop} does with $m=1$.
Stop ends the output at the first firing.
Escape then stop saves most of the forward passes that stopping saves, and its quality stays within the paired interval of the lossless run (Figure~\ref{fig:ex2}g).

\begin{table*}[t]
\caption{\textbf{The three loop rules in full.} Each rule against the lossless run of the same draft on the same 900 prompts, 300 for InternSVG-8B image-to-SVG. Columns give the outputs that end normally, the outputs still in a loop, the mean length, the target forward passes saved, and the change in the main quality metric with its paired 95\% bootstrap interval. We use escape then stop. A strict periodicity detector counts the outputs still in a loop.}
\label{app:loop-rules}
\centering
\footnotesize
\setlength{\tabcolsep}{3.0pt}
\begin{tabular*}{\textwidth}{@{\extracolsep{\fill}}l*{7}{c}@{}}
\toprule
 & ended & still in & mean & fewer & \multicolumn{3}{c}{quality vs lossless} \\
\cmidrule(lr){6-8}
Rule & normally (\%) & a loop (\%) & tokens & forwards & $\Delta$ & 95\% low & 95\% high \\
\midrule
\multicolumn{8}{l}{\textit{OmniSVG-8B, text-to-SVG, greedy} (quality: CLIP)} \\
lossless & 64 & 27 & 756 & 1.00$\times$ & \multicolumn{3}{c}{reference (0.272)} \\
rollback & 90 & 2 & 469 & 1.20$\times$ & -0.0002 & -0.0009 & +0.0004 \\
\textbf{escape then stop} & 90 & 2 & 469 & 1.25$\times$ & -0.0006 & -0.0012 & 0.0000 \\
stop & 91 & 2 & 447 & 1.32$\times$ & -0.0007 & -0.0013 & -0.0002 \\
\addlinespace[2pt]
\multicolumn{8}{l}{\textit{OmniSVG-8B, text-to-SVG, official sampling} (quality: CLIP)} \\
lossless & 67 & 23 & 772 & 1.00$\times$ & \multicolumn{3}{c}{reference (0.272)} \\
rollback & 87 & 3 & 547 & 1.10$\times$ & +0.0002 & -0.0005 & +0.0010 \\
\textbf{escape then stop} & 89 & 2 & 522 & 1.17$\times$ & -0.0003 & -0.0010 & +0.0004 \\
stop & 91 & 2 & 483 & 1.25$\times$ & -0.0006 & -0.0013 & -0.0001 \\
\addlinespace[2pt]
\multicolumn{8}{l}{\textit{InternSVG-8B, text-to-SVG, greedy} (quality: CLIP)} \\
lossless & 49 & 33 & 2352 & 1.00$\times$ & \multicolumn{3}{c}{reference (0.269)} \\
rollback & 70 & 11 & 1672 & 1.16$\times$ & +0.0015 & +0.0003 & +0.0028 \\
\textbf{escape then stop} & 73 & 10 & 1569 & 1.25$\times$ & -0.0009 & -0.0018 & +0.0001 \\
stop & 75 & 10 & 1464 & 1.34$\times$ & -0.0017 & -0.0026 & -0.0009 \\
\addlinespace[2pt]
\multicolumn{8}{l}{\textit{InternSVG-8B, text-to-SVG, official sampling} (quality: CLIP)} \\
lossless & 72 & 0 & 2091 & 1.00$\times$ & \multicolumn{3}{c}{reference (0.288)} \\
rollback & 72 & 0 & 2088 & 1.00$\times$ & -0.0002 & -0.0006 & +0.0001 \\
\textbf{escape then stop} & 72 & 0 & 2086 & 1.00$\times$ & -0.0001 & -0.0004 & 0.0000 \\
stop & 72 & 0 & 2077 & 1.01$\times$ & -0.0002 & -0.0006 & 0.0000 \\
\addlinespace[2pt]
\multicolumn{8}{l}{\textit{OmniSVG-8B, image-to-SVG, greedy} (quality: DINO)} \\
lossless & 65 & 26 & 798 & 1.00$\times$ & \multicolumn{3}{c}{reference (0.911)} \\
rollback & 88 & 2 & 571 & 1.05$\times$ & +0.0051 & +0.0030 & +0.0076 \\
\textbf{escape then stop} & 90 & 3 & 544 & 1.13$\times$ & +0.0024 & +0.0005 & +0.0045 \\
stop & 93 & 2 & 485 & 1.24$\times$ & -0.0022 & -0.0037 & -0.0007 \\
\addlinespace[2pt]
\multicolumn{8}{l}{\textit{OmniSVG-8B, image-to-SVG, official sampling} (quality: DINO)} \\
lossless & 65 & 26 & 808 & 1.00$\times$ & \multicolumn{3}{c}{reference (0.911)} \\
rollback & 86 & 3 & 596 & 1.05$\times$ & +0.0045 & +0.0025 & +0.0066 \\
\textbf{escape then stop} & 88 & 3 & 570 & 1.12$\times$ & +0.0018 & -0.0001 & +0.0040 \\
stop & 90 & 2 & 514 & 1.22$\times$ & -0.0025 & -0.0040 & -0.0011 \\
\addlinespace[2pt]
\multicolumn{8}{l}{\textit{InternSVG-8B, image-to-SVG, greedy} (quality: DINO)} \\
lossless & 72 & 23 & 1926 & 1.00$\times$ & \multicolumn{3}{c}{reference (0.940)} \\
rollback & 93 & 0 & 1285 & 1.15$\times$ & +0.0047 & -0.0001 & +0.0096 \\
\textbf{escape then stop} & 93 & 0 & 1299 & 1.18$\times$ & +0.0012 & -0.0044 & +0.0063 \\
stop & 95 & 0 & 1128 & 1.32$\times$ & -0.0044 & -0.0076 & -0.0018 \\
\addlinespace[2pt]
\multicolumn{8}{l}{\textit{InternSVG-8B, image-to-SVG, official sampling} (quality: DINO)} \\
lossless & 86 & 8 & 1640 & 1.00$\times$ & \multicolumn{3}{c}{reference (0.956)} \\
rollback & 95 & 0 & 1407 & 1.04$\times$ & -0.0007 & -0.0036 & +0.0024 \\
\textbf{escape then stop} & 95 & 0 & 1417 & 1.05$\times$ & -0.0021 & -0.0056 & +0.0009 \\
stop & 96 & 0 & 1318 & 1.12$\times$ & -0.0056 & -0.0093 & -0.0024 \\
\bottomrule
\end{tabular*}
\end{table*}

\subsection{Stopped Outputs}
\label{app:loop-stopped}
By an offline check, 67\% to 87\% of the outputs that loop control stops are already in a loop (Figure~\ref{app:a15}).

\sidecapfig{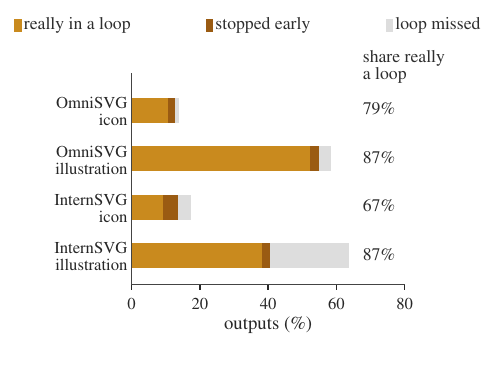}{\textbf{Outputs that loop control stops.} For each model and set, the outputs where loop control fires, split by an offline check into outputs in a loop and outputs stopped early, with the loops it does not catch in gray. The number gives the share of stopped outputs that are in a loop.}{app:a15}

\FloatBarrier
\section{The Best of Eight}
\label{app:quality}
Every best of $N$ here gives the original model and Sightline the same loop control and the same rendered selection, so the rows differ only in the decoder.

\subsection{Protocol and Cost}
\label{app:bo8-protocol}
\begin{figure*}[t]
\centering
\includegraphics{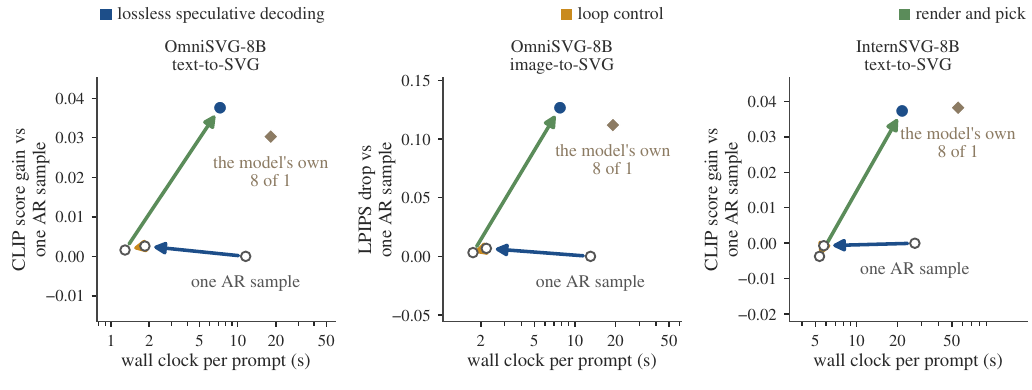}
\caption{\textbf{From one sample to the best of eight.} Quality against seconds per prompt for three model--task pairs, starting from one sample of the original model. Lossless speculative decoding and loop control move the point left, rendered selection moves it up, and the star marks the original model's own best of eight.}
\label{app:a11}
\end{figure*}

Figure~\ref{fig:ex2}h counts the forward passes of the target per prompt, as a share of the passes of one sample of the original model, which makes one pass per token.
A batch makes one pass per step for all of its samples and runs until its longest sample ends.
The seconds of Figure~\ref{fig:ex2}i include drawing, rendering and scoring every sample.
Figure~\ref{app:a11} takes the steps from one sample to the best of eight one at a time.
From one sample of the original model, lossless speculation and loop control move the point left, and rendered selection moves it up, above and to the left of the original model's best of eight.
On OmniSVG-8B text-to-SVG, keeping the best of the first $k$ samples to finish adds two points to the frontier of quality against time (Table~\ref{app:first-k}).

\begin{table*}[t]
\caption{\textbf{Best of the first $k$ to finish.} $N$ candidates are decoded together, and once $k$ of them have finished, the rest are dropped and the best finished one is kept. Seconds per prompt to decode, render and score on one H200, and the quality change against one sample of the original model with its paired 95\% bootstrap interval, 300 prompts per setting.}
\label{app:first-k}
\centering
\footnotesize
\setlength{\tabcolsep}{2.2pt}
\begin{tabular*}{\textwidth}{@{\extracolsep{\fill}}l*{12}{c}@{}}
\toprule
 & \multicolumn{4}{c}{OmniSVG-8B, text-to-SVG} & \multicolumn{4}{c}{OmniSVG-8B, image-to-SVG} & \multicolumn{4}{c}{InternSVG-8B, text-to-SVG} \\
\cmidrule(lr){2-5}\cmidrule(lr){6-9}\cmidrule(lr){10-13}
 & s & \multicolumn{3}{c}{CLIP gain (95\%)} & s & \multicolumn{3}{c}{LPIPS drop (95\%)} & s & \multicolumn{3}{c}{CLIP gain (95\%)} \\
\midrule
original model, 1 sample & 12.06 & 0.000 & 0.000 & 0.000 & 13.50 & 0.000 & 0.000 & 0.000 & 25.61 & 0.000 & 0.000 & 0.000 \\
\quad + loop control & 7.19 & 0.000 & -0.002 & +0.001 & 8.15 & -0.003 & -0.008 & +0.002 & 20.96 & -0.004 & -0.006 & -0.001 \\
\quad + loop control, best of 8 & 18.88 & +0.030 & +0.027 & +0.034 & 19.74 & +0.112 & +0.094 & +0.130 & 54.84 & +0.038 & +0.034 & +0.043 \\
\quad + loop control, first 6 of 8 & 12.27 & +0.027 & +0.023 & +0.030 & 13.28 & +0.100 & +0.082 & +0.117 & 46.12 & +0.035 & +0.031 & +0.040 \\
\quad + loop control, first 4 of 8 & 7.84 & +0.021 & +0.017 & +0.025 & 9.56 & +0.071 & +0.055 & +0.089 & 38.78 & +0.028 & +0.024 & +0.033 \\
\textbf{Ours} + loop control & 1.35 & +0.002 & -0.001 & +0.004 & 1.80 & +0.003 & -0.008 & +0.015 & 5.24 & -0.004 & -0.009 & +0.001 \\
\quad best of 8 & 4.84 & +0.031 & +0.027 & +0.035 & 5.08 & +0.105 & +0.087 & +0.122 & 21.80 & +0.037 & +0.033 & +0.042 \\
\quad first 6 of 8 & 2.91 & +0.027 & +0.023 & +0.030 & 3.52 & +0.085 & +0.064 & +0.104 & 17.75 & +0.030 & +0.025 & +0.035 \\
\quad first 4 of 8 & 1.91 & +0.020 & +0.016 & +0.024 & 2.71 & +0.061 & +0.040 & +0.080 & 14.87 & +0.023 & +0.018 & +0.027 \\
\quad best of 16 & 7.35 & +0.038 & +0.034 & +0.041 & 7.84 & +0.127 & +0.108 & +0.145 & 37.34 & +0.046 & +0.041 & +0.051 \\
\quad first 8 of 16 & 3.18 & +0.024 & +0.020 & +0.028 & 4.56 & +0.078 & +0.059 & +0.096 & 27.72 & +0.034 & +0.029 & +0.038 \\
\bottomrule
\end{tabular*}
\end{table*}

\subsection{Judges}
\label{app:judges}
\begin{figure*}[t]
\centering
\includegraphics{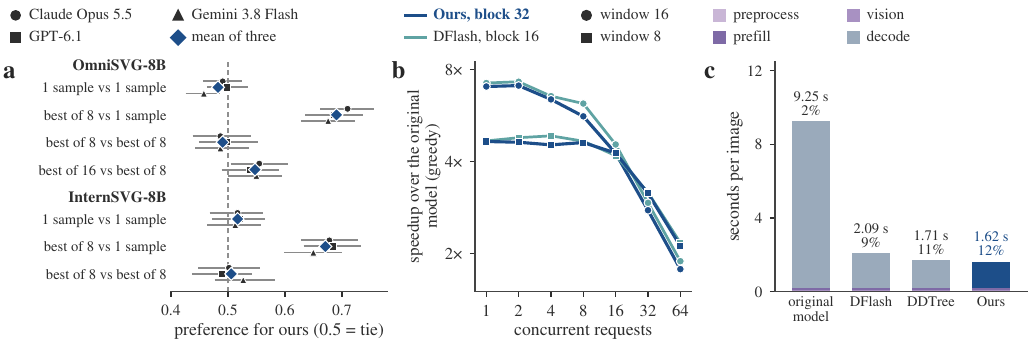}
\caption{\textbf{Judges, serving at one window, and one-off cost.} (a)~Preference for ours with its 95\% interval, per VLM judge, and the mean of the three as a diamond, for OmniSVG-8B and InternSVG-8B text-to-SVG. The first row of each model compares two single samples that follow the same distribution, so its ideal value is 0.5. (b)~Speedup over the original model under greedy decoding against the number of concurrent requests, for our head (block length 32) and DFlash's (block length 16), both verified at windows of 16 and 8. (c)~Seconds per image of OmniSVG-8B image-to-SVG by stage, preprocessing, vision encoding, prefill and decoding, with the share of the one-off stages on top of each bar.}
\label{app:row-judges}
\end{figure*}

Figure~\ref{app:row-judges}a gives each VLM judge of Figure~\ref{fig:ex2}j separately.
The calibration pair compares two single samples that follow the same distribution, so its ideal value is 0.5, and each judge's reading on it shows the judge's own bias.

\subsection{Selection Rules}
\label{app:selection}
Figure~\ref{fig:ex2}k compares five rules for choosing among eight samples.
Highest and lowest likelihood take the sample whose tokens have the highest or lowest mean log-probability under the target.
Text consensus takes the sample with the highest mean n-gram F1 score against the other seven, for $n$ from 1 to 4.
Pixel consensus takes the sample whose rendering, in gray at $64\times64$ pixels, has the lowest mean squared error to the other seven.
The render score is the scorer of rendered selection.
The panel counts only metrics that the render score does not use, the aesthetic score and HPS for text-to-SVG and SSIM, LPIPS and MSE for image-to-SVG.
Each gain over a random pick is divided by the standard deviation of the per-prompt difference, so that metrics of different scales share one axis.

Choosing by the rendering beats a random pick at three temperatures (Table~\ref{app:selection-temps}) and under every scorer we try, including metrics that the scorer does not use (Tables~\ref{app:scorer-swap-text} and~\ref{app:scorer-swap-image}).

\begin{table*}[t]
\caption{\textbf{Choosing among eight samples at three temperatures.} InternSVG-8B text-to-SVG, 300 prompts, eight samples each. Quality of the chosen sample minus a random pick, paired per prompt. CLIP is the render score's own signal, and Aesthetic and HPS are independent of it. The render score's 95\% bootstrap interval excludes zero in every column except Aesthetic at $T$ 1.0. \textbf{Best} in bold, \underline{second}{} underlined.}
\label{app:selection-temps}
\centering
\footnotesize
\setlength{\tabcolsep}{2.0pt}
\begin{tabular*}{\textwidth}{@{\extracolsep{\fill}}l*{9}{c}@{}}
\toprule
 & \multicolumn{3}{c}{$T$ 0.3} & \multicolumn{3}{c}{$T$ 0.6 (official)} & \multicolumn{3}{c}{$T$ 1.0} \\
\cmidrule(lr){2-4}\cmidrule(lr){5-7}\cmidrule(lr){8-10}
Pick by & CLIP & Aesthetic & HPS & CLIP & Aesthetic & HPS & CLIP & Aesthetic & HPS \\
\midrule
random pick & 0.000 & 0.000 & 0.000 & 0.000 & 0.000 & 0.000 & 0.000 & 0.000 & 0.000 \\
highest likelihood & -0.014 & -0.069 & -0.006 & -0.017 & -0.058 & -0.006 & -0.006 & \textbf{+0.035} & -0.001 \\
lowest likelihood & \underline{+0.009} & \underline{+0.030} & \underline{+0.003} & \underline{+0.006} & \underline{+0.049} & \underline{+0.003} & 0.000 & -0.039 & -0.002 \\
text consensus & +0.004 & +0.021 & \underline{+0.003} & +0.005 & +0.041 & \underline{+0.003} & \underline{+0.002} & +0.011 & \underline{+0.002} \\
\textbf{render score} & \textbf{+0.040} & \textbf{+0.070} & \textbf{+0.009} & \textbf{+0.041} & \textbf{+0.069} & \textbf{+0.009} & \textbf{+0.036} & \underline{+0.026} & \textbf{+0.007} \\
\bottomrule
\end{tabular*}
\end{table*}

\begin{table}[htbp]\centering
\begingroup
\renewenvironment{table}[1][]{\begin{minipage}[t]{0.485\textwidth}\vspace{0pt}\centering}{\end{minipage}}%
\renewenvironment{table*}[1][]{\begin{minipage}[t]{0.485\textwidth}\vspace{0pt}\centering}{\end{minipage}}%
\begin{table}[t]
\caption{\textbf{Swapping the scorer, text-to-SVG.} OmniSVG-8B text-to-SVG, 900 prompts, eight of our samples each. Each row picks a sample by one metric and scores it on every metric, minus a random pick. Every 95\% bootstrap interval is on the improving side.}
\label{app:scorer-swap-text}
\centering
\footnotesize
\setlength{\tabcolsep}{3.0pt}
\begin{tabular*}{\columnwidth}{@{\extracolsep{\fill}}l*{3}{c}@{}}
\toprule
Pick by & CLIP & Aesthetic & HPS \\
\midrule
CLIP & +0.0287 & +0.0221 & +0.0066 \\
Aesthetic & +0.0044 & +0.3484 & +0.0039 \\
HPS & +0.0158 & +0.1075 & +0.0117 \\
\bottomrule
\end{tabular*}
\end{table}
\hfill\begin{table}[t]
\caption{\textbf{Swapping the scorer, image-to-SVG.} OmniSVG-8B image-to-SVG, 900 prompts, eight of our samples each. Each row picks a sample by one metric and scores it on every metric, minus a random pick. Every 95\% bootstrap interval is on the improving side. LPIPS and MSE are better when lower.}
\label{app:scorer-swap-image}
\centering
\footnotesize
\setlength{\tabcolsep}{3.0pt}
\begin{tabular*}{\columnwidth}{@{\extracolsep{\fill}}l*{4}{c}@{}}
\toprule
Pick by & DINO & SSIM & LPIPS & MSE \\
\midrule
DINO & +0.0481 & +0.0360 & -0.0862 & -0.0270 \\
SSIM & +0.0298 & +0.0547 & -0.0814 & -0.0357 \\
LPIPS & +0.0428 & +0.0469 & -0.0968 & -0.0340 \\
MSE & +0.0370 & +0.0501 & -0.0868 & -0.0380 \\
\bottomrule
\end{tabular*}
\end{table}
\endgroup
\end{table}

\FloatBarrier
\section{Serving and Cost}
\label{app:serving}

\subsection{Serving}
\label{app:serving-load}
Figure~\ref{fig:ex2}l serves InternSVG-8B text-to-SVG in SGLang 0.5.22 on one H200, with 1 to 64 concurrent requests under greedy decoding and official sampling.
SGLang verifies the drafts of these heads as a chain, and both heads run that way.
Our head verifies windows of 32, 24, 16 and 8, and DFlash's head windows of 16 and 8.
At each load, each head runs at the window that serves it fastest, and Appendix~\ref{app:runs} shows how the identity prices such shorter windows.
Tables~\ref{app:p5-speed} and~\ref{app:p5-accept} give every load and window of Figure~\ref{fig:ex2}l.
At the same window, our head runs within 10\% of DFlash's (Figure~\ref{app:row-judges}b), so the lead at low load comes from the longer window that only a head trained on longer blocks can serve.
From 8 concurrent requests on, both heads do best at windows of 16 or less, and ours runs at 0.91$\times$ to 1.03$\times$ the speed of DFlash's head (Table~\ref{app:p5-speed}).

\begin{table}[htbp]\centering
\begingroup
\renewenvironment{table}[1][]{\begin{minipage}[t]{0.485\textwidth}\vspace{0pt}\centering}{\end{minipage}}%
\renewenvironment{table*}[1][]{\begin{minipage}[t]{0.485\textwidth}\vspace{0pt}\centering}{\end{minipage}}%
\begin{table}[t]
\caption{\textbf{Serving in SGLang, every load and window, speedup.} Throughput speedup over the original model at the same concurrency, InternSVG-8B text-to-SVG on one H200, with both heads verifying chains. The verification window is set separately from the block length of the head. Settings measured in a later job are scaled to the node of the main run by their own rerun of DFlash's head at a window of 16 (factors 0.96 to 1.07). The last column gives the original model's throughput. \textbf{Best} in bold, \underline{second}{} underlined, per row.}
\label{app:p5-speed}
\centering
\footnotesize
\setlength{\tabcolsep}{1.2pt}
\begin{tabular*}{\columnwidth}{@{\extracolsep{\fill}}rr*{6}{c}@{}}
\toprule
 & & \multicolumn{2}{c}{DFlash (block 16)} & \multicolumn{4}{c}{\textbf{Ours} (block 32)} \\
\cmidrule(lr){3-4}\cmidrule(lr){5-8}
 & & \multicolumn{6}{c}{verification window} \\
Concurrency & AR tok/s & 16 & 8 & 32 & 24 & 16 & 8 \\
\midrule
\multicolumn{8}{l}{\textit{Greedy}} \\
1 & 216 & 7.22$\times$ & 4.67$\times$ & \textbf{9.29}$\times$ & \underline{8.31}$\times$ & 7.03$\times$ & 4.65$\times$ \\
2 & 416 & 7.29$\times$ & 4.78$\times$ & \underline{8.69}$\times$ & \textbf{9.03}$\times$ & 7.08$\times$ & 4.62$\times$ \\
4 & 811 & 6.54$\times$ & 4.85$\times$ & \underline{7.12}$\times$ & \textbf{7.77}$\times$ & 6.38$\times$ & 4.53$\times$ \\
8 & 1567 & \textbf{6.19}$\times$ & 4.66$\times$ & 5.11$\times$ & 5.33$\times$ & \underline{5.62}$\times$ & 4.61$\times$ \\
16 & 2886 & \textbf{4.55}$\times$ & 4.18$\times$ & 3.26$\times$ & 3.58$\times$ & 4.25$\times$ & \underline{4.26}$\times$ \\
32 & 5197 & 2.93$\times$ & \textbf{3.17}$\times$ & 2.07$\times$ & 2.34$\times$ & 2.77$\times$ & \underline{3.15}$\times$ \\
64 & 8778 & 1.89$\times$ & \textbf{2.17}$\times$ & 1.26$\times$ & 1.48$\times$ & 1.78$\times$ & \underline{2.12}$\times$ \\
\midrule
\multicolumn{8}{l}{\textit{Official sampling}} \\
1 & 211 & 4.45$\times$ & 3.54$\times$ & \textbf{5.53}$\times$ & \underline{4.98}$\times$ & 4.75$\times$ & 3.63$\times$ \\
2 & 405 & 4.46$\times$ & 3.48$\times$ & \underline{5.36}$\times$ & \textbf{5.62}$\times$ & 4.53$\times$ & 3.74$\times$ \\
4 & 783 & \textbf{4.33}$\times$ & 3.44$\times$ & 3.82$\times$ & \underline{4.17}$\times$ & 4.03$\times$ & 3.37$\times$ \\
8 & 1505 & \textbf{3.22}$\times$ & 2.79$\times$ & 2.36$\times$ & 2.66$\times$ & \underline{3.04}$\times$ & 2.94$\times$ \\
16 & 2776 & 2.39$\times$ & \textbf{2.90}$\times$ & 1.52$\times$ & 1.74$\times$ & 2.25$\times$ & \underline{2.81}$\times$ \\
32 & 5000 & 1.63$\times$ & \underline{2.09}$\times$ & 1.01$\times$ & 1.28$\times$ & 1.56$\times$ & \textbf{2.16}$\times$ \\
64 & 8480 & 1.04$\times$ & \underline{1.36}$\times$ & 0.63$\times$ & 0.74$\times$ & 0.97$\times$ & \textbf{1.38}$\times$ \\
\bottomrule
\end{tabular*}
\end{table}
\hfill\begin{table}[t]
\caption{\textbf{Serving in SGLang, every load and window, acceptance.} Mean accepted tokens per verification step reported by the engine, for the runs of Table~\ref{app:p5-speed}. \textbf{Best} in bold, \underline{second}{} underlined, per row.}
\label{app:p5-accept}
\centering
\footnotesize
\setlength{\tabcolsep}{1.2pt}
\begin{tabular*}{\columnwidth}{@{\extracolsep{\fill}}rr*{6}{c}@{}}
\toprule
 & & \multicolumn{2}{c}{DFlash (block 16)} & \multicolumn{4}{c}{\textbf{Ours} (block 32)} \\
\cmidrule(lr){3-4}\cmidrule(lr){5-8}
 & & \multicolumn{6}{c}{verification window} \\
Concurrency & AR tok/s & 16 & 8 & 32 & 24 & 16 & 8 \\
\midrule
\multicolumn{8}{l}{\textit{Greedy}} \\
1 & 216 & 9.9 & 6.3 & \textbf{13.1} & \underline{11.5} & 9.7 & 6.3 \\
2 & 416 & 10.2 & 6.4 & \underline{12.4} & \textbf{12.8} & 9.7 & 6.2 \\
4 & 811 & 9.5 & 6.2 & \textbf{11.6} & \underline{11.3} & 9.0 & 6.3 \\
8 & 1567 & 9.6 & 6.3 & \textbf{11.4} & \underline{10.5} & 8.7 & 6.1 \\
16 & 2886 & 9.6 & 6.3 & \textbf{12.0} & \underline{10.3} & 9.0 & 6.2 \\
32 & 5197 & 9.9 & 6.4 & \textbf{12.6} & \underline{11.3} & 9.4 & 6.3 \\
64 & 8778 & 10.1 & 6.4 & \textbf{12.8} & \underline{11.4} & 9.5 & 6.3 \\
\midrule
\multicolumn{8}{l}{\textit{Official sampling}} \\
1 & 211 & 6.4 & 4.9 & \textbf{8.2} & \underline{7.3} & 6.8 & 5.1 \\
2 & 405 & 6.5 & 4.9 & \underline{8.3} & \textbf{8.4} & 6.6 & 5.2 \\
4 & 783 & \textbf{6.8} & 4.9 & \underline{6.7} & \underline{6.7} & 6.4 & 4.8 \\
8 & 1505 & \underline{5.5} & 4.3 & \textbf{5.7} & \textbf{5.7} & 5.1 & 4.3 \\
16 & 2776 & \underline{5.5} & 4.4 & \textbf{6.1} & 5.3 & 5.2 & 4.4 \\
32 & 5000 & 6.1 & 4.7 & \textbf{6.6} & \underline{6.3} & 5.8 & 4.7 \\
64 & 8480 & 6.1 & 4.7 & \textbf{6.8} & \underline{6.2} & 5.6 & 4.6 \\
\bottomrule
\end{tabular*}
\end{table}
\endgroup
\end{table}

\subsection{One-Off Cost of Image-to-SVG}
\label{app:oneoff}
Image-to-SVG pays for preprocessing, vision encoding and prefill once per image.
As decoding gets faster, this one-off part grows to about a tenth of the time (Figure~\ref{app:row-judges}c).

\FloatBarrier
\section{Limitations}
\label{app:limits}
Like every lossless speculative decoder, Sightline keeps the loops of the target.
Loop control is shared by every decoder and catches exact repeats only.
Like other speculative decoders, it gains more when the target is confident and few requests share the hardware.
Like other block drafters, it needs a draft head trained on the outputs of each target, and a head serves windows up to the block length it was trained for.
The theory gives each position one candidate set, which holds for block drafters and approximately for drafters that condition on earlier draft tokens.
The lossless guarantee holds in exact arithmetic, and all timings use one GPU type.

\end{document}